\documentclass[twoside,leqno,twocolumn]{article}

\usepackage[letterpaper]{geometry}

\usepackage{ltexpprt}

\usepackage{amsmath,amssymb,amsfonts}
\usepackage{algorithmic}
\usepackage{graphicx}
\usepackage{textcomp}
\usepackage{xcolor}
\usepackage{hyperref}
\usepackage{threeparttable}
\usepackage{subfigure}
\usepackage{color}
\usepackage{subfloat}
\usepackage{algorithm}
\usepackage{amsfonts}
\usepackage{latexsym}
\usepackage{soul}
\usepackage{enumitem}
\usepackage[nopar]{lipsum} 

\setstcolor{red}

\begin{document}



\title{\Large Fairness-aware Outlier Ensemble}
\author{Haoyu Liu\thanks{Zhejiang University, \{haoyu\_liu, s18he, cjm\}@zju.edu.cn}
\and Fenglong Ma\thanks{Pennsylvania State University, fenglong@psu.edu}
\and Shibo He\footnotemark[1]
\and Jiming Chen\footnotemark[1]
\and Jing Gao\thanks{University at Buffalo, jing@buffalo.edu}
}

\date{}

\maketitle


\fancyfoot[R]{\scriptsize{Copyright \textcopyright\ 20XX by SIAM\\
Unauthorized reproduction of this article is prohibited}}





\begin{abstract} \small\baselineskip=9pt 
Outlier ensemble methods have shown outstanding performance on the discovery of instances that are significantly different from the majority of the data. However, without the awareness of fairness, their applicability in the ethical scenarios, such as fraud detection and judiciary judgement system, could be degraded.
In this paper, we propose to reduce the bias of the outlier ensemble results through a fairness-aware ensemble framework.
Due to the lack of ground truth in the outlier detection task, the key challenge is how to mitigate the degradation in the detection performance with the improvement of fairness.
To address this challenge, we define a distance measure based on the output of conventional outlier ensemble techniques to estimate the possible cost associated with detection performance degradation. 
Meanwhile, we propose a post-processing framework to tune the original ensemble results through a stacking process so that we can achieve a trade off between fairness  and detection performance. Detection performance is measured by the area under ROC curve (AUC) while fairness is measured at both group and individual level. 
Experiments on eight public datasets are conducted.
Results demonstrate the effectiveness of the proposed framework in improving fairness of outlier ensemble results. We also analyze the trade-off between AUC and fairness.
\end{abstract}

\section{Introduction}
Machine learning methods have shown huge success in automated decision making in terms of decision performance measures such as accuracy and recall rate. Recently, the ethical issues regarding machine learning results have been recognized~\cite{chouldechova2017fair,hardt2016equality,lahoti2019ifair}, and there has been a growing interest in designing fairness-aware machine learning methods. 
Great efforts have been made to address this fairness issue in the tasks such as classification~\cite{agarwal2018reductions,kamishima2012fairness,menon2018cost}, clustering~\cite{backurs2019scalable,chen2019proportionally,chierichetti2017fair}, and representation learning~\cite{bolukbasi2016man,creager2019flexibly,moyer2018invariant}. However, for the outlier detection task, fairness-aware techniques are relatively unexplored.

Outlier detection is the process to distinguish outliers from inliers~\cite{chandola2009anomaly}, which has potential applications in multiple gender/racial related scenarios, such as credit card/insurance fraud detection and judiciary judgement system. Thus, the applicability of outlier detection methods could be degraded if there are no constraints on the unethical outputs.
In addition,  for the outlier detection task, outlier ensemble methods which rely on multiple different outlier detectors are usually employed to obtain superior detection performance~\cite{campos2018unsupervised,lazarevic2005feature,rayana2016less,schubert2012evaluation,zimek2013subsampling}. To develop techniques that promote the fairness of outlier ensemble results, one strategy is to transform every employed outlier detector into a fairness-aware outlier detector. For example, for LOF outlier detector~\cite{breunig2000lof}, P \emph{et~al.}~\cite{p2020fair} proposed FairLOF to reduce bias within the results. However, to achieve superior performance, outlier ensemble may consist of multiple detectors based on different outlier detection algorithms, and it could be inefficient to transform base detectors into fairness-aware detecotrs separately. Moreover, even when base detectors are fairness-aware, the process of combining base detectors in the ensemble could still introduce bias into the final outlier detection outputs.   


To tackle these issues, in this paper, we propose to develop a fairness-aware outlier ensemble framework that aims to achieve fair outlier detection results. Any conventional outlier detection method could be plugged into the framework as a base detector. 
The key idea is to post-process  ensemble results considering both fairness and AUC performance. This is still a non-trivial task due to the following two major challenges.

\smallskip
\noindent\textbf{Challenge I: }\emph{How to formally define the fairness measure for the unsupervised outlier detection task?}
Outlier detection methods are mostly in an unsupervised manner, where the ground truth information is unavailable during the deployment. Thus, some widely used fairness measures such as equalized odds and equal opportunity~\cite{hardt2016equality} are not applicable. To tackle this challenge,  we define group and individual fairness measure based on outlier scores of individual outlier detectors  without using any ground truth labels.

For \textbf{group fairness}, the proposed definition is largely motivated by \emph{demographic parity}~\cite{dwork2012fairness,kusner2017counterfactual}. Inspired by ``\emph{four-fifth rule}'', demographic parity measures fairness through the difference in the acceptance rates of the applicants from the different ethic groups. In outlier detection results, the outlier score for each instance can be regarded as an estimate on how likely this instance could be selected as an outlier. Therefore, the average of outlier scores of the instances from one ethic group describes the adverse impact towards this group in outlier detection. Thus, we define the group fairness of the outlier detection task by the difference in the average outlier scores among different ethic groups. 

For \textbf{individual fairness}, we aim to measure the difference in the outlier scores of two instances from different ethic groups. If these two instances are similar, then the higher the difference, the larger the bias. Also, considering the challenge brought by the lack of ground truth labels, we cannot directly apply previous works which use labels to measure instance similarity~\cite{berk2017convex}. Therefore, we first propose to quantify the similarity between instances based on original feature values excluding sensitive features. Then we use the obtained similarity values to derive a weight for each instance pair, and a weight sum operation on all instance-wise differences is conducted to measure the individual fairness.

\smallskip
\noindent
\textbf{Challenge II: }\emph{How to prevent the degradation of detection performance while improving fairness?} 
In supervised tasks, during the training procedure, a trade-off optimization problem between fairness and the explicitly measured performance (e.g., accuracy) is usually formulated to obtain a compromise~\cite{goel2018non,krasanakis2018adaptive,menon2018cost}. However, outlier detection is unsupervised, and thus directly applying previous strategies is not possible. 

To tackle this challenge, we employ originally generated outlier scores as an anchor. The key idea is as follow: First, with an arbitrarily employed conventional outlier ensemble method, the outlier scores for all instances are produced. Since conventional outlier ensembles do not consider the fairness issue, these generated outlier scores could be biased against some ethic groups, though they may have superior detection performance in terms of measures such as AUC. Then, we use these outlier scores, which can be represented as an outlier score vector, as the target outlier scores, i.e., \emph{target outlier vector}. During the improvement of the fairness, we propose to minimize the distance between newly generated ensemble result and the target outlier vector. Since the instances with higher outlier scores are more important for the outlier detection task, we also propose to plug in a weight that measures the importance of each instance into the framework to preserve the outlier detection performance.  

\smallskip
\noindent
\textbf{Framework. }Using a stacking structure, we fuse the aforementioned strategies for solving the two challenges into one framework. The framework is based on the solution to an optimization problem, which is formulated to transform original ensemble results into a fairness-aware results. Within the optimization formulation, a parameter $\alpha$ is designed to tune the ``fairness degree'': a higher $\alpha$ induces a fairer outlier detection result, and vice versa. The closed-form solutions are derived. Extensive experiments based on eight public datasets are conducted to illustrate the effectiveness of the solutions. Besides, two different outlier ensemble methods are leveraged to show that, the proposed framework is able to transform any conventional outlier ensemble method into a fairness-aware method.
In summary, our main contributions are as follows:
\begin{itemize}[noitemsep,topsep=0pt]
\item To the best of our knowledge, this is the first work addressing the important fairness issue for the outlier ensemble problem. 

\item Both group and individual fairness measures are defined under the setting of unsupervised outlier detection task.

\item An effective framework is designed to transform any outlier ensemble method into fairness-aware outlier ensemble via a stacking structure. Closed-form solutions to an optimization problem that balances fairness and AUC measures are given.

\item Based on a broad range of datasets, we provide experimental analysis on the effectiveness and the fairness cost of the proposed framework.
\end{itemize}

\section{Notations}
\label{sec:Notations}
Let $\mathcal{D} = {(X_i,a_i)}_{i=1}^n$ be denoted as a dataset, where $n$ is the number of instances. For each instance $i$, $X_i$ denotes the feature vector, and $a_i \in \mathbf{g} = \{0,1,...,v\}$ represents the value of the protected attribute. If there are multiple protected attributes, the proposed framework is still applicable by encoding values from different protected attributes into one attribute.
Before the ensemble procedure, $k$ base outlier detectors are firstly conducted, producing an outlier score matrix $\mathbf{S} \in \mathbb{R}^{k \times n}$, where $\mathbf{S}_{i,.}$ denotes the score vector generated by the $i$th detector for all of the observations, and $\mathbf{S}_{.,j}$ denotes the score vector generated by all of the base detectors for the $j$th instance.

Then, we use $\mathbf{t} \in \mathbb{R}^n$ to denote the \emph{target outlier vector}, which contains outlier scores generated by the base outlier ensemble method. Any mature outlier ensemble methods can be employed as the base outlier ensemble method to produce $\mathbf{t}$.
Generated from a stacking structure, the newly obtained fairness-aware outlier score vector is represented by $\mathbf{y} \in \mathbb{R}^{n}$. Further, utilizing an ensemble weights vector $\mathbf{W} \in \mathbb{R}^k$, we derive $\mathbf{y}$ through $\mathbf{y} = \mathbf{W}\mathbf{S}$. Thus, our goal falls to producing the weights vector $\mathbf{W}$ of base detectors so that the generated $\mathbf{y}$ can satisfy both fairness and performance requirements.

\section{Methods}
\label{sec:Methods}
A stacking structure is proposed to solve the problem. We first introduce the overall framework and then provide the corresponding solutions for the scenarios with group and individual fairness respectively. 

\subsection{Framework}
\begin{figure}
    \centering
    \includegraphics[width=0.45\textwidth,height=0.3\textwidth]{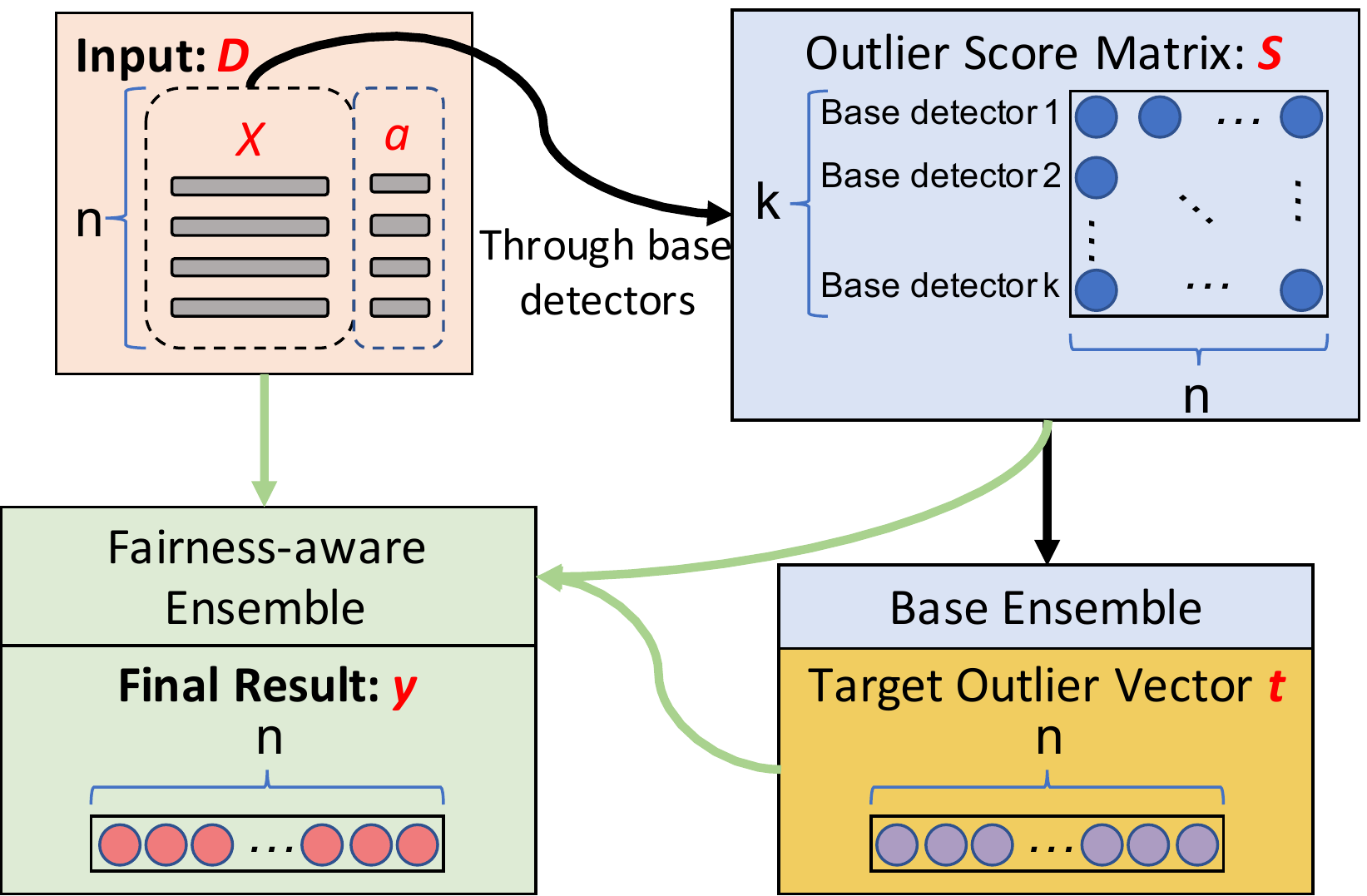}
    \caption{Overall pipeline for the proposed framework.}
    \label{fig:framework}
\end{figure}

The overall pipeline is presented in Fig. \ref{fig:framework}.
Based on outlier scores matrix $\mathbf{S}$ calculated from base detectors, 
the base ensemble method is firstly conducted to generate target outlier vector $\mathbf{t}$.
Then, both $\mathbf{S}$ and $\mathbf{t}$ together with the values ${(X_i,a_i)}_{i=1}^n$ from original dataset $\mathcal{D}$ are leveraged to calculate final result $\mathbf{y}$ through the optimization procedure in the fairness-aware ensemble structure.
We formulate this optimization process as follows:
\begin{equation}
\label{equa:framework}
\text{minimize} \;f_1(\mathbf{y},\mathbf{t}) + \alpha f_2(\mathbf{y},\mathcal{D}),
\end{equation}
where $f_1$ denotes a function quantifying the difference between $\mathbf{y}$ and $\mathbf{t}$, $f_2$ represents the measurements of fairness, and $\alpha$ is a trade-off parameter to control the relative importance of two functions.
Assume the outlier score vector $\mathbf{t}$ obtained from the base outlier ensemble method is an optimal result.
Thus, the larger the difference between $\mathbf{y}$ and $\mathbf{t}$, the higher the degradation of the detection performance may be induced.
For the $f_2$ function, we consider both group-level and individual-level fairness to measure the bias. The less the value of $f_2$, the fairer the result of $\mathbf{y}$.
Finally, by minimizing both $f_1$ and $f_2$ together, the proposed framework can improve fairness while indirectly preventing large AUC drop of the detection results.
Next, we will introduce the details of $f_1$, $f_2$, and our proposed solutions, respectively.

\subsection{$f_1$ Function}
we first introduce how to measure the distance between output $\mathbf{y}$ and the original ensemble result $\mathbf{t}$.
Without the loss of generality, $\mathbf{t}$ is firstly normalized into $[0,1]$, where the closer the instance's score to the $1$, the higher possibility to be an outlier.
Therefore, the instances with larger outlier scores are more meaningful for the objective to identify outliers , which are further expected to have less degradation during the improvement of fairness.
We design $f_1$ function to quantify their difference:
\begin{equation}
\label{equa:f_1_1}
f_1(\mathbf{y},\mathbf{t}) = \sum_{i=1}^{n}\beta_i(Y_i - T_i)^2 = \sum_{i=1}^{n}\beta_i(\mathbf{W} \cdot \mathbf{S}_{.,i} - T_i)^2,
\end{equation}
where $Y_i = \mathbf{W} \cdot \mathbf{S}_{.,i}$ denotes the obtained outlier score for the instance $i$ after the ensemble of base detectors based on weights $\mathbf{W}$.
$\beta_i$ is leveraged to denote the importance of each instance.

To calculate $\beta_i$, we first rank all of the instances in ascending order according to the normalized $\mathbf{t}$.
Then, $\beta_i$ is derived by:
\begin{flalign*}
    \beta_i = e^{\frac{\text{rank}(i)}{n}},
\end{flalign*}
where $\text{rank}(i)$ denotes the rank of the $i$th instance.
The scale of $\beta_i$ is $(1,e]$.
Clearly, the instances having larger outlier scores in $\mathbf{t}$ have higher rank values so that the assigned importance weights $\beta$ are larger.
Consequently, the change of the outlier scores for those instances exhibiting abnormal characteristics will invoke a larger increase of $f_1$ than those normal instances.
Therefore, during the improvement of fairness, minimization process will preferentially increase the differences of outlier scores from $\mathbf{t}$ and $\mathbf{y}$ for those instances showing less abnormality status according to $\mathbf{t}$.
Inserting Eq.~(\ref{equa:f_1_1}) into Eq.~(\ref{equa:framework}), the objective function can be presented as:
\begin{flalign*}
\underset{\mathbf{W}}{\text{minimize}} \;\sum_{i=1}^{n}\beta_i(\mathbf{W} \cdot \mathbf{S}_{.,i} - T_i)^2 + \alpha f_2(\mathbf{W},\mathbf{S},{(X_i,a_i)}_{i=1}^n).
\end{flalign*}

\subsection{$f_2$ Function}
In this subsection, we demonstrate the definitions of group fairness and individual fairness for the outlier detection problem respectively.

\subsubsection{Group Fairness}
Demographic parity~\cite{dwork2012fairness} requests that every group should receive the same positive rate.
Similarly, we use average outlier scores to indicate the group positive rate in outlier detection task.
The definition is given by:
\begin{flalign*}
\text{DP} = \frac{1}{N}{\underset{p,q \in \mathbf{g},p\neq q}{\sum}(\frac{\sum_{\{i|a_i=p\}}Y_i}{|\{i|a_i = p\}|} - \frac{\sum_{\{j|a_j=q\}}Y_j}{|\{j|a_j = q\}|})^2},
\end{flalign*}
where $p$ and $q$ represent different groups.
$\{i|a_i=p\}$ denotes the index set of outliers belonging to group $p$, while $\{j|a_j=q\}$ denotes the index set of outliers belonging to group $q$.
$N = |\mathbf{g}|(|\mathbf{g}|-1)/2$ denotes the number of pairs of the groups.
The measurement sums up squared differences between all of $N$ group pairs in terms of their average outlier scores.
Since DP is describing the bias among groups, thus the group fairness is improved when the value of DP is reducing.

\subsubsection{Individual Fairness}
Most previous works on individual fairness use the ground truth to indicate the real distance between two instances, such as \cite{berk2017convex,kusner2017counterfactual}.
However, it is inapplicable in unsupervised outlier ensemble problem.
We define the individual fairness as:
\begin{flalign*}
\text{IF} = \frac{1}{N}{\underset{p,q\in \mathbf{g},p \neq q}{\sum}\frac{\underset{\{(i,j)|a_i=p,a_j=q\}}{\sum}d(X_i,X_j)(Y_i-Y_j)^2}{|\{i|a_i = p\}||\{j|a_j = q\}|}}.
\end{flalign*}
Here, we use $\{(i,j)|a_i=p,a_j=q\}$ to denote the pair of instances from two different groups $p$ and $q$.
$d(X_i,X_j)$ indicates the original relation between a pair of instances $i$ and $j$.
We define $d(X_i,X_j)$ by:
\begin{flalign*}
d(X_i,X_j) = e^{-f_\text{Distance}(X_i,X_j)},
\end{flalign*}
where $f_\text{Distance}(.,.)$ is the estimation of the real distance between two instances, which is based on original feature values $X_i$ and $X_j$.
The larger the difference between these two instances, the higher the value of $f_\text{Distance}(.,.)$ , and the lower the value of $d(.,.)$.
Consequently, if two instances $i$ and $j$ are originally different, the value of $d(X_i,X_j)$ will be small. Thus, the difference between $Y_i$ and $Y_j$ will induce little increase of IF.
Conversely, if two instances are very similar according to the distance estimation, then the difference between $Y_i$ and $Y_j$ will largely imply a high bias.
To summarize, the above procedure aims to obtain similar outputs $Y_i$ and $Y_j$ for those similar instances $i$ and $j$ during the minimization procedure in Eq.~(\ref{equa:framework}).
Without the loss of generality, we use a euclidean distance measurement as $f_\text{Distance}$, which will be further scaled by min-max normalization and then used to calculate $d(.,.)$.

\subsection{Solutions}
Since the definitions of $f_1$ and $f_2$ functions are both convex, we derive the closed-form solutions for two definitions of fairness.

\subsubsection{Solution for Group Fairness}

Replacing $f_2$ by group fairness measurement DP, we have the following objective function:

\begin{flalign*}
&\underset{\mathbf{W}}{\text{minimize}} \;\text{L}_1(\mathbf{W}) = \sum_{i=1}^{n}\beta_i(\mathbf{W} \cdot \mathbf{S}_{.,i} - T_i)^2 +&&\\\nonumber
            & \frac{\alpha}{N}{\underset{p,q \in \mathbf{a},p\neq q}{\sum}(\frac{\sum_{\{i|a_i=p\}}\mathbf{W} \cdot \mathbf{S}_{.,i}}{|\{i|a_i = p\}|} - \frac{\sum_{\{j|a_j=q\}}\mathbf{W} \cdot \mathbf{S}_{.,j}}{|\{j|a_i = q\}|})^2}.
\end{flalign*}
Solutions can be obtained by calculating derivative w.r.t. $\mathbf{W}$:
\begin{flalign*}
\mathbf{W} = 
(\mathbf{B}\odot\mathbf{S}^\intercal \mathbf{S} + \frac{\alpha}{N}{\underset{p,q \in \mathbf{g},p\neq q}{\sum} \mathbf{d}_{pq}\mathbf{d}_{pq}^\intercal})^{-1}(\mathbf{B}\odot\mathbf{S}^\intercal \mathbf{t}),
\end{flalign*}
where $\mathbf{d}_{pq}$ represents $\frac{\sum_{\{i|a_i=p\}} \mathbf{S}_{.,i}}{|\{i|a_i = p\}|} - \frac{\sum_{\{j|a_j=q\}} \mathbf{S}_{.,j}}{|\{j|a_j = q\}|}$, $\mathbf{B} \in \mathbb{R}^{k \times n}$ denotes the stack of $k$ $\beta$. Details can be found in the supplementary materials.

\subsubsection{Solution for Individual Fairness}
Next, we introduce the solution with individual fairness.
Inserting $\text{IF}$ as the $f_2$ function in the framework, we have:
\begin{flalign*}
&\underset{\mathbf{W}}{\text{minimize}} \;\text{L}_2(\mathbf{W}) = \sum_{i=1}^{n}\beta_i(\mathbf{W} \cdot \mathbf{S}_{.,i} - T_i)^2 +&\\\nonumber
&\frac{\alpha}{N}{\underset{p,q\in \mathbf{g},p \neq q}{\sum}\frac{\underset{\{(i,j)|a_i=p,a_j=q\}}{\sum}d(X_i,X_j)(\mathbf{W} \cdot \mathbf{S}_{.,i}-\mathbf{W} \cdot \mathbf{S}_{.,j})^2}{|\{i|a_i = p\}||\{j|a_j = q\}|}}.
\end{flalign*}
Similarly, we calculate the derivative w.r.t. $\mathbf{W}$. The solution is as follows:
\begin{flalign*}
&\mathbf{W} = 
[(\mathbf{B}\odot\mathbf{S}^\intercal) \mathbf{S} + &\\\nonumber
&\frac{\alpha}{N}{\underset{p,q \in \mathbf{g},p\neq q}{\sum}\frac{(\mathbf{M}_{pq} \odot \mathbf{D}_{pq}^\intercal)\mathbf{D}_{pq}\mathbf{W}}{|\{i|a_i = p\}||\{j|a_j = q\}|}}]^{-1}[(\mathbf{B}\odot\mathbf{S}^\intercal) \mathbf{t}],
\end{flalign*}
where $\mathbf{M}_{pq} \in \mathbb{R}^{k\times{|\{i|a_i=p\}||\{j|a_j=q\}|}}$ is the stack of $k$ $\mathbf{m}_{pq}$, $\mathbf{m}_{pq} = \{d(X_i,X_j)|a_i=p,a_j = q\} \in \mathbb{R}^{|\{i|a_i=p\}||\{j|a_j=q\}|}$. $\mathbf{D}_{pq} = \{\mathbf{S}_{.,i} - \mathbf{S}_{.,j} |a_i = p,a_j = q\} \in \mathbb{R}^{{|\{i|a_i=p\}||\{j|a_j=q\}|}\times k}$. Details can be found in the supplementary materials.

\section{Experiments}
\label{sec:Experiments}
In this section, we conduct experiments to evaluate the proposed methods. We aim to answer the following questions. \textbf{Q1:} Is the proposed framework effective to reduce bias while increasing the trade-off parameter $\alpha$? \textbf{Q2:} How the detection performance change during the improvement of fairness? \textbf{Q3:} Is the proposed distance measurement in $f_1$ function effective for reducing the AUC loss?

\subsection{Experimental Settings}
We first introduce the public datasets used in the experiments.
Then, we present the constitutions of the base detectors as well as two base ensemble methods, which will be used to generate outlier scores $\mathbf{S}$ and {target outlier vector} $\mathbf{t}$.
Last, we provide a comparison $f_1$ function and give the solution, which will be further used to verify the effectiveness of the plugged weights parameter $\beta$.

\subsubsection{Datasets}
Eight public datasets are used to demonstrate the performance of the designed framework. 
The statistics of four datasets are summarized in Table \ref{tab:datasets}. The other four datasets and the corresponding experimental results will be given in the supplementary materials due to space limitations.
\begin{table}[ht]
\centering
\begin{threeparttable}
\caption{Statistics for the datasets.}\label{tab:datasets}
\begin{tabular}{@{} ccc @{}}\hline
    \bfseries Datasets &\bfseries Inlier/Outlier Size & \bfseries Groups\\\hline
    \bfseries Communities   &1,717/277 &4 \\
    \bfseries German Credit    &700/300 &4 \\
    \bfseries Annthyroid & 6,666/534 & 2\\
    \bfseries Cardio   &1,655/176 & 2 \\\hline
    
\end{tabular}
\begin{tablenotes}
\centering
      \footnotesize
      \item ``Groups'' denotes the number of the protected groups.
    \end{tablenotes}
    \end{threeparttable}
\end{table}

\begin{itemize}[noitemsep,topsep=0pt]
\item \textbf{Communities Dataset}\footnote{\url{http://archive.ics.uci.edu/ml/datasets/communities+and+crime}} contains census and judicial information of the communities within US.
The original dataset has $4$ attributes representing the percentage of the population, including African American, Caucasian, Asian, and Hispanic heritage separately.
For each instance, we select the ethic group with the largest population as the ethic label.
Thus, there are $4$ groups in total.
The original ground truth of the dataset is the total number of violent crimes per 100K population (crime rate).
To use this dataset in the outlier detection problem, we regard the instances with ``$>0.5$'' values in the crime rate as the outliers, the instances with ``$<=0.5$'' of the crime rate as the inliers.

\item \textbf{German Credit Dataset}\footnote{\url{https://archive.ics.uci.edu/ml/datasets/statlog+(german+credit+data)}} contains credit records such as personal status, credit score, credit amount, age, etc. 
We use the numeric version, which has encoded the qualitative descriptions into numerical values.
$23$ attributes are used to describe the credit-related information for one person in the dataset.
The $24$th attribute labels the ``good'' or ``bad'' for the credit risk of the corresponding person.
We treat the instances with ``good'' labels as the inliers and the others as the outliers.
Four protected groups are generated through the combination of personal status and gender information.

\item \textbf{Annthyroid Dataset}\footnote{\url{http://odds.cs.stonybrook.edu/annthyroid-dataset/}} originates from a classification dataset for thyroid disease.
The instances are labeled as outliers and inliers.
Since the dataset does not have the protected attribute, we manually add a protected attribute containing two different ethic groups.

\item \textbf{Cardiotocogrpahy (Cardio) Dataset}\footnote{\url{http://odds.cs.stonybrook.edu/cardiotocogrpahy-dataset/}} 
measures the fetal heart rate and uterine contraction features.
One attribute denotes the ``normal'' and ``pathologic'' of the fetal states, where ``normal'' instances are treated as inliers, while ``pathologic'' instances are outliers. Similarly, we add a protected attribute with two ethic groups for this dataset.

\end{itemize}

Among these datasets, Communities dataset and German Credit dataset are generally used for fairness research, while the other datasets are originally used for outlier detection.
To evaluate fair outlier detection problem based on these datasets, we add ground truth of outliers for Communities dataset and German Credit dataset according to their original labels.
For the outlier detection datasets, we manually add one protected attribute to denote the protected group.
Instances are sampled into different groups to induce bias.

\begin{figure*}[htb!]
    \centering
    \subfigure[Communities Dataset.]{
        \includegraphics[height=0.17\textwidth,width=0.23\textwidth]{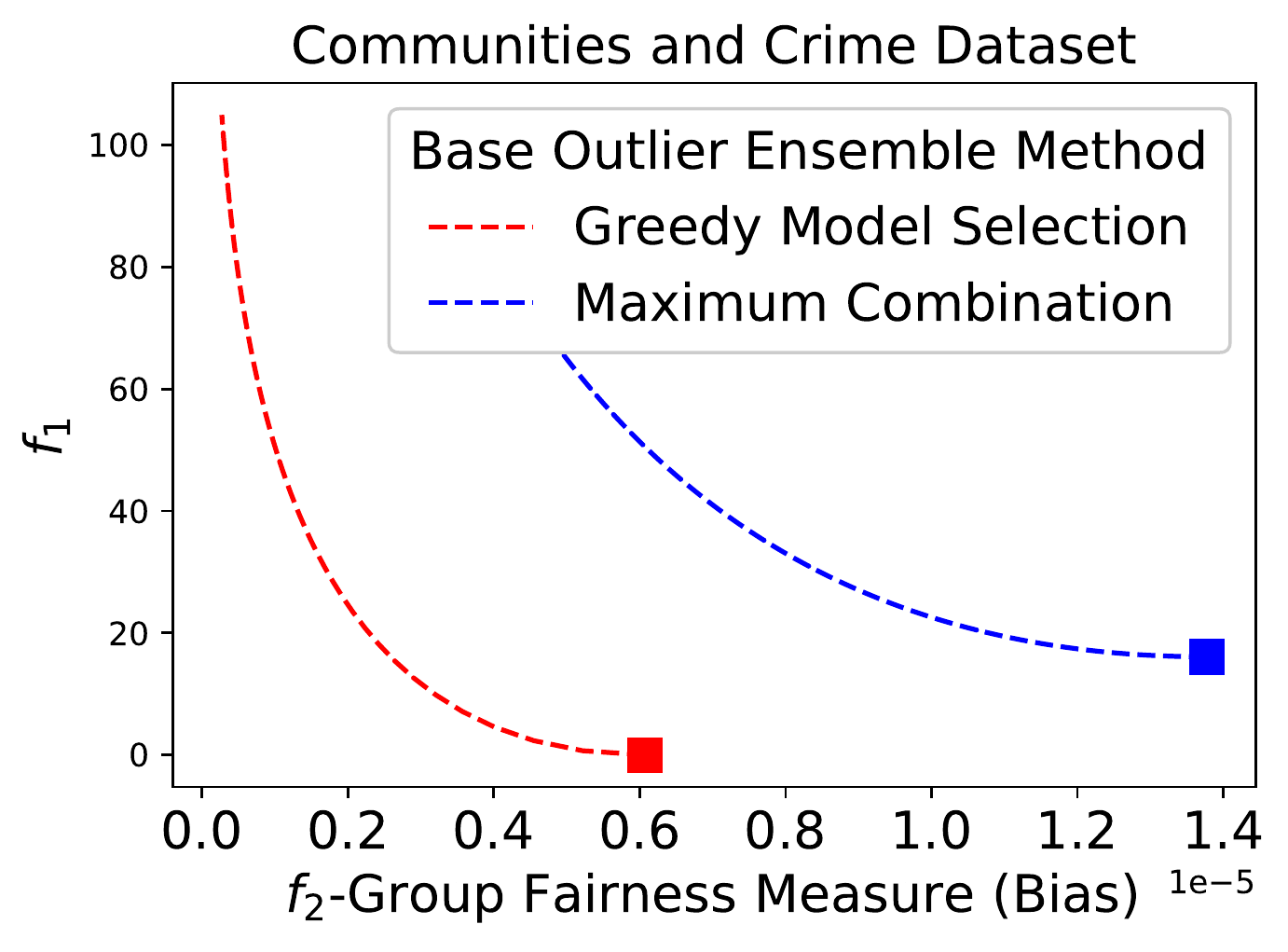}
    }
    \subfigure[{German Credit Dataset.}]{
        \includegraphics[height=0.17\textwidth,width=0.23\textwidth]{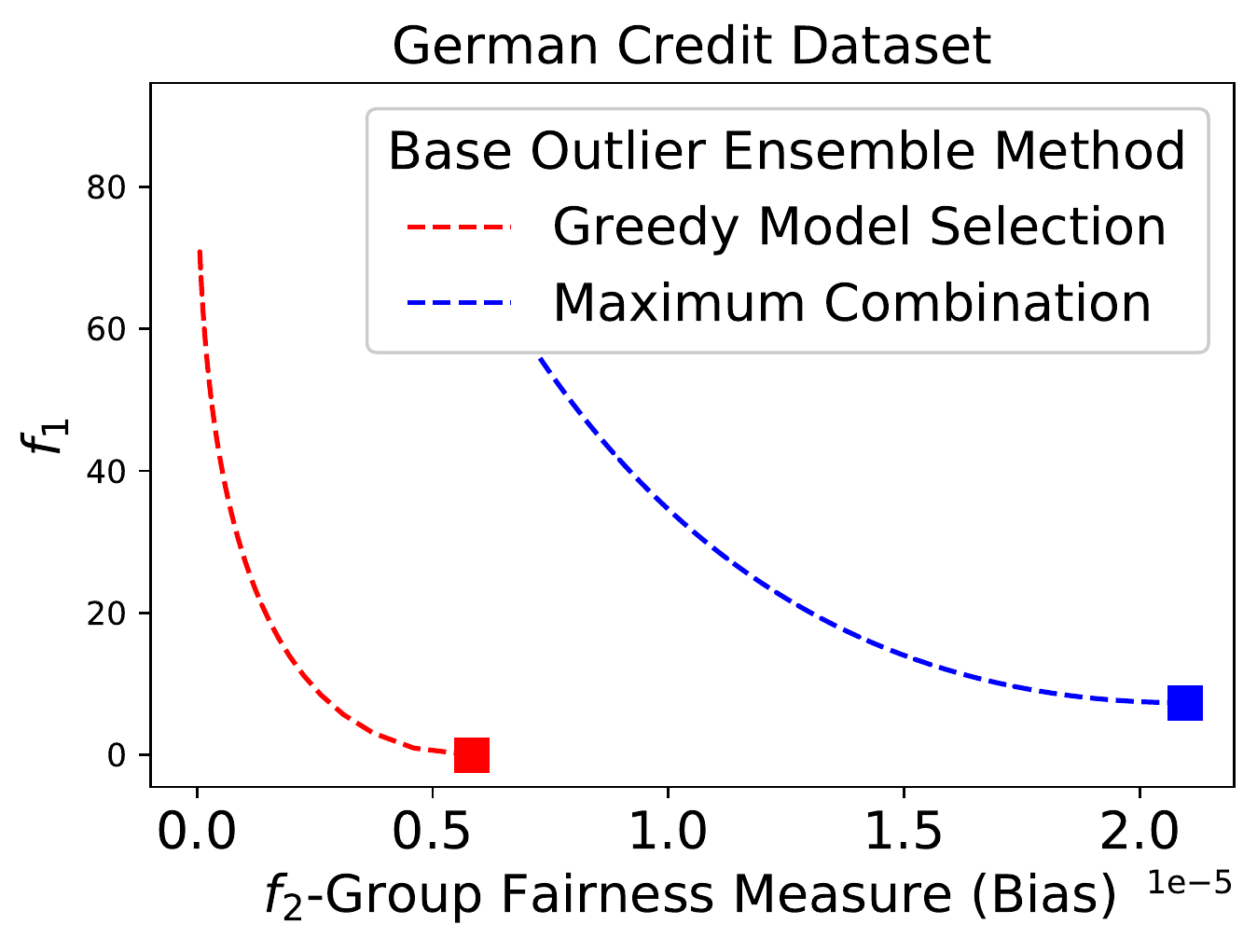}
    }    
    \subfigure[Annthyroid Dataset.]{
        \includegraphics[height=0.17\textwidth,width=0.23\textwidth]{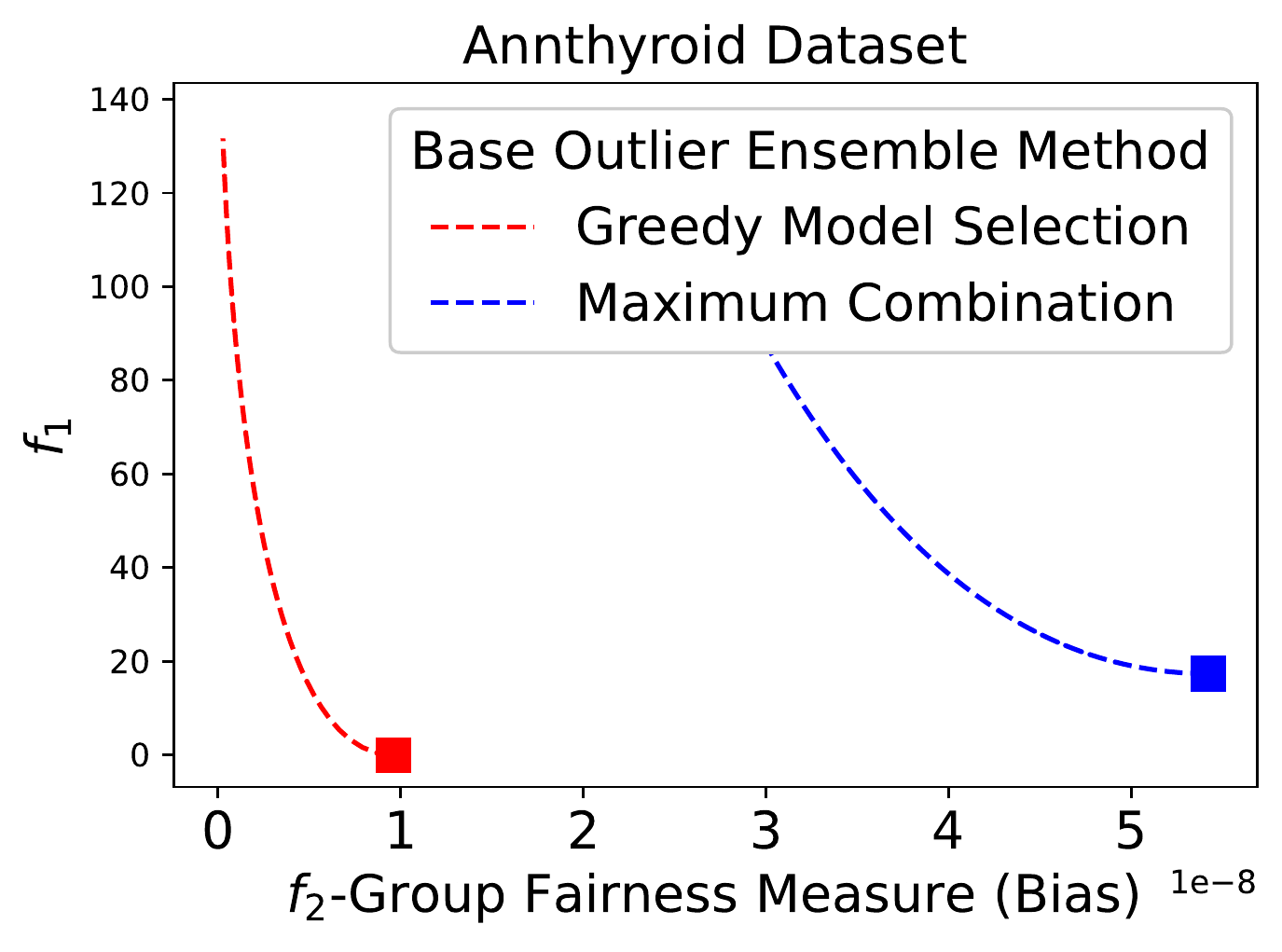}
    }
    \subfigure[Cardiotocogrpahy Dataset.]{
        \includegraphics[height=0.17\textwidth,width=0.23\textwidth]{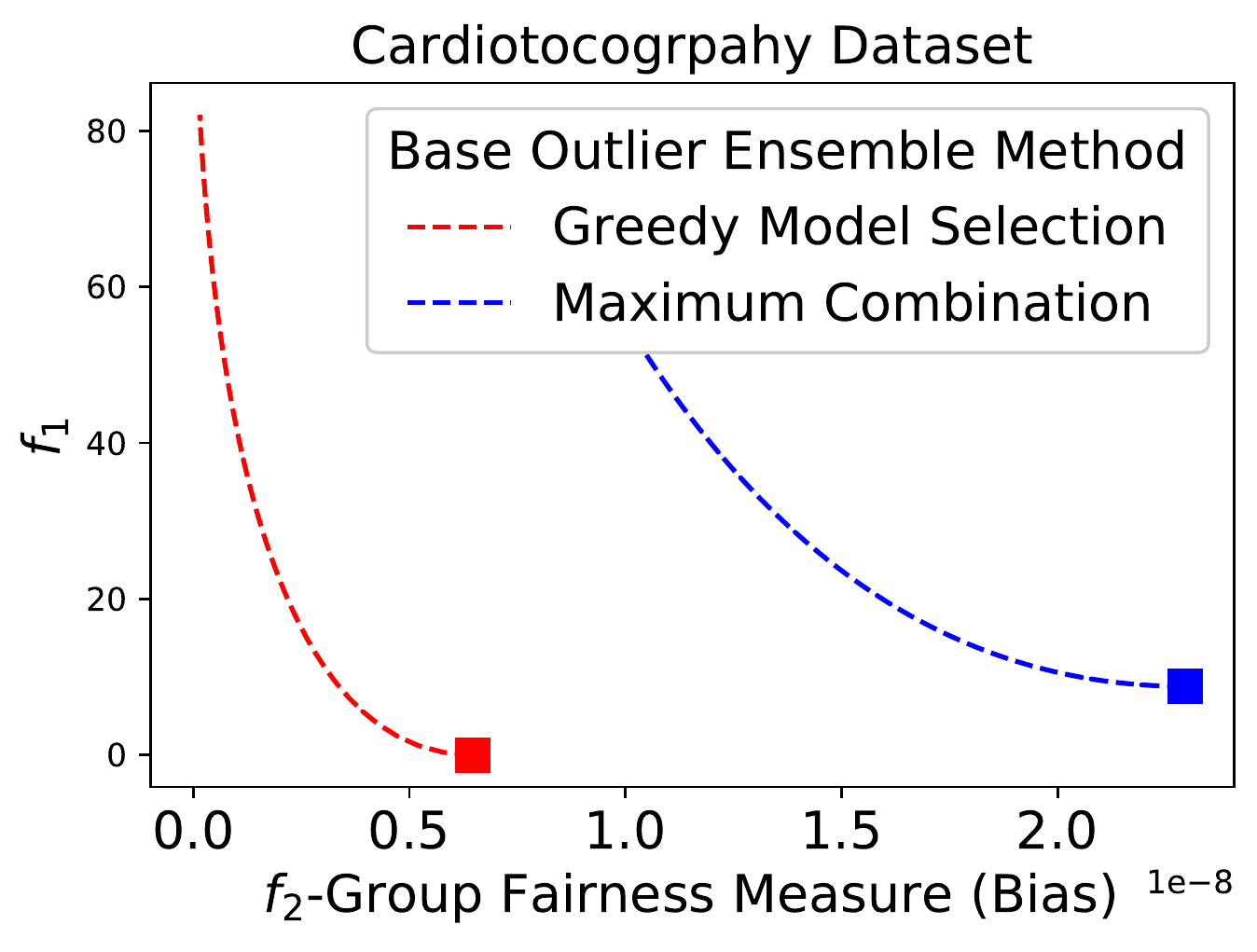}
    }
    \caption{$f_2$-$f_1$ curves for group fairness.}
    \label{fig:f1f2groupfairness}

\end{figure*}

\begin{figure*}[htb!]
    \centering
     \subfigure[Communities Dataset.]{
        \includegraphics[height=0.17\textwidth,width=0.23\textwidth]{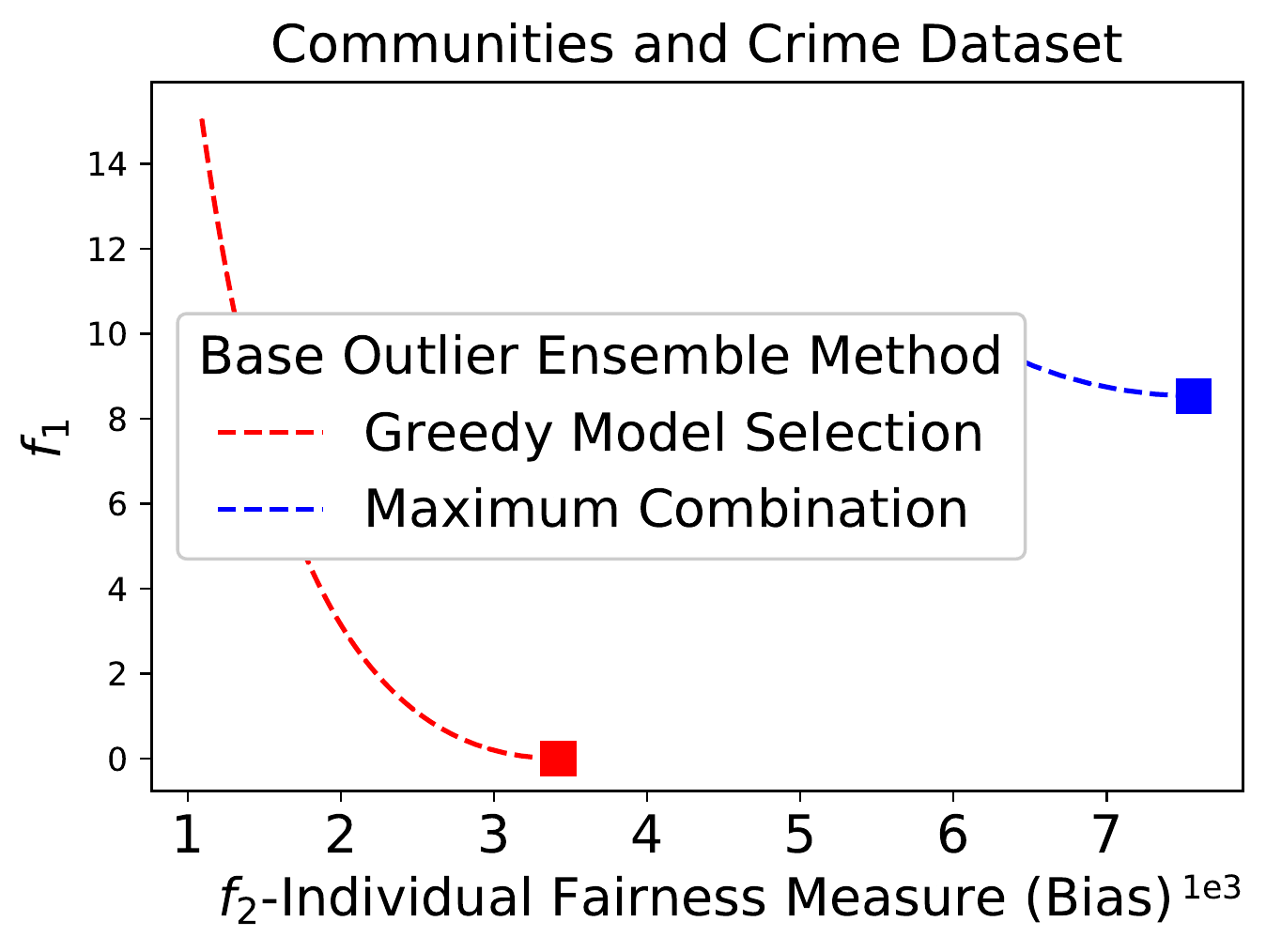}
    }
    \subfigure[{German Credit Dataset.}]{
        \includegraphics[height=0.17\textwidth,width=0.23\textwidth]{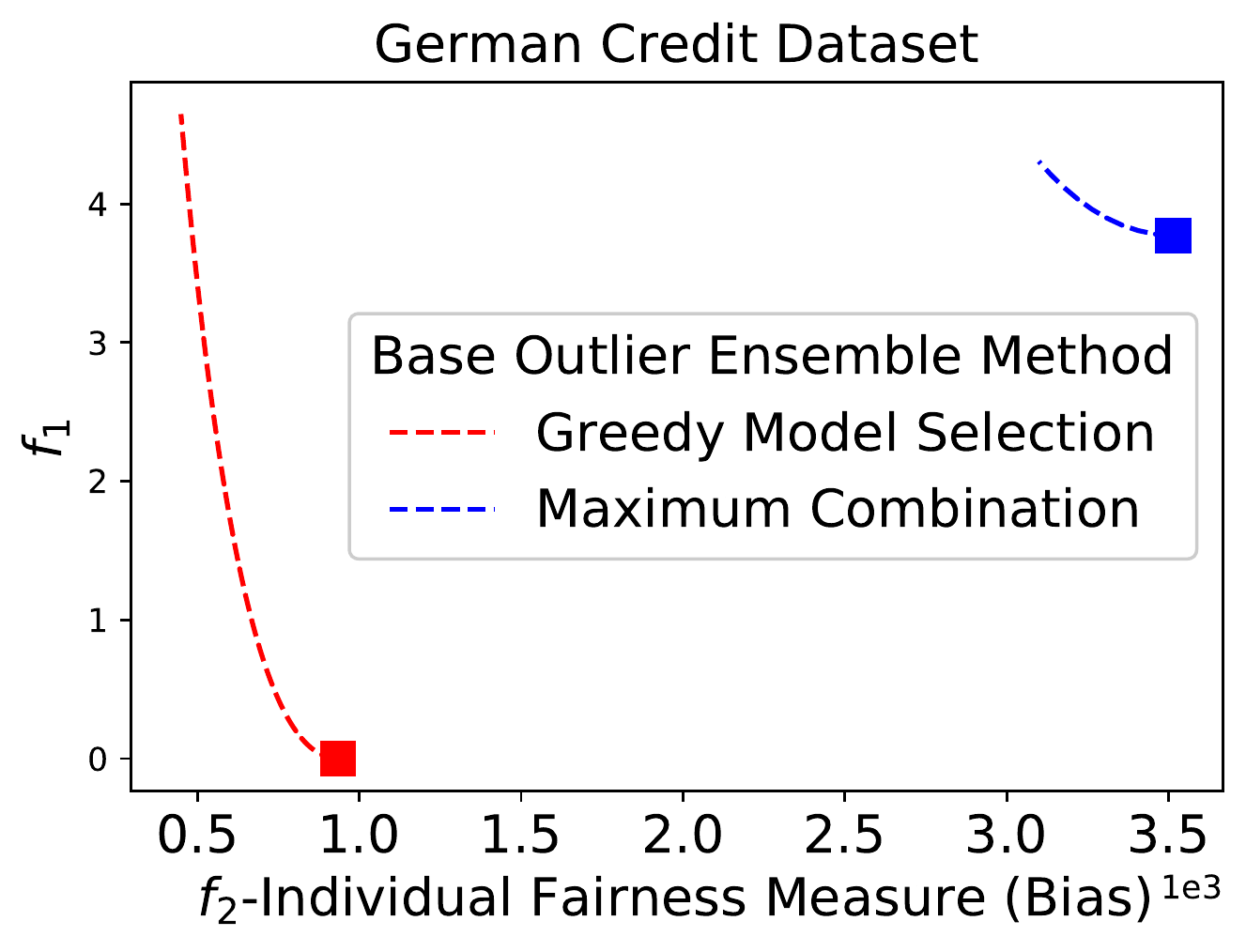}
    }    
    \subfigure[Annthyroid Dataset.]{
        \includegraphics[height=0.17\textwidth,width=0.23\textwidth]{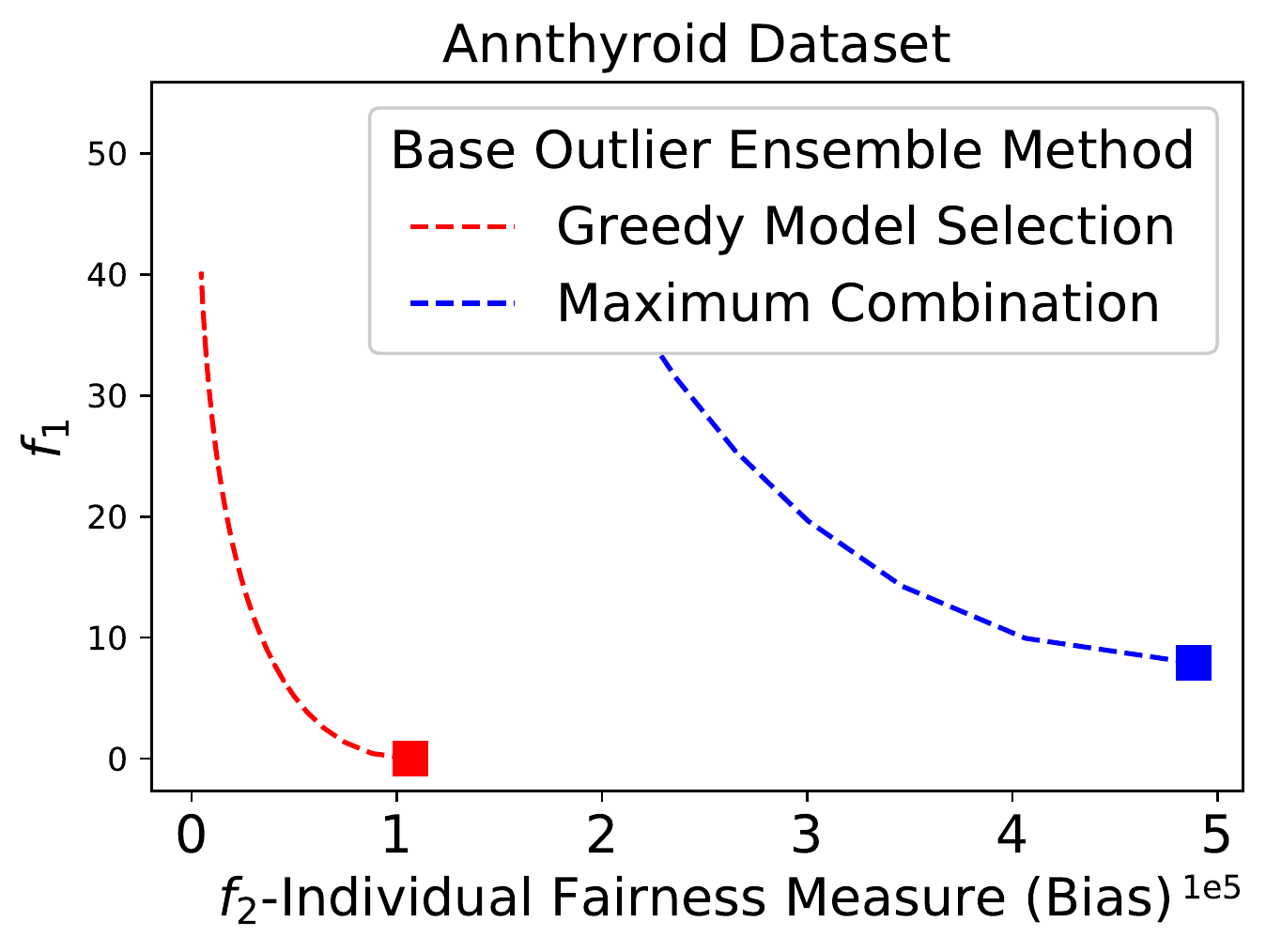}
    }
    \subfigure[Cardiotocogrpahy Dataset.]{
        \includegraphics[height=0.17\textwidth,width=0.23\textwidth]{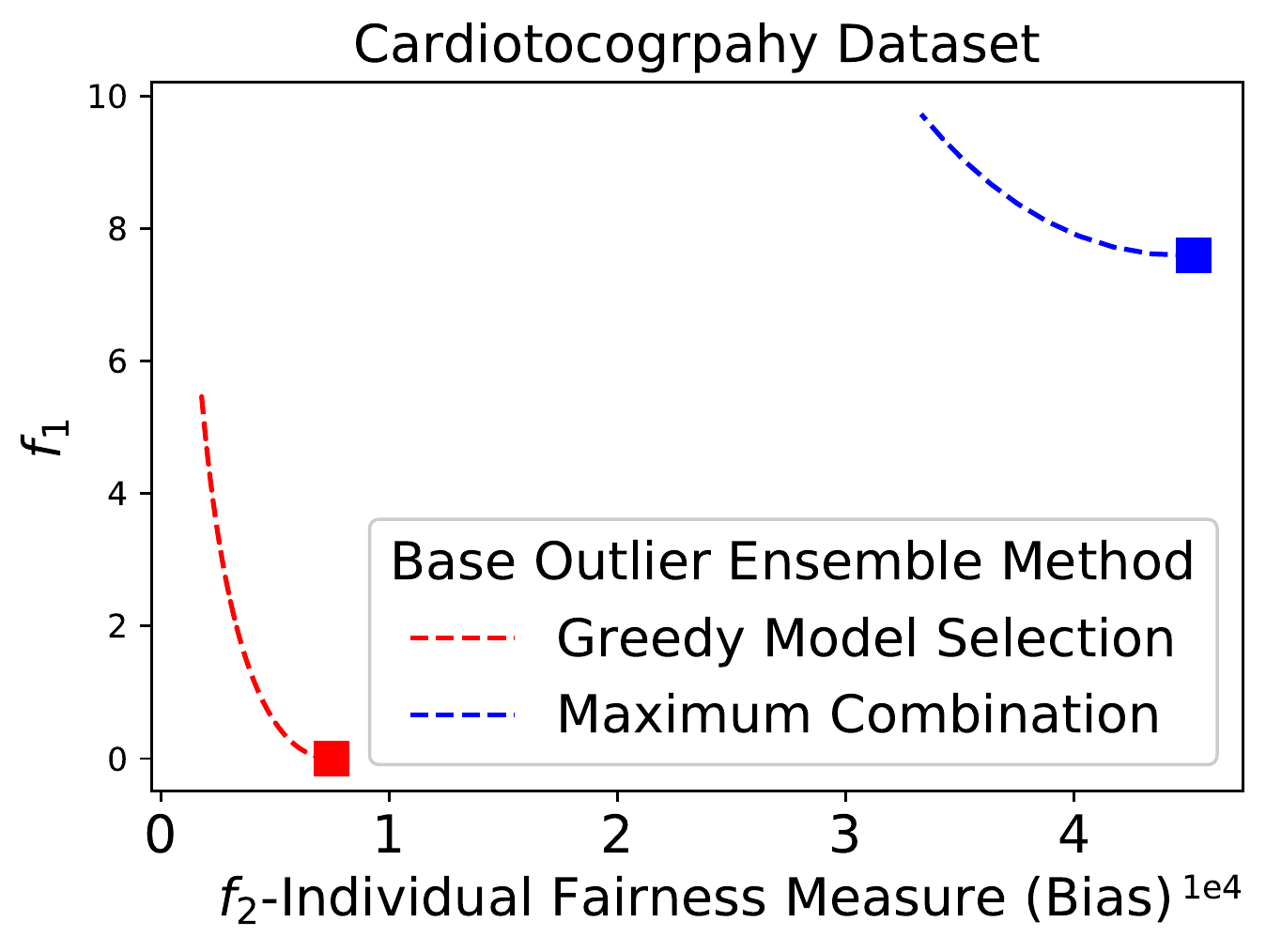}
    }
    \caption{$f_2$-$f_1$ curves for individual fairness.}
    \label{fig:f1f2individualfairness}
\end{figure*}

\subsubsection{Base Detectors}
Base detectors generate $\mathbf{S}$ for the following ensemble operation. 
In the experiments, we utilize three different kinds of outlier detection methods with various parameters as the base detectors.
\begin{itemize}[noitemsep,topsep=0pt]
\item \textbf{Local Outlier Factor}~\cite{breunig2000lof} chooses $5$, $10$, $15$, $20$, $25$, and $30$ as the number of neighbors.

\item \textbf{k-Nearest Neighbors}~\cite{ramaswamy2000efficient} sets $k$ as $2$, $4$, $6$, $8$, and $10$ respectively.

\item \textbf{Isolation Forest}~\cite{liu2008isolation} uses $25$, $50$, $75$, $100$, $125$, $150$, and $175$ as the number of base estimators.
\end{itemize}
The outputs from these base detectors will be normalized to $[0,1]$.
The closer the value to $1$, the more abnormality of the observation.
Finally, $\mathbf{S}$ is obtained from the results of these base detectors.

\subsubsection{Base Ensemble Methods}
Before running the fairness-aware ensemble process, we need to obtain the {target outlier vector} $\mathbf{t}$ as an optimal ensemble result.
Any existing outlier ensemble method can be used to generate the {target outlier vector}.
In this paper, we employ Average Combination and the Greedy Model Selection method~\cite{schubert2012evaluation} as examples to demonstrate the effectiveness of the proposed framework.

\begin{itemize}[noitemsep,topsep=0pt]
\item \textbf{Maximum Combination} calculates the maximum outlier score from all of base detectors for each instance, which is presented as $\{\frac{\max\mathbf{S}_{.,1}}{k}, \frac{\max\mathbf{S}_{.,2}}{k},...,\frac{\max\mathbf{S}_{.,n}}{k}\}$.
The maximum combination can find out the outliers, which are ``difficult'' to be identified, and thus, may improve the detection performance~\cite{aggarwal2015theoretical}.

\item \textbf{Greedy Model Selection}~\cite{schubert2012evaluation} is a diversity-based outlier ensemble method. 
It heuristically analyzes the correlation between the outputs of the base detectors with the ensemble results.
The base detector with the highest marginal gain on the diversity will be greedily adopted.
The final result is the average combination for all of the selected base detectors.
\end{itemize}

\begin{figure*}[!htb]
    \centering
     \subfigure[Communities Dataset.]{
        \includegraphics[height=0.17\textwidth,width=0.23\textwidth]{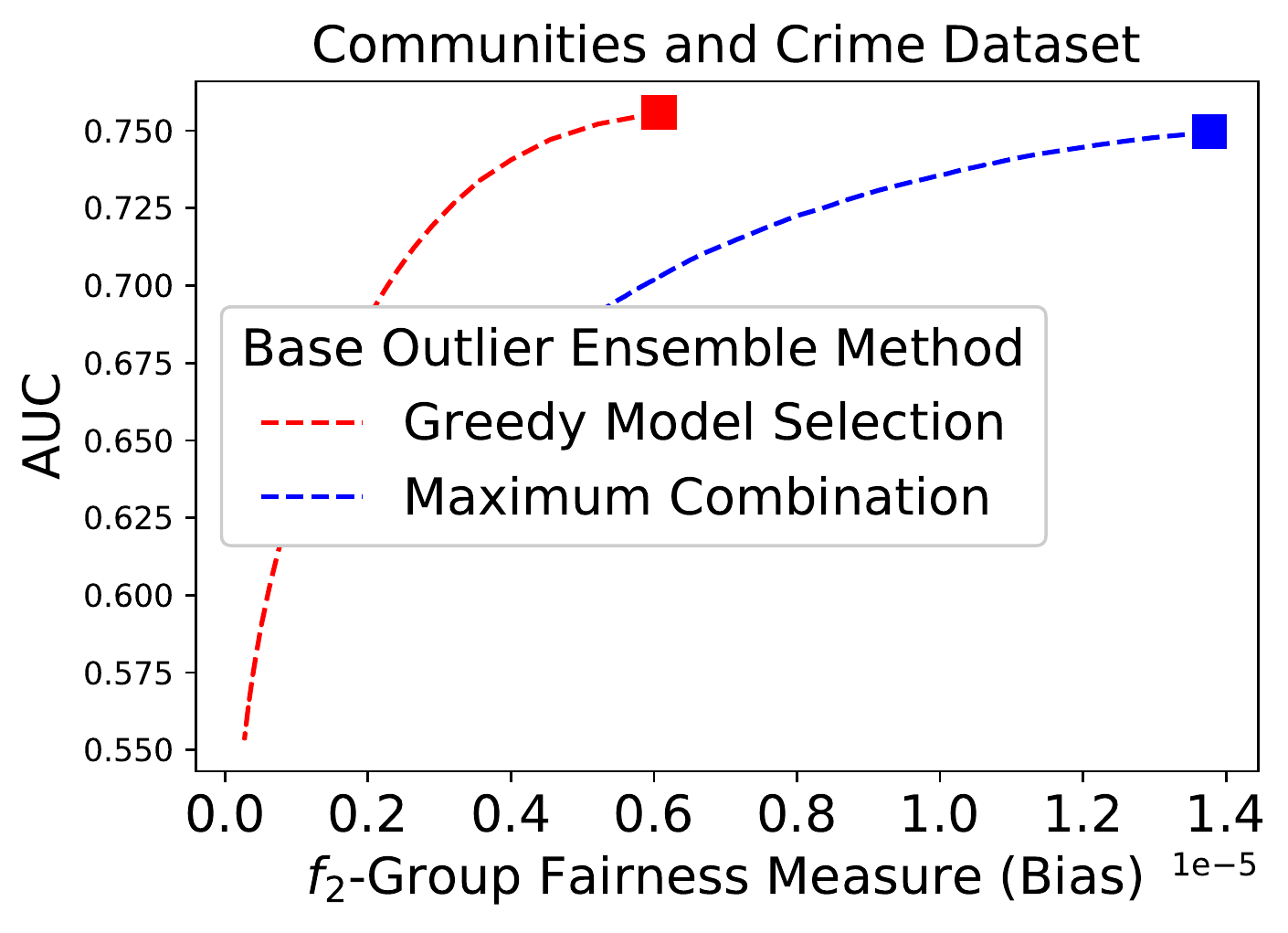}
    }
    \subfigure[{German Credit Dataset.}]{
    \label{fig:german_bias_AUC_group_fairness}
        \includegraphics[height=0.17\textwidth,width=0.23\textwidth]{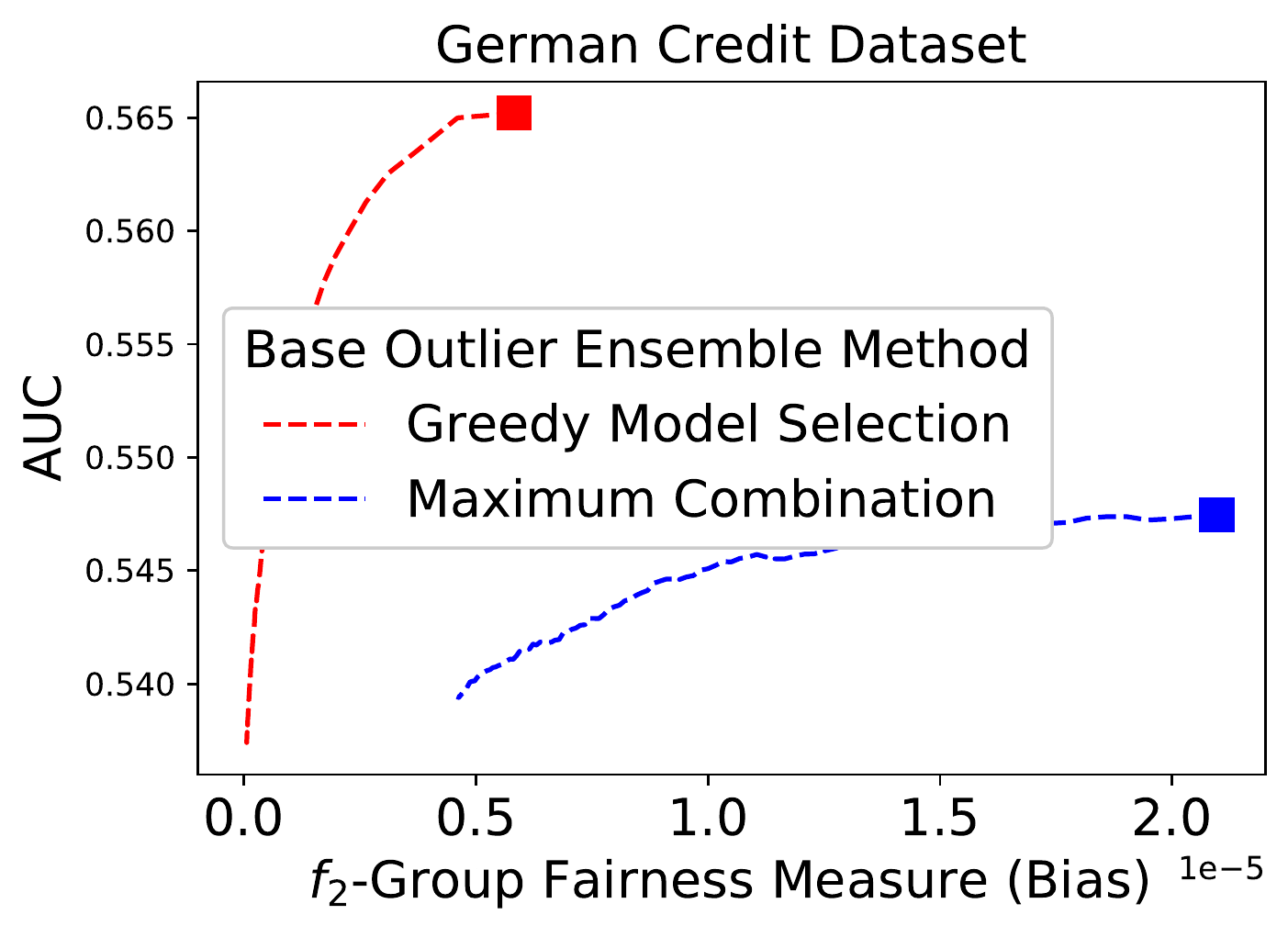}
    }    
    \subfigure[Annthyroid Dataset.]{
        \includegraphics[height=0.17\textwidth,width=0.23\textwidth]{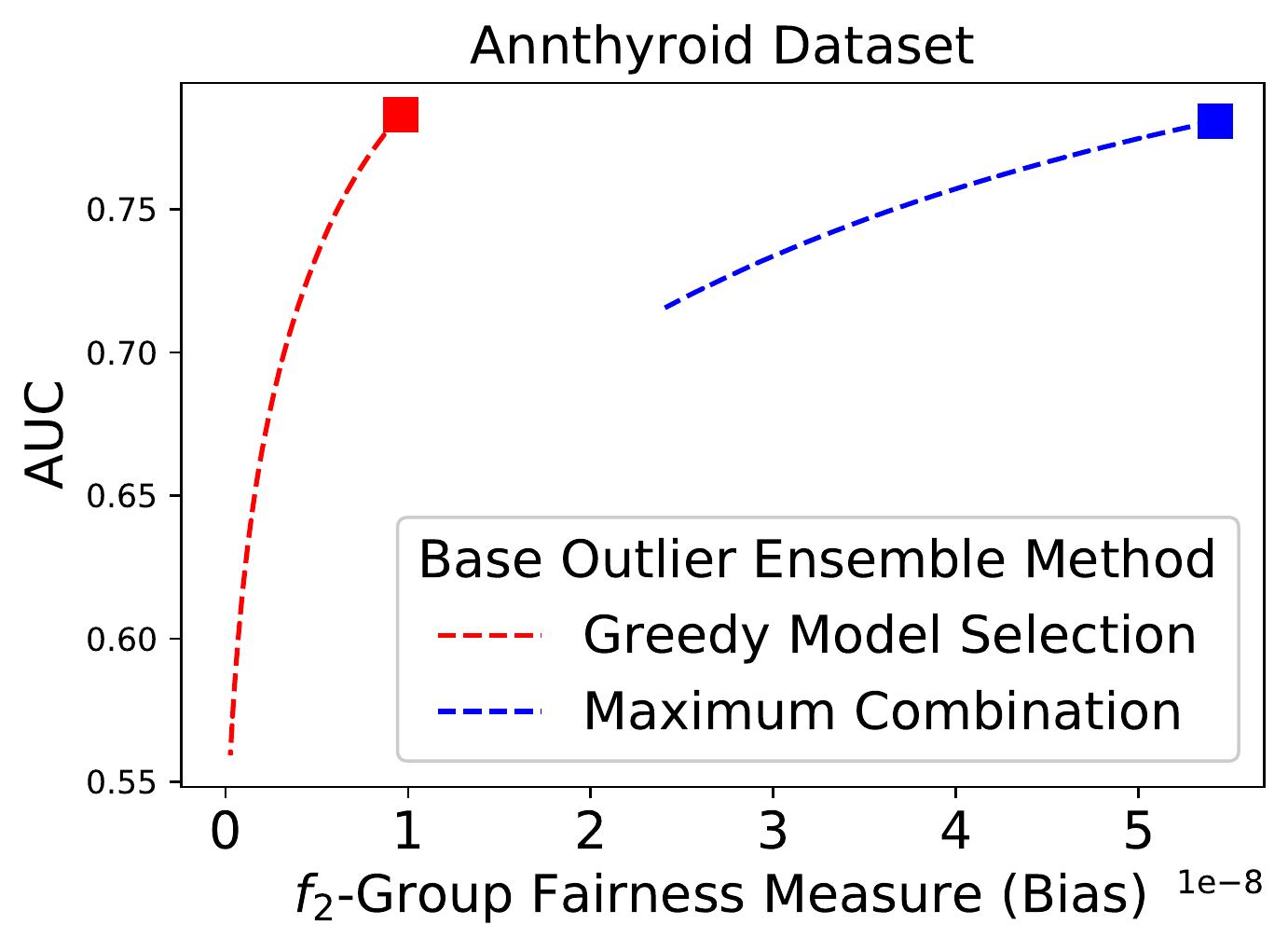}
    }
    \subfigure[Cardiotocogrpahy Dataset.]{
        \includegraphics[height=0.17\textwidth,width=0.23\textwidth]{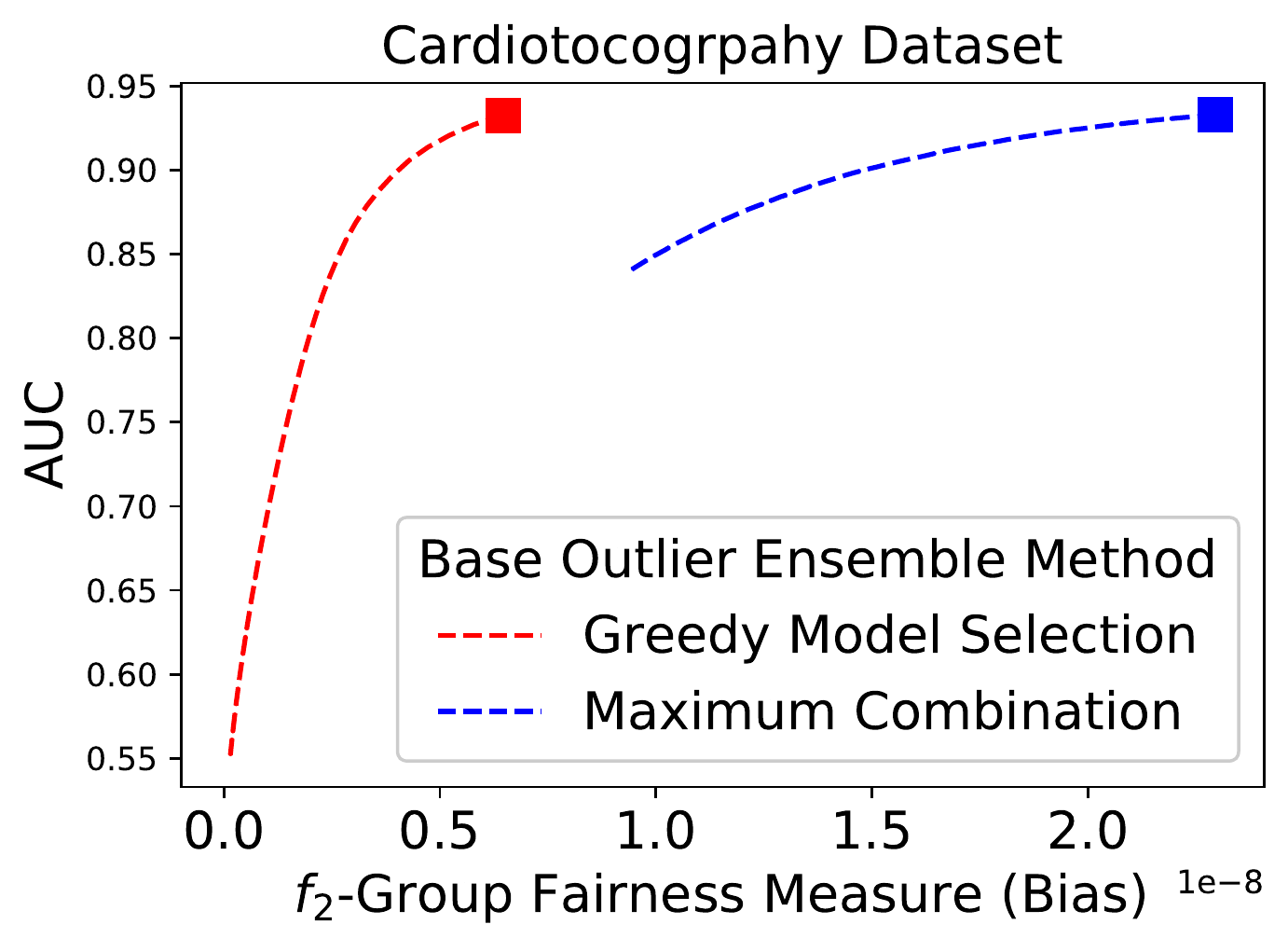}
    }
    \caption{Bias-AUC curves for group fairness.}
    \label{fig:barelation1}
\end{figure*}

\begin{figure*}[!h]
    \centering
    \subfigure[Communities Dataset.]{
        \includegraphics[height=0.17\textwidth,width=0.23\textwidth]{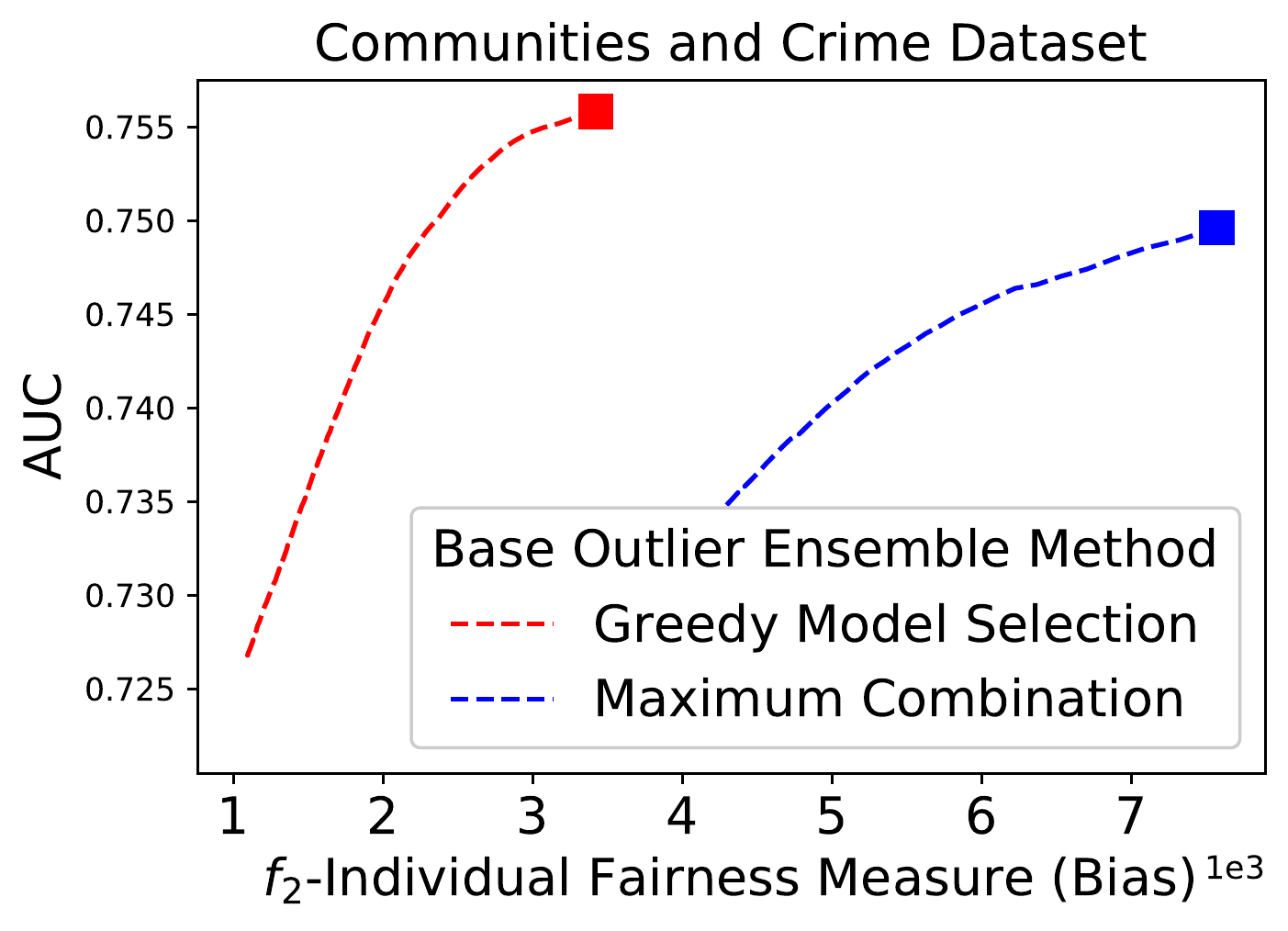}
    }
    \subfigure[{German Credit Dataset.}]{
        \includegraphics[height=0.17\textwidth,width=0.23\textwidth]{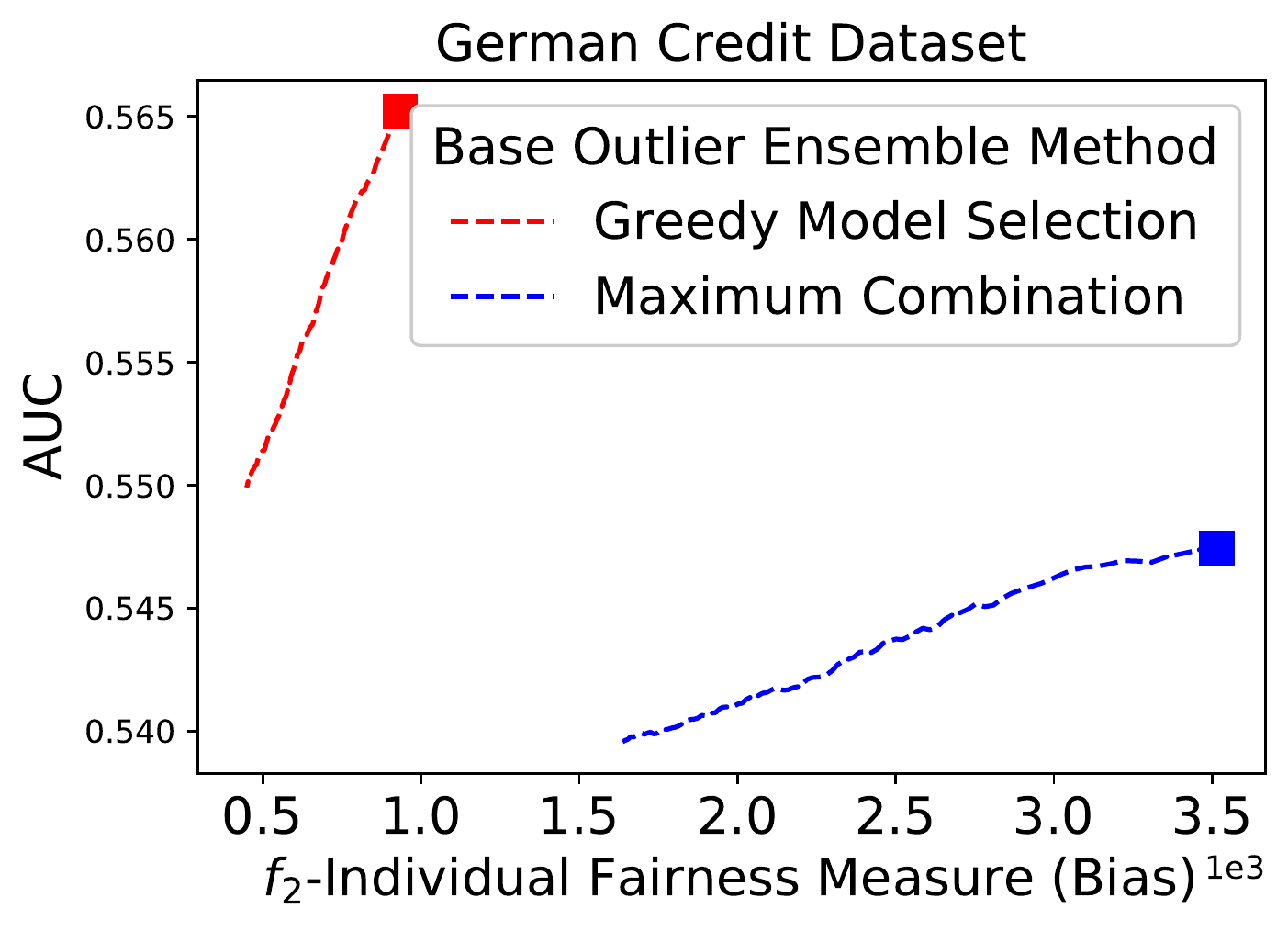}
    }    
    \subfigure[Annthyroid Dataset.]{
        \includegraphics[height=0.17\textwidth,width=0.23\textwidth]{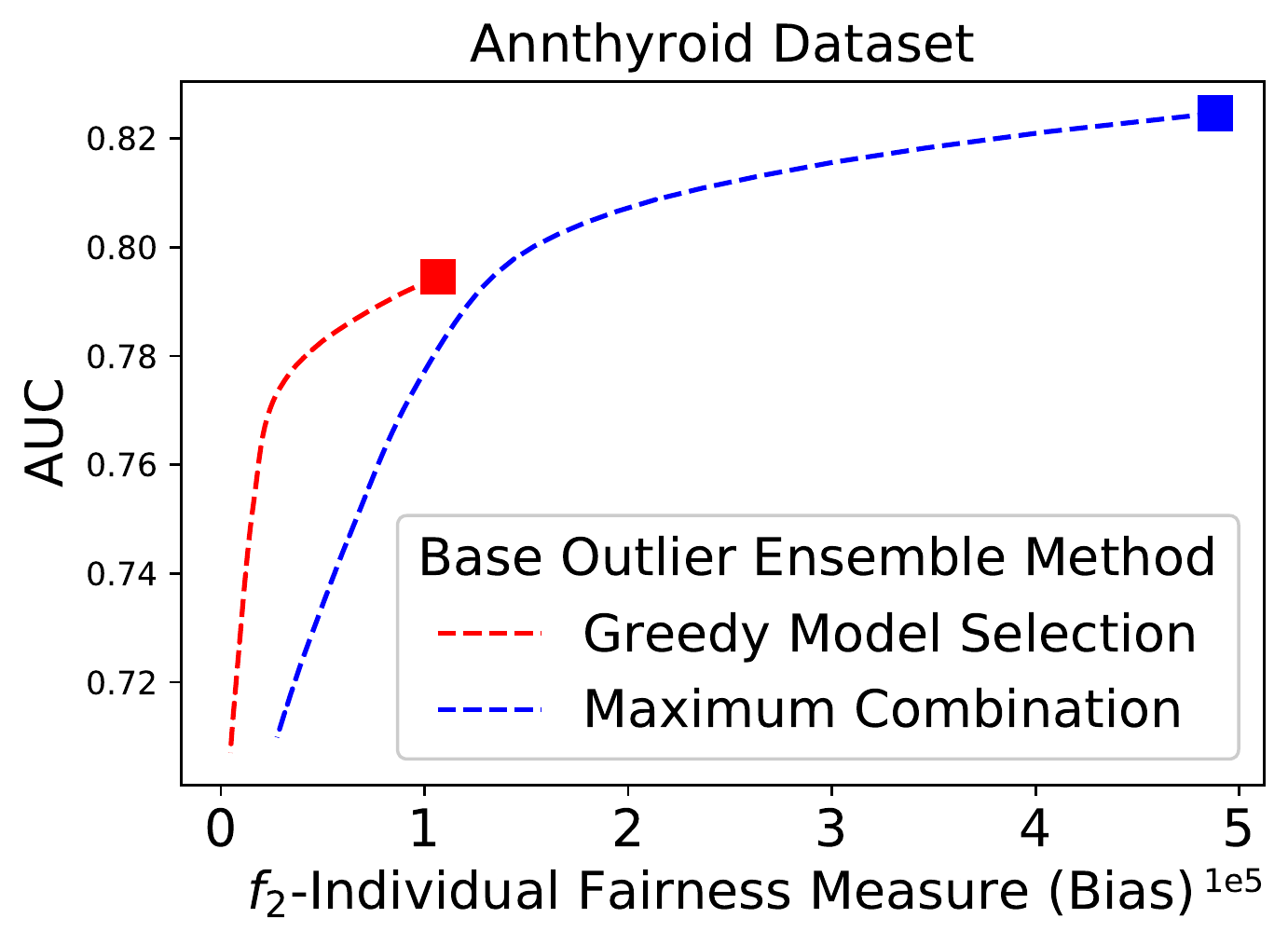}
    }
    \subfigure[Cardiotocogrpahy Dataset.]{
        \includegraphics[height=0.17\textwidth,width=0.23\textwidth]{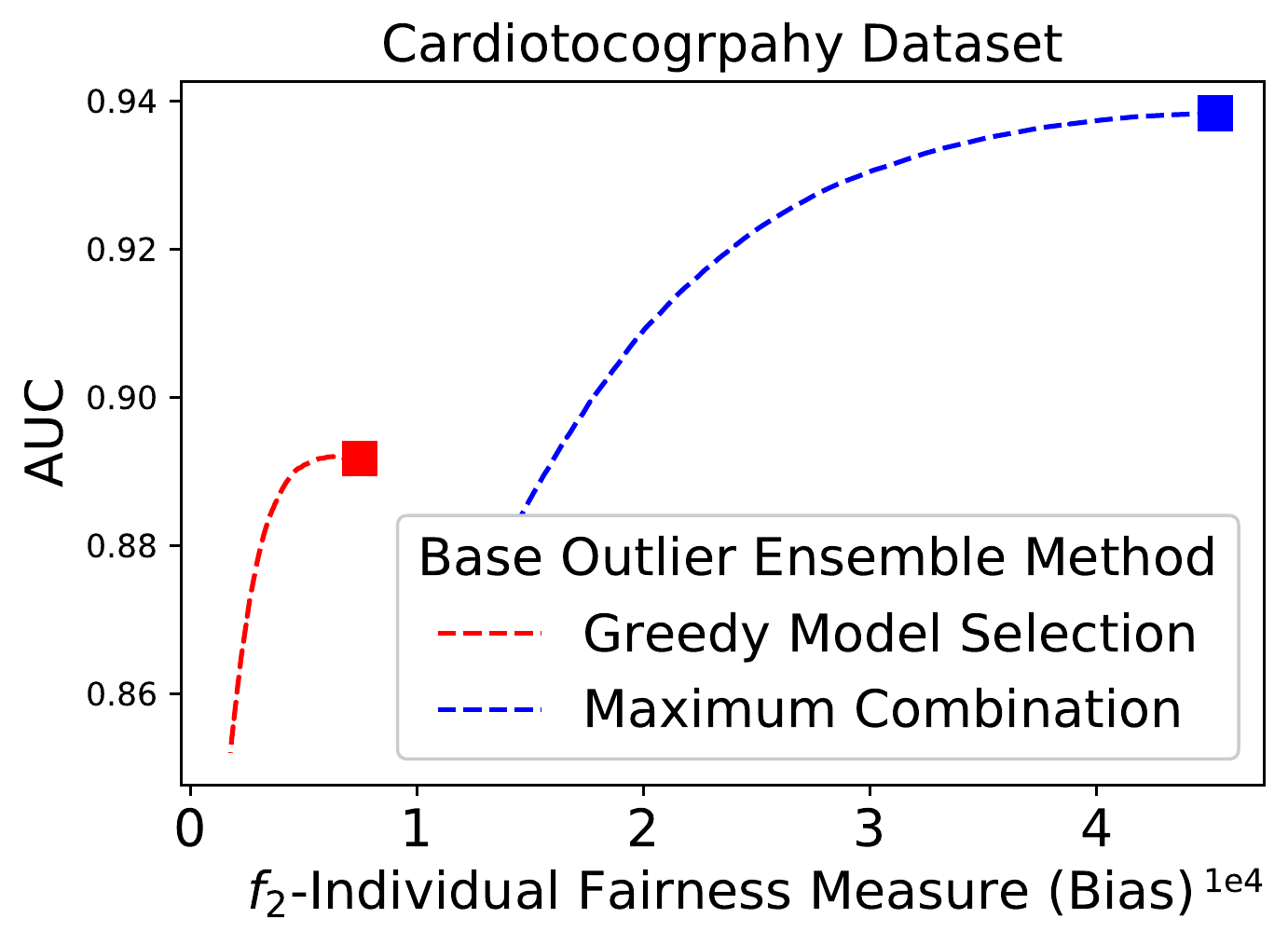}
    }
    \caption{Bias-AUC curves for individual fairness.}
    \label{fig:barelation2}
\end{figure*}

\subsubsection{Method for Comparison}
The introduced $f_1$ function in Sec. \ref{sec:Methods} has higher weights for those instances having higher outlier scores in the target outlier vector.
To verify the effectiveness of this weighted mechanism, we introduce a method without $\beta$ to make a comparison.
We name this method as \emph{unweighted $f_1$ framework}.
The objective function can be written as:
\begin{flalign*}
\underset{\mathbf{W}}{\text{minimize}} \;\sum_{i=1}^{n}(\mathbf{W} \cdot \mathbf{S}_{.,i} - T_i)^2 + \alpha f_2(\mathbf{W},\mathbf{S},{(X_i,a_i)}_{i=1}^n).
\end{flalign*}
Similarly, as the objective function is convex, we obtain the solutions by calculating the derivatives under the definitions of group fairness and individual fairness respectively.
Using DP as the group fairness measure, we have:
\begin{flalign*}
\mathbf{W} = 
(\mathbf{S}^\intercal \mathbf{S} + \frac{\alpha}{N} {\underset{p,q \in \mathbf{g},p\neq q}{\sum} \mathbf{d}_{pq}\mathbf{d}_{pq}^\intercal})^{-1}(\mathbf{S}^\intercal \mathbf{t}).
\end{flalign*}
Using IF as the individual fairness, we then have:
\begin{flalign*}
&\mathbf{W} = 
(\mathbf{S}^\intercal \mathbf{S} + &\\\nonumber
&\frac{\alpha}{N}{\underset{p,q \in \mathbf{g},p\neq q}{\sum}\frac{(\mathbf{M}_{pq} \odot \mathbf{D}_{pq}^\intercal)\mathbf{D}_{pq}\mathbf{W}}{|\{i|a_i = p\}||\{j|a_j = q\}|}})^{-1}(\mathbf{S}^\intercal \mathbf{t}).
\end{flalign*}
Intuitively, without distinguishing the importance among different observations, when $\mathbf{t}$ is the optimal result, the outlier detection performance of the unweighted $f_1$ framework would be worse than that of the proposed fairness-aware outlier ensemble framework in terms of the AUC value.
The evaluations will be given based on the following experiments.

\subsection{Experimental Results}
Leveraging the aforementioned base outlier detectors and base ensemble methods, we propose to answer \textbf{Q1}-\textbf{Q3}, respectively.
\begin{itemize}[noitemsep,topsep=0pt]
    \item \textbf{A1:} For \textbf{Q1}, we evaluate the relation between $f_1$ and $f_2$ to verify the effectiveness of the framework.
    \item \textbf{A2:} For \textbf{Q2}, we analyze the bias-AUC curve to illustrate the change of performance during the improvement of the fairness.
    \item \textbf{A3:} For \textbf{Q3}, we compare the proposed framework with the unweighted $f_1$ framework to demonstrate the superior performance of the proposed $f_1$ function in terms of cost of fairness.
\end{itemize}

\begin{figure*}[!h]
    \centering
    \subfigure[Communities Dataset.]{
        \includegraphics[height=0.17\textwidth,width=0.23\textwidth]{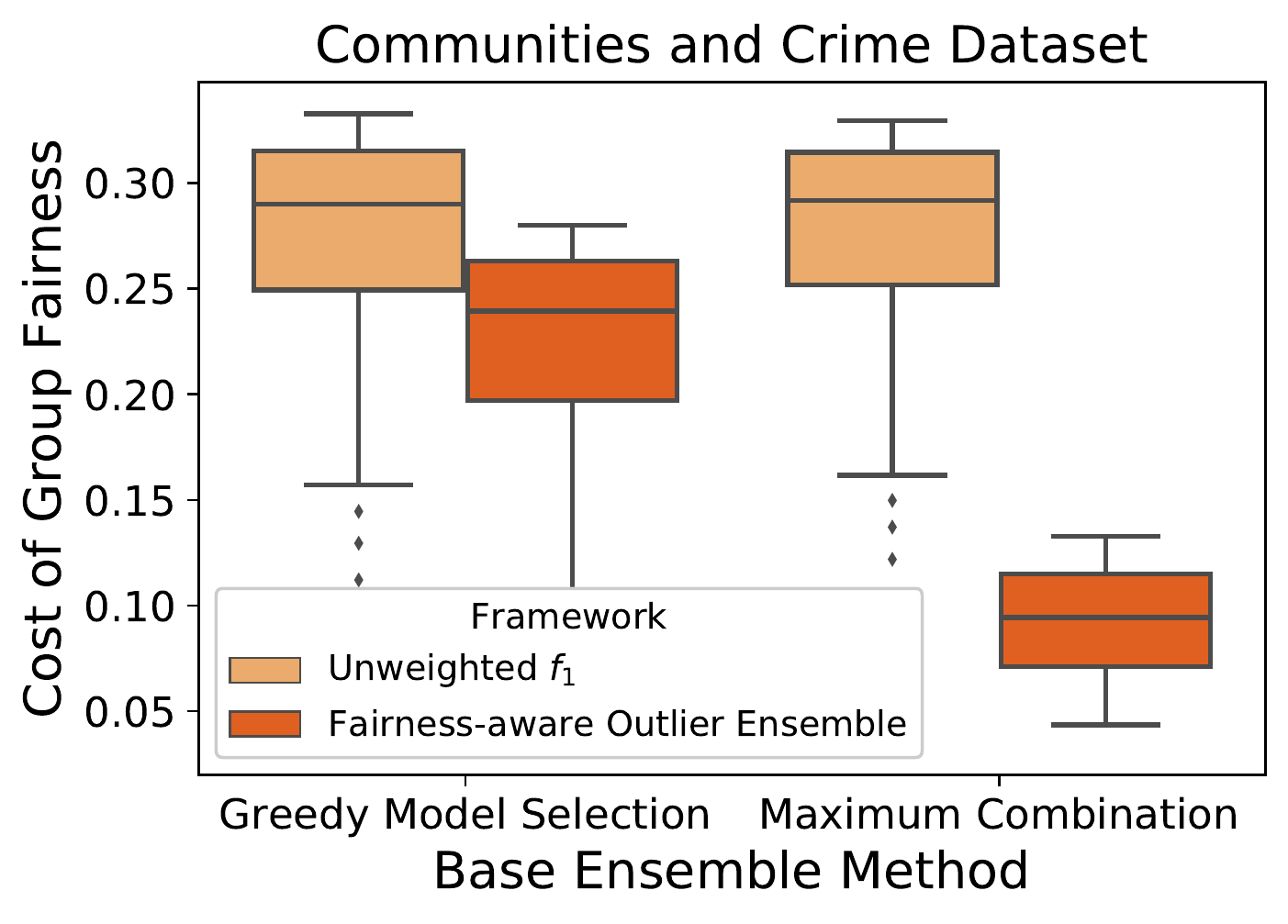}
    }
    \subfigure[{German Credit Dataset.}]{
        \includegraphics[height=0.17\textwidth,width=0.23\textwidth]{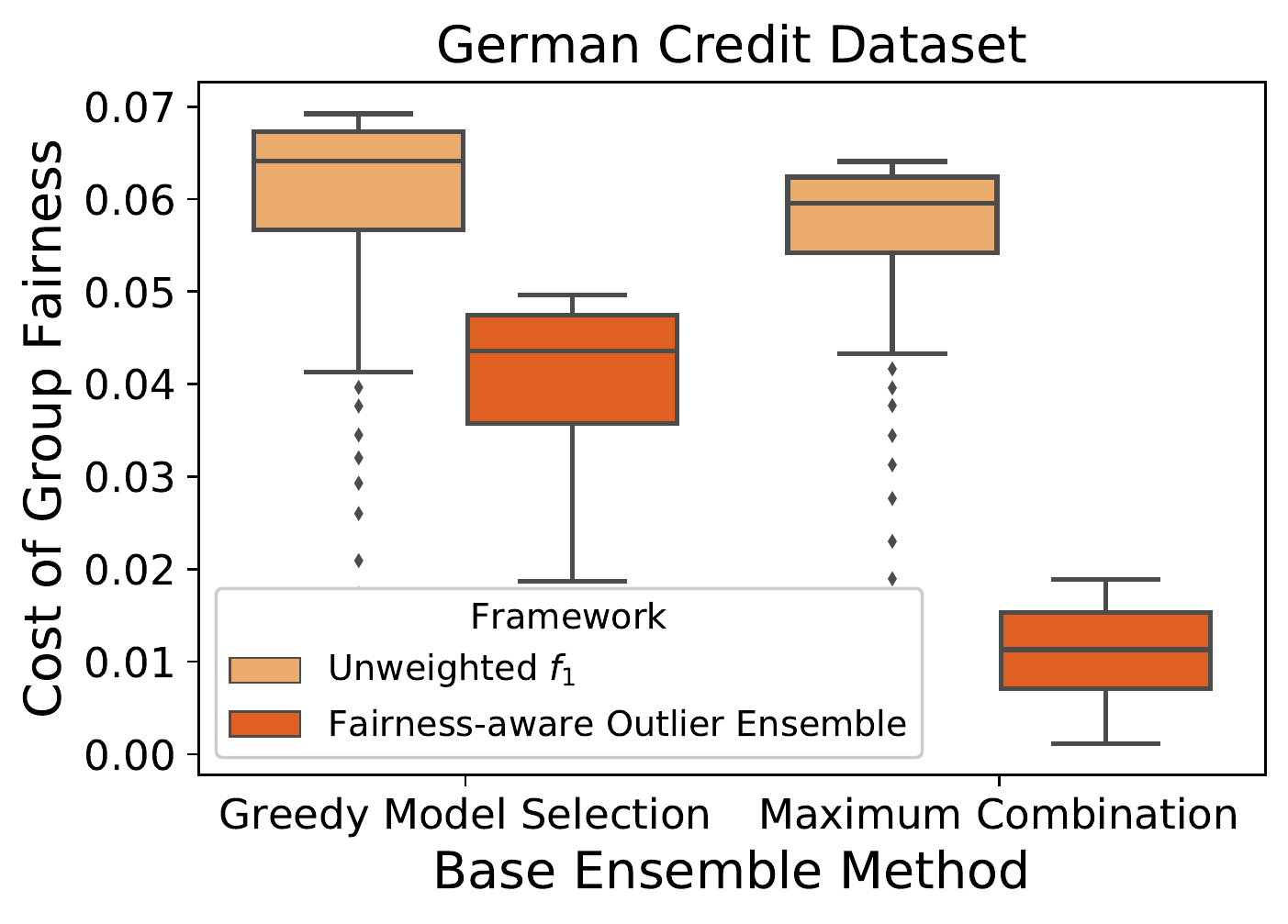}
    }    
    \subfigure[Annthyroid Dataset.]{
        \includegraphics[height=0.17\textwidth,width=0.23\textwidth]{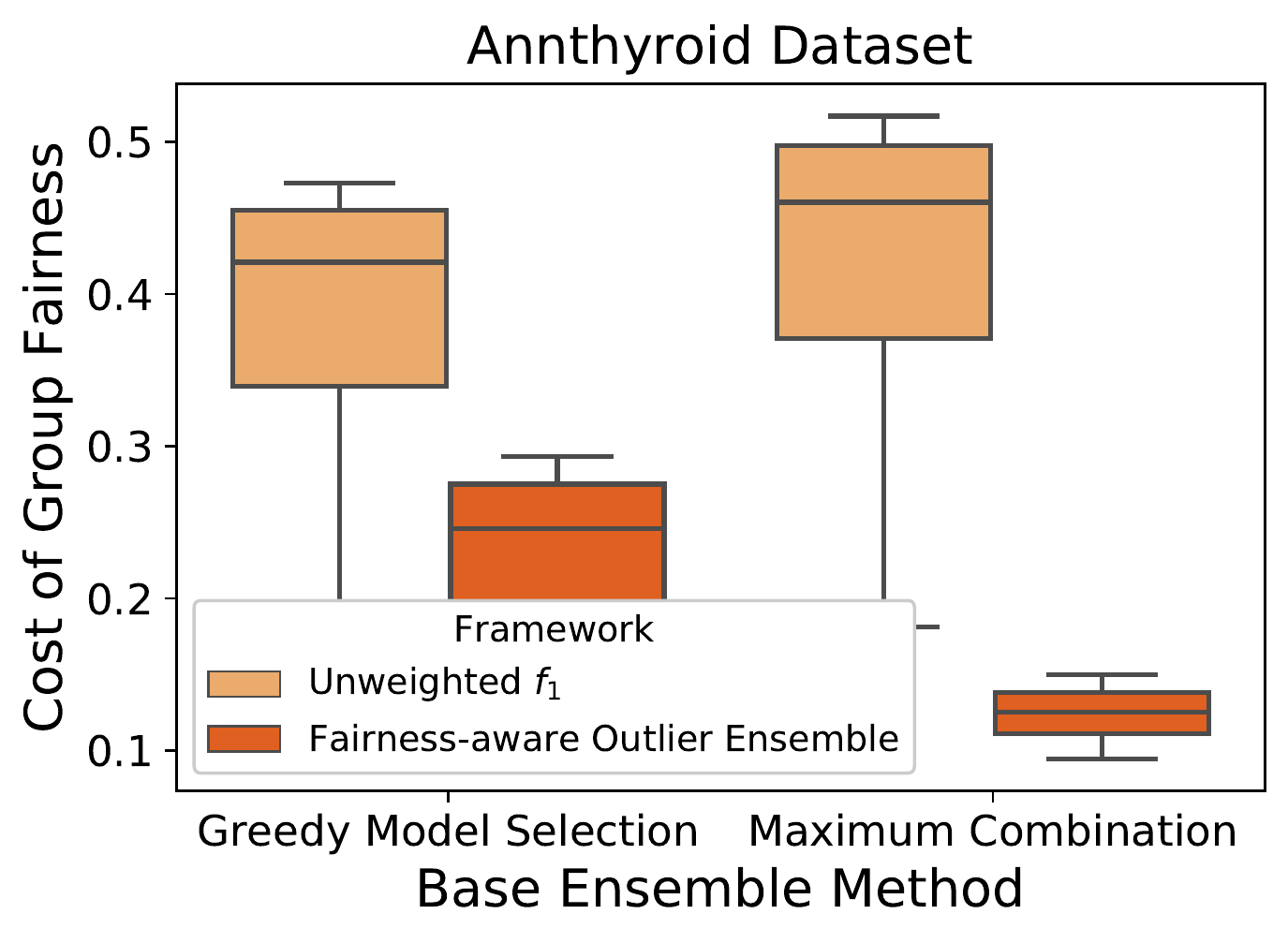}
    }
    \subfigure[Cardiotocogrpahy Dataset.]{
        \includegraphics[height=0.17\textwidth,width=0.23\textwidth]{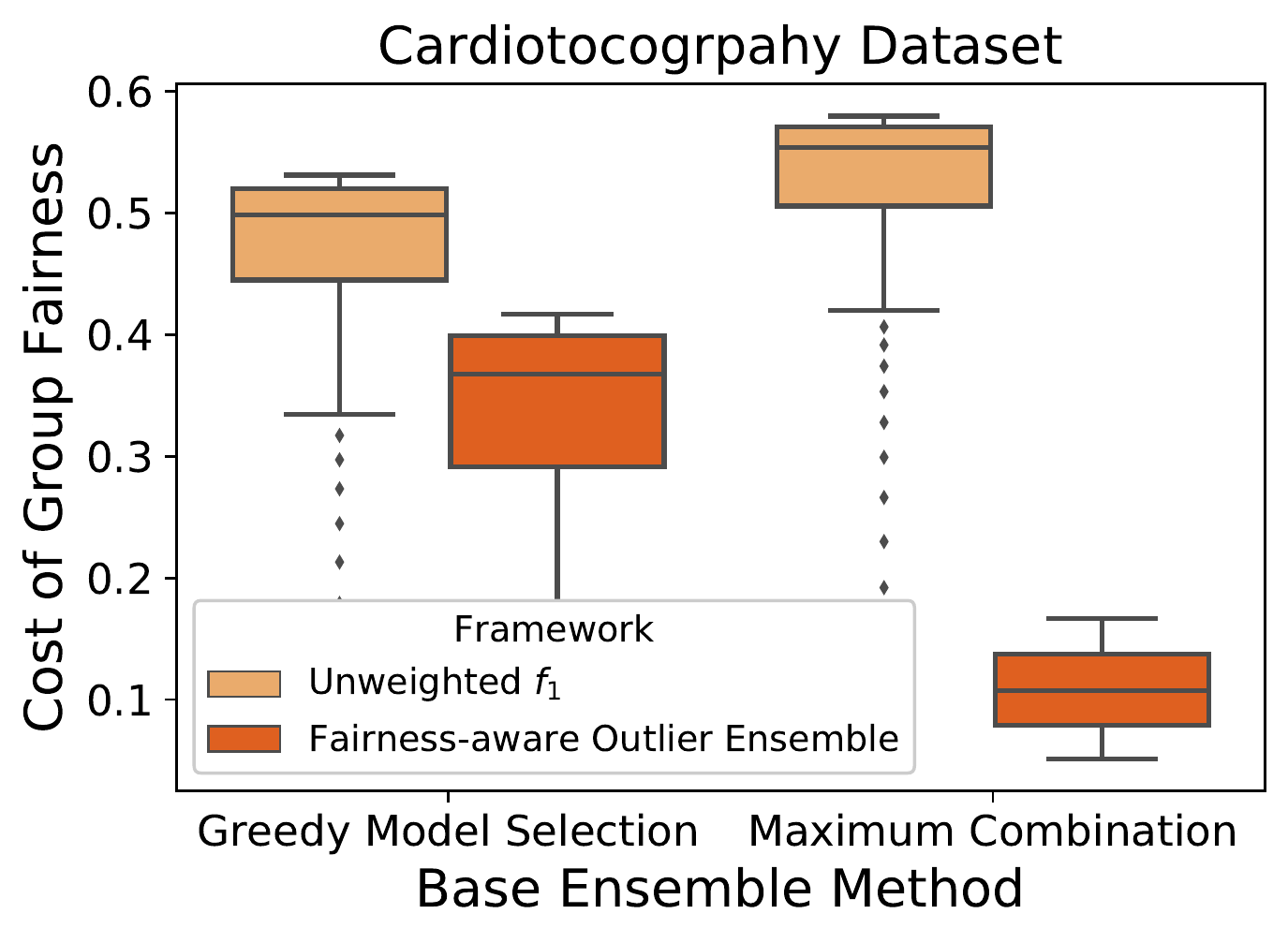}
    }
    \caption{Cost of group fairness.}
    \label{fig:cogf}
\end{figure*}

\begin{figure*}[!h]
    \centering
    
    \subfigure[Communities Dataset.]{
        \includegraphics[height=0.17\textwidth,width=0.23\textwidth]{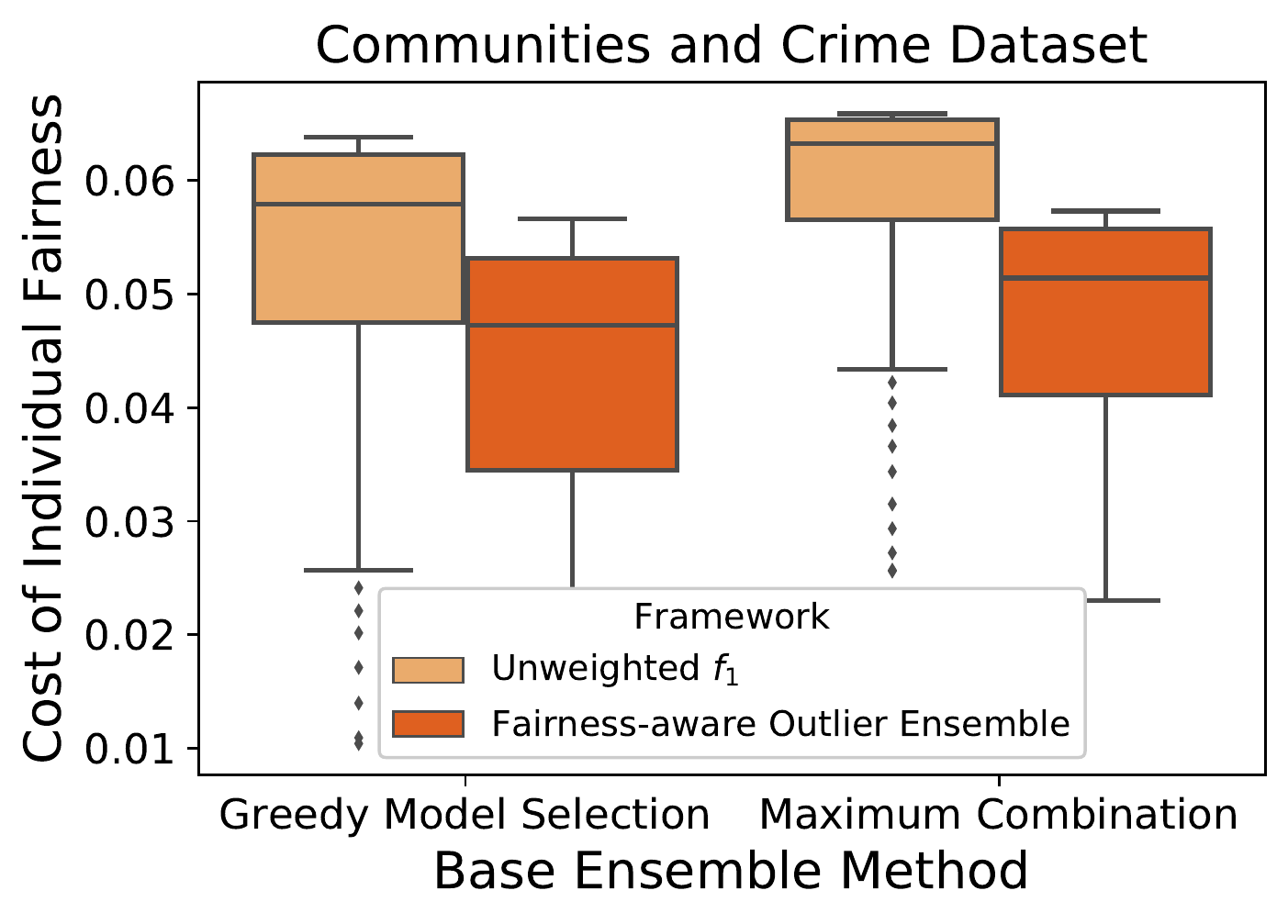}
    }
    \subfigure[{German Credit Dataset.}]{
        \includegraphics[height=0.17\textwidth,width=0.23\textwidth]{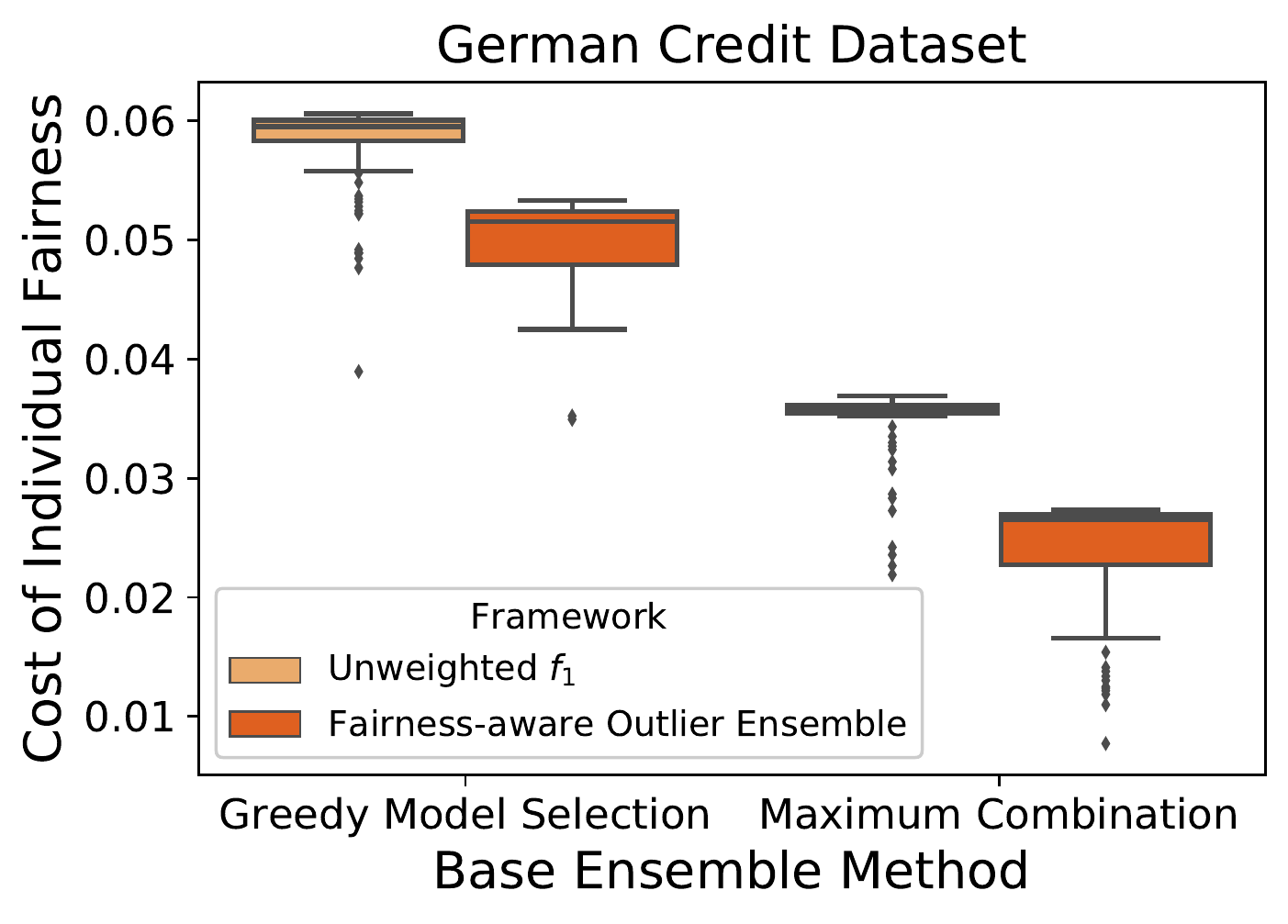}
    }    
    \subfigure[Annthyroid Dataset.]{
        \includegraphics[height=0.17\textwidth,width=0.23\textwidth]{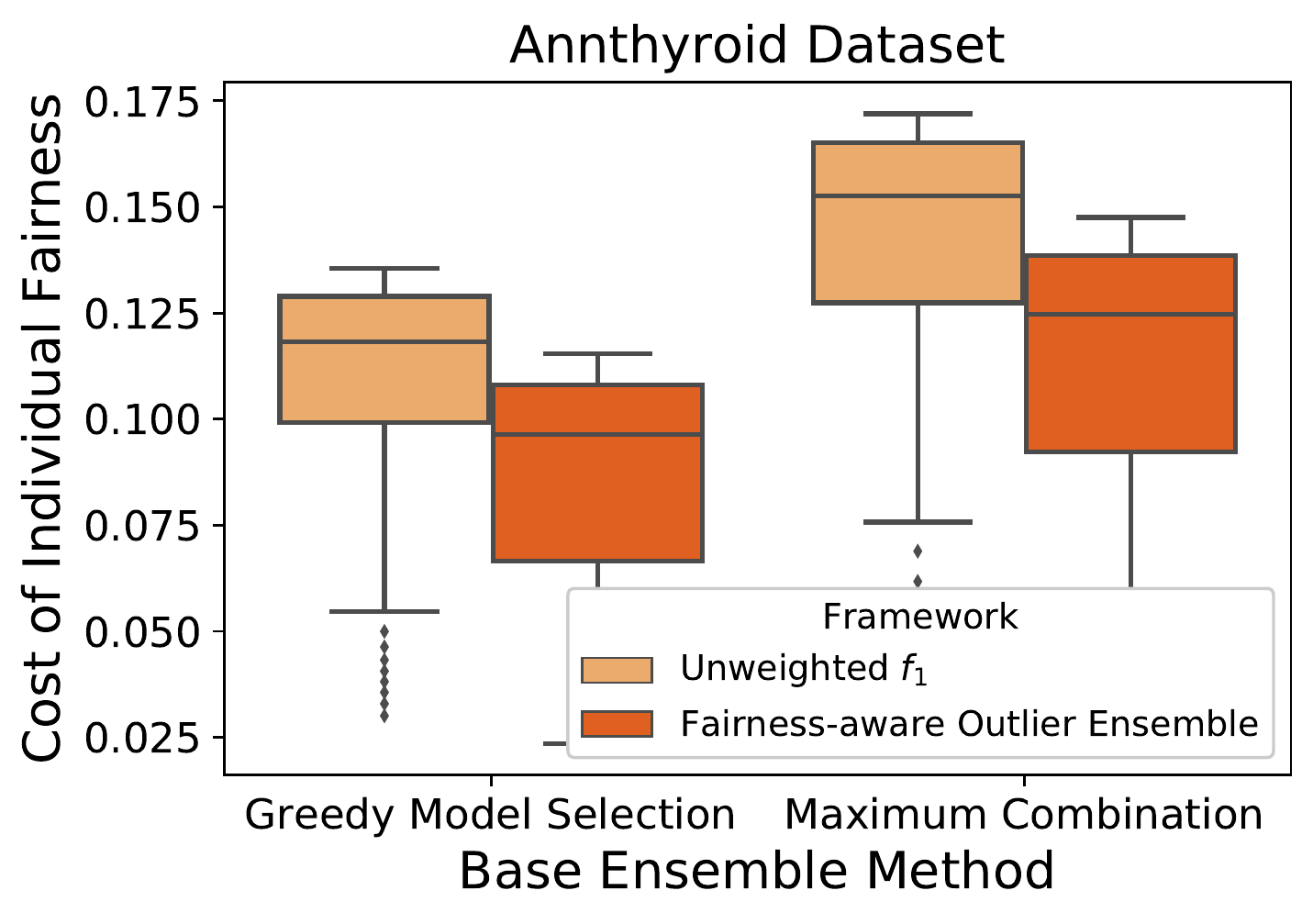}
    }
    \subfigure[Cardiotocogrpahy Dataset.]{
        \includegraphics[height=0.17\textwidth,width=0.23\textwidth]{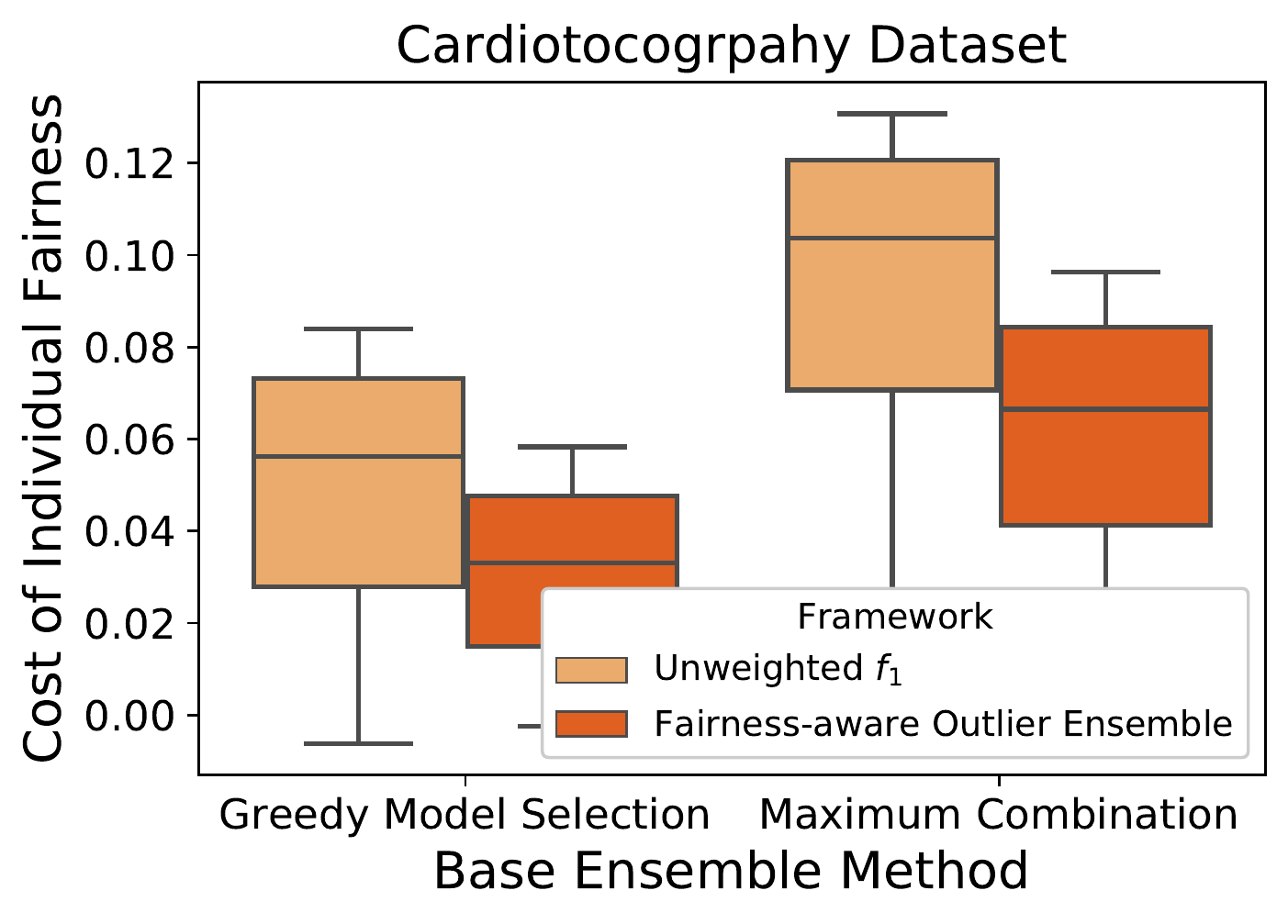}
    }
    \caption{Cost of individual fairness.}
    \label{fig:coif}
\end{figure*}
\subsubsection{\textbf{A1}} \emph{Relations between $f_1$ and $f_2$}.
Employing the trade-off parameter $\alpha \in [0,+\infty)$, the relative importance of $f_1$ and $f_2$ can be changed.
The larger the $\alpha$, the lower the $f_2$ can be obtained.
During the experiments, we select multiple $\alpha$ in an ascending order to generate different $\mathbf{W}$ and the corresponding values of $f_1$ and $f_2$.
We plot the $f_2$-$f_1$ curves for group fairness and individual fairness in Fig. \ref{fig:f1f2groupfairness} and Fig. \ref{fig:f1f2individualfairness} respectively.
Also, we present $f_1$ v.s. $f_2$ for both base outlier ensemble methods.
The \emph{square marker} denotes values where $\alpha$ is the smallest.
Note that, our definitions of group fairness and individual fairness indicate the bias of the results.
The results of Fig. \ref{fig:f1f2groupfairness} and Fig. \ref{fig:f1f2individualfairness} show that regardless of base ensemble methods, when bias is reducing, the difference between the newly generated outlier vector and the target outlier vector will increase.
Furthermore, when $\alpha$ is the smallest, the ensemble results have the largest bias and the lowest $f_1$ value.

\subsubsection{\textbf{A2}} \emph{Relations between AUC and bias}. 
The bias-AUC curves for group fairness and individual fairness are presented in Fig.~\ref{fig:barelation1} and Fig.~\ref{fig:barelation2}, respectively.
Intuitively, if the target outlier vector $\mathbf{t}$ is an optimal detection result, then even a small difference with $\mathbf{t}$ may degrade the AUC.
Therefore, the minimization operation on $f_1$ function is expected to help maintain a \emph{similar} detection performance with the base ensemble method.
From Fig.~\ref{fig:barelation1} and Fig.~\ref{fig:barelation2}, the experiments show that when the bias is decreasing, AUC value is always degraded simultaneously.
Therefore, a \emph{cost} on AUC for improving the fairness of the outlier ensemble result obviously exists.
We investigate this cost and compare the proposed framework with the unweighted $f_1$ framework in the next subsection.

\subsubsection{\textbf{A3}} \emph{Cost of fairness}.
We use the fairness improvement per AUC degradation to measure the cost of fairness (cof):
\begin{equation*}
\text{cof} = \frac{f_2(\mathbf{W}_{\alpha=0}) - f_2(\mathbf{W})}{\text{AUC}(\mathbf{W}_{\alpha=0})-\text{AUC}(\mathbf{W})}
\end{equation*}
Here, $\mathbf{W}_{\alpha=0}$ denotes the obtained $\mathbf{W}$ when $\alpha$ is $0$.
AUC(.) represents the AUC value for the obtained $\mathbf{y}$ derived by using the corresponding $\mathbf{W}$ as the ensemble weights.
We randomly sample $100$ values of $\alpha$ for each dataset and obtain different $\mathbf{W}$ through two frameworks.
Based on the produced $\mathbf{W}$, the corresponding cof values are calculated and presented in boxplots showing their distributions.
The results for group and individual fairness are shown in Fig.~\ref{fig:cogf} and Fig.~\ref{fig:coif} respectively. Clearly, a lower cof can be obtained by employing the proposed fairness-aware outlier ensemble framework, while the unweighted $f_1$ framework usually have a higher cof value. Thus, the effectiveness of $\beta$ in $f_1$ can be experimentally verified.

\section{Related Works}
\label{sec:related works}

Since the lack of the prior works in fair outlier detection, in this section, we first provide a review on the outlier ensemble, then we introduce fair machine learning methods in three different categories.
\subsection{Outlier Ensemble}
Ensemble analysis for outlier detection is an challenging and emerging area~\cite{aggarwal2013outlier,schubert2012evaluation,zimek2014ensembles}.
Aggarwal \emph{et~al.} \cite{aggarwal2015theoretical} provided the theoretical foundations for methods such as subsampling, bagging, averaging/maximum combination in outlier ensemble, which mostly aim to reduce the variance or induce the diversity.
Similarly, some other important works designed combination mechanisms on all of the base models to directly reduce the variance~\cite{gao2006converting,kriegel2011interpreting,liu2008isolation}.
Another strategy is to select base models to optimize the diversity of the outlier scores, which can also improve the detection performance~\cite{campos2018unsupervised,rayana2016less,schubert2012evaluation,zimek2014ensembles} .
However, none of the previous outlier ensemble structures addressed the fairness issue.

\subsection{Fairness-aware Machine Learning Methods}
There are three categories of fairness-aware machine learning methods, which are \emph{pre-processing}, \emph{in-processing} and \emph{post-processing} methods.
First, pre-processing methods mitigate bias by transforming original datasets, and will not change the process of the machine learning algorithms~\cite{kamiran2012data,calmon2017optimized,sharma2020data}.
In-processing methods directly transform the machine learning method to remove discrimination during the training process, they usually formulate an optimization problem to trade off between fairness and performance~\cite{kamishima2012fairness,zafar2015fairness,berk2017convex}.
For the post-processing, original machine learning methods are usually utilized as black-box methods and will not be modified. 
The outputs will be directly modified to fit the fairness requirements while maintaining an acceptable performance~\cite{hardt2016equality,kim2019multiaccuracy}.
As for the fair outlier detection, P \emph{et~al.} \cite{p2020fair} proposed FairLOF to reformulate LOF~\cite{breunig2000lof} and Davidson \emph{et~al.} focused on a special fairness measure which is further used to detect unfairness~\cite{Davidson2019AFF}. Both of these two works are very different with the task we solved.

\section{Conclusions}
\label{sec:Conclusion}
In this paper, we present a fairness-aware outlier ensemble framework to reduce the bias of the outlier ensemble results while preventing a big drop in its AUC measure. 
By employing any existing outlier ensemble method, the proposed framework has a stacking structure and transforms the original outlier ensemble result into a fairer result in a post-processing manner. 
To measure the bias of the outlier ensemble results under the unsupervised setting, we define two fairness measures for group fairness and individual fairness respectively.
Further, we propose an indirect trade-off formulation to control the loss in the detection performance with the improvement of the fairness.
The effectiveness of the proposed framework is verified through experiments on a variety of public datasets.
We also designed a comparative method to illustrate the superior performance of the proposed framework in terms of the loss in detection performance with improved fairness.

\bibliographystyle{siamplain}
\bibliography{fairness}

\appendix
\section{Supplementary Materials}
We provide detail demonstrations of solutions and the experimental results on other four datasets in the supplementary materials.

\subsection{Solutions}
The definitions of $f_1$ and $f_2$ functions are both convex, we derive the closed-form solutions for two definitions of fairness.

\subsubsection{Solution for Group Fairness}
Replacing $f_2$ by group fairness measurement DP, the objective function under group fairness is as follows:
\begin{flalign*}
&\underset{\mathbf{W}}{\text{minimize}} \;\text{L}_1(\mathbf{W}) = \sum_{i=1}^{n}\beta_i(\mathbf{W} \cdot \mathbf{S}_{.,i} - T_i)^2 +\\\nonumber
&\frac{\alpha}{N}{\underset{p,q \in \mathbf{a},p\neq q}{\sum}(\frac{\sum_{\{i|a_i=p\}}\mathbf{W} \cdot \mathbf{S}_{.,i}}{|\{i|a_i = p\}|} - \frac{\sum_{\{j|a_j=q\}}\mathbf{W} \cdot \mathbf{S}_{.,j}}{|\{j|a_i = q\}|})^2}.
\end{flalign*}
Using $\mathbf{d}_{pq}$ replace $\frac{\sum_{\{i|a_i=p\}} \mathbf{S}_{.,i}}{|\{i|a_i = p\}|} - \frac{\sum_{\{j|a_j=q\}} \mathbf{S}_{.,j}}{|\{j|a_j = q\}|}$, the objective function can be written as:
\begin{flalign*}
&\underset{\mathbf{W}}{\text{minimize}} \;\text{L}_1(\mathbf{W}) =&\\\nonumber
&\sum_{i=1}^{n}\beta_i(\mathbf{W} \cdot \mathbf{S}_{.,i} - T_i)^2 +\frac{\alpha}{N} {\underset{p,q \in \mathbf{g},p\neq q}{\sum}(\mathbf{W}\cdot \mathbf{d}_{pq})^2}.
\end{flalign*}
As the problem is convex, we then directly solve the problem by calculating derivative w.r.t. $\mathbf{W}$:
\begin{flalign*}
&\frac{d\text{L}_1}{d\mathbf{W}} = 2\mathbf{B}\odot\mathbf{S}^\intercal \mathbf{S} \mathbf{W} - 2\mathbf{B}\odot\mathbf{S}^\intercal \mathbf{t} +&\\\nonumber &\frac{2\alpha}{N} {\underset{p,q \in \mathbf{g},p\neq q}{\sum} \mathbf{d}_{pq}\mathbf{d}_{pq}^\intercal \mathbf{W}},
\end{flalign*}
where $\mathbf{B} \in \mathbb{R}^{k \times n}$ is the stack of $k$ $\beta$.
We obtain the solution of $\mathbf{W}$ when $\frac{d\text{L}_1}{d\mathbf{W}} = 0$:
\begin{flalign*}
&2\mathbf{B}\odot\mathbf{S}^\intercal \mathbf{S} \mathbf{W} - 2\mathbf{B}\odot\mathbf{S}^\intercal \mathbf{t} + \frac{2\alpha} {N}{\underset{p,q \in \mathbf{g},p\neq q}{\sum} \mathbf{d}_{pq}\mathbf{d}_{pq}^\intercal \mathbf{W}} = 0.&
\end{flalign*}
Thus:
\begin{flalign*}
&\mathbf{W} = 
(\mathbf{B}\odot\mathbf{S}^\intercal \mathbf{S} + \frac{\alpha}{N}{\underset{p,q \in \mathbf{g},p\neq q}{\sum} \mathbf{d}_{pq}\mathbf{d}_{pq}^\intercal})^{-1}(\mathbf{B}\odot\mathbf{S}^\intercal \mathbf{t}).&
\end{flalign*}

\subsubsection{Solution for Individual Fairness}
Inserting $\text{IF}$ as the $f_2$ function in the framework, the objective function under individual fairness is as follows:
\begin{flalign*}
&\underset{\mathbf{W}}{\text{minimize}} \;\text{L}_2(\mathbf{W}) = \sum_{i=1}^{n}\beta_i(\mathbf{W} \cdot \mathbf{S}_{.,i} - T_i)^2 +&\\\nonumber
&\frac{\alpha}{N}{\underset{p,q\in \mathbf{g},p \neq q}{\sum}\frac{\underset{\{(i,j)|a_i=p,a_j=q\}}{\sum}d(X_i,X_j)(\mathbf{W} \cdot \mathbf{S}_{.,i}-\mathbf{W} \cdot \mathbf{S}_{.,j})^2}{|\{i|a_i = p\}||\{j|a_j = q\}|}}.
\end{flalign*}
Similarly, we calculate the derivative w.r.t. $\mathbf{W}$.
We first define a vector $\mathbf{m}_{pq} = \{d(X_i,X_j)|a_i=p,a_j = q\} \in \mathbb{R}^{|\{i|a_i=p\}||\{j|a_j=q\}|}$, which contains all of $d(.,.)$ values for every pair of instances from groups $p$ and $q$. 
Then, $\mathbf{M}_{pq}$ is the stack of $k$ $\mathbf{m}_{pq}$, thus $\mathbf{M}_{pq} \in \mathbb{R}^{k\times{|\{i|a_i=p\}||\{j|a_j=q\}|}}$.
Next, we define a matrix $\mathbf{D}_{pq} = \{\mathbf{S}_{.,i} - \mathbf{S}_{.,j} |a_i = p,a_j = q\} \in \mathbb{R}^{{|\{i|a_i=p\}||\{j|a_j=q\}|}\times k}$, which represents the values of $\mathbf{S}_{.,i} - \mathbf{S}_{.,j}$ for every pair of instances $i$,$j$ from group $p$ and $q$ respectively.
The derivative result is:
\begin{flalign*}
&\frac{d\text{L}_2}{d\mathbf{W}} = 2(\mathbf{B}\odot\mathbf{S}^\intercal) \mathbf{S} \mathbf{W} -&\\\nonumber 
&2(\mathbf{B}\odot\mathbf{S}^\intercal) \mathbf{t} + \frac{2\alpha}{N}{\underset{p,q \in \mathbf{g},p\neq q}{\sum}\frac{(\mathbf{M}_{pq} \odot \mathbf{D}_{pq}^\intercal)\mathbf{D}_{pq}\mathbf{W}}{|\{i|a_i = p\}||\{j|a_j = q\}|}}.
\end{flalign*}
Then, let $\frac{d\text{L}_2}{d\mathbf{W}} = 0$, we get:
\begin{flalign*}
&0=2(\mathbf{B}\odot\mathbf{S}^\intercal) \mathbf{S} \mathbf{W} -&\\\nonumber
&2(\mathbf{B}\odot\mathbf{S}^\intercal) \mathbf{t} + 
\frac{2\alpha}{N}{\underset{p,q \in \mathbf{g},p\neq q}{\sum}\frac{(\mathbf{M}_{pq} \odot \mathbf{D}_{pq}^\intercal)\mathbf{D}_{pq}\mathbf{W}}{|\{i|a_i = p\}||\{j|a_j = q\}|}}.
\end{flalign*}
Thus, the final solution is:
\begin{flalign*}
&\mathbf{W} = 
[(\mathbf{B}\odot\mathbf{S}^\intercal) \mathbf{S} + &\\\nonumber
&\frac{\alpha}{N}{\underset{p,q \in \mathbf{g},p\neq q}{\sum}\frac{(\mathbf{M}_{pq} \odot \mathbf{D}_{pq}^\intercal)\mathbf{D}_{pq}\mathbf{W}}{|\{i|a_i = p\}||\{j|a_j = q\}|}}]^{-1}[(\mathbf{B}\odot\mathbf{S}^\intercal) \mathbf{t}].
\end{flalign*}

\subsection{Supplementary Experiments}
The other four datasets and the corresponding experimental results are presented as follows.
\subsubsection{Datasets}
The statistics of these four datasets are summarized in the Table \ref{tab:datasets}. 
\begin{table}[ht]
\centering
\begin{threeparttable}
\caption{Statistics for the datasets.}\label{tab:datasets}
\begin{tabular}{@{} ccc @{}}\hline
    \bfseries Dataset &\bfseries Inlier/Outlier Size & \bfseries Groups\\\hline
    \bfseries Japanese Vowels   &1,406/50 & 3 \\
    \bfseries Breast Cancer    &444/239 & 3 \\
    \bfseries Mammography   & 10,923/260 &4 \\
    \bfseries Pima   & 500/268 &4 \\\hline
\end{tabular}
\begin{tablenotes}
\centering
      \footnotesize
      \item ``Groups'' denotes the number of the protected groups.
    \end{tablenotes}
    \end{threeparttable}
\end{table}

\begin{itemize}
\item \textbf{Japanese Vowels DataSet}\footnote{\url{http://odds.cs.stonybrook.edu/japanese-vowels-data/}} 
describes the pronunciation of the vowels /ae/.
The ground truth attribute denotes the outliers and inliers.

\item \textbf{Breast Cancer Dataset}\footnote{\url{http://odds.cs.stonybrook.edu/breast-cancer-wisconsin-original-dataset/}} 
describes the status of the breast cancer. 
``Benign'' and ``malignant'' classes are treated as inliers and outliers separately.

\item \textbf{Mammography Dataset}\footnote{\url{http://odds.cs.stonybrook.edu/mammography-dataset/}} 
describes the microcalcifications information for mammography. 
The ground truth attribute divides instances into outliers and inliers.

\item \textbf{Pima Dataset}\footnote{\url{http://odds.cs.stonybrook.edu/pima-indians-diabetes-dataset/}} describes diabetes status of human.
The outliers are those observations having diabetes, otherwise, the observations are labeled as inliers.

\end{itemize}

We manually add one protected attribute to denote the protected group in these four datasets.
Instances are sampled into different groups to induce bias.

\begin{figure}[!htb]
    \centering
    \subfigure[Japanese Vowels Dataset.]{
        \includegraphics[height=0.15\textwidth,width=0.20\textwidth]{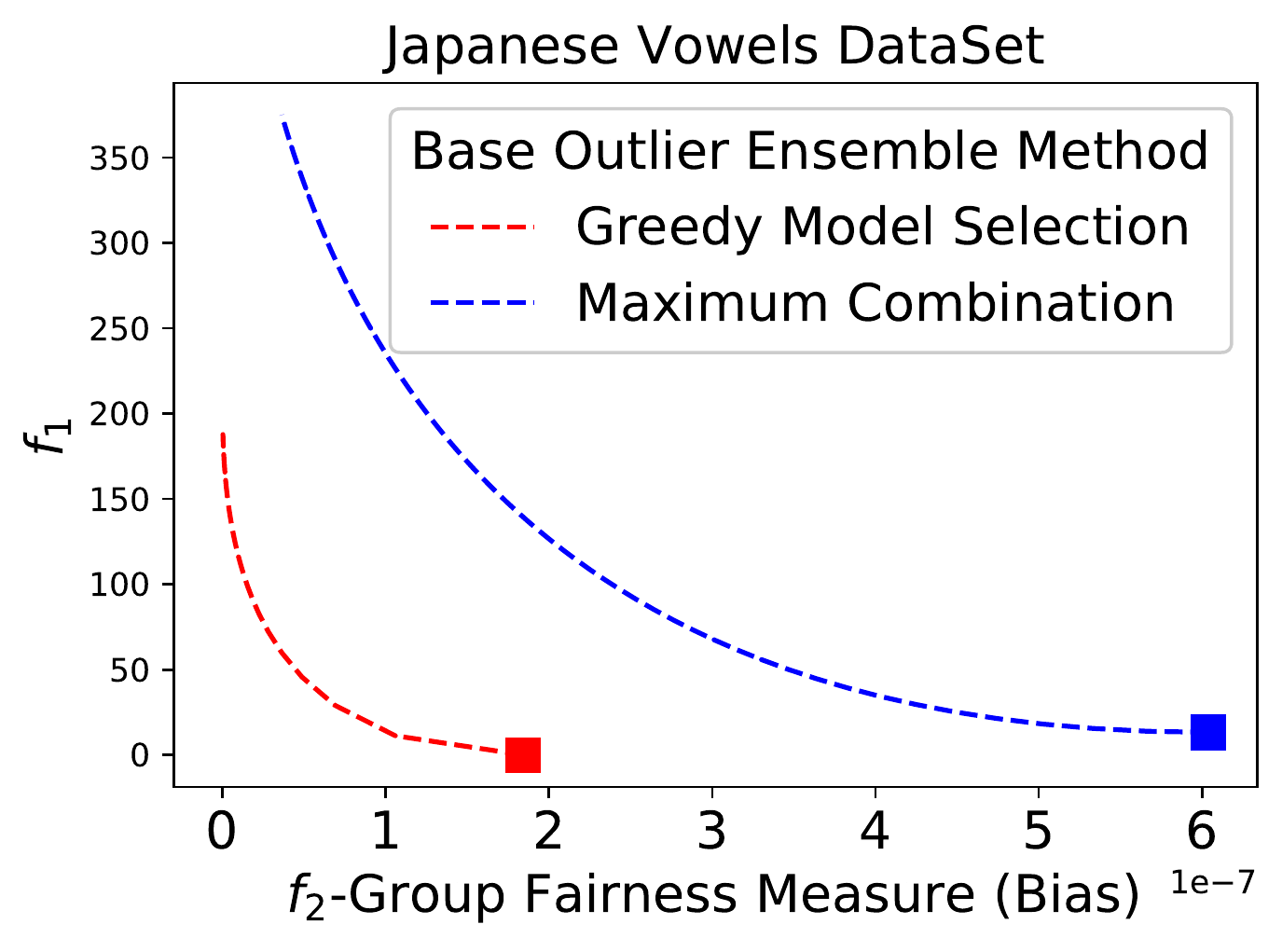}
    }
    \subfigure[Breast Cancer Dataset]{
        \includegraphics[height=0.15\textwidth,width=0.20\textwidth]{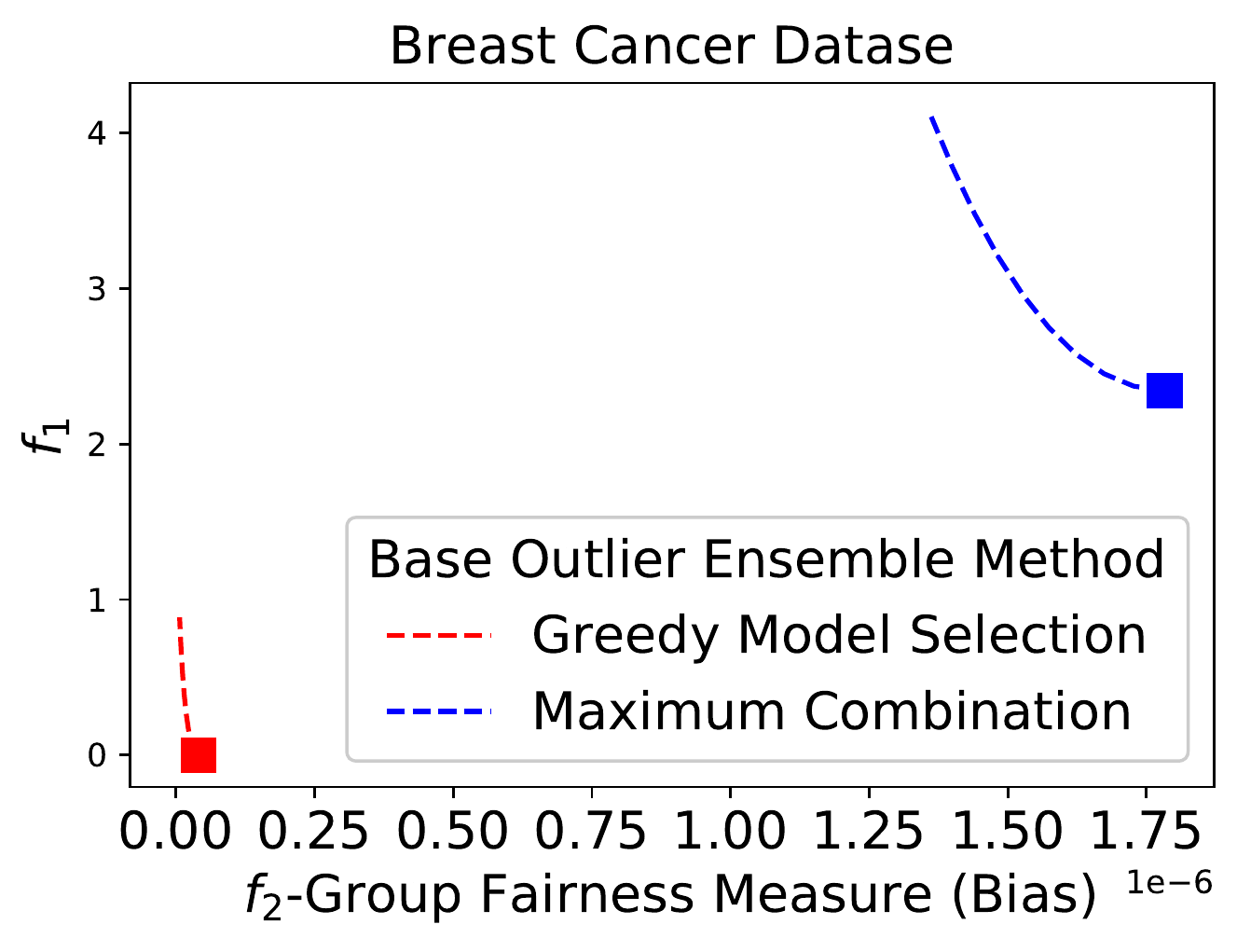}
    }
    
    \subfigure[Mammography Dataset]{
        \includegraphics[height=0.15\textwidth,width=0.20\textwidth]{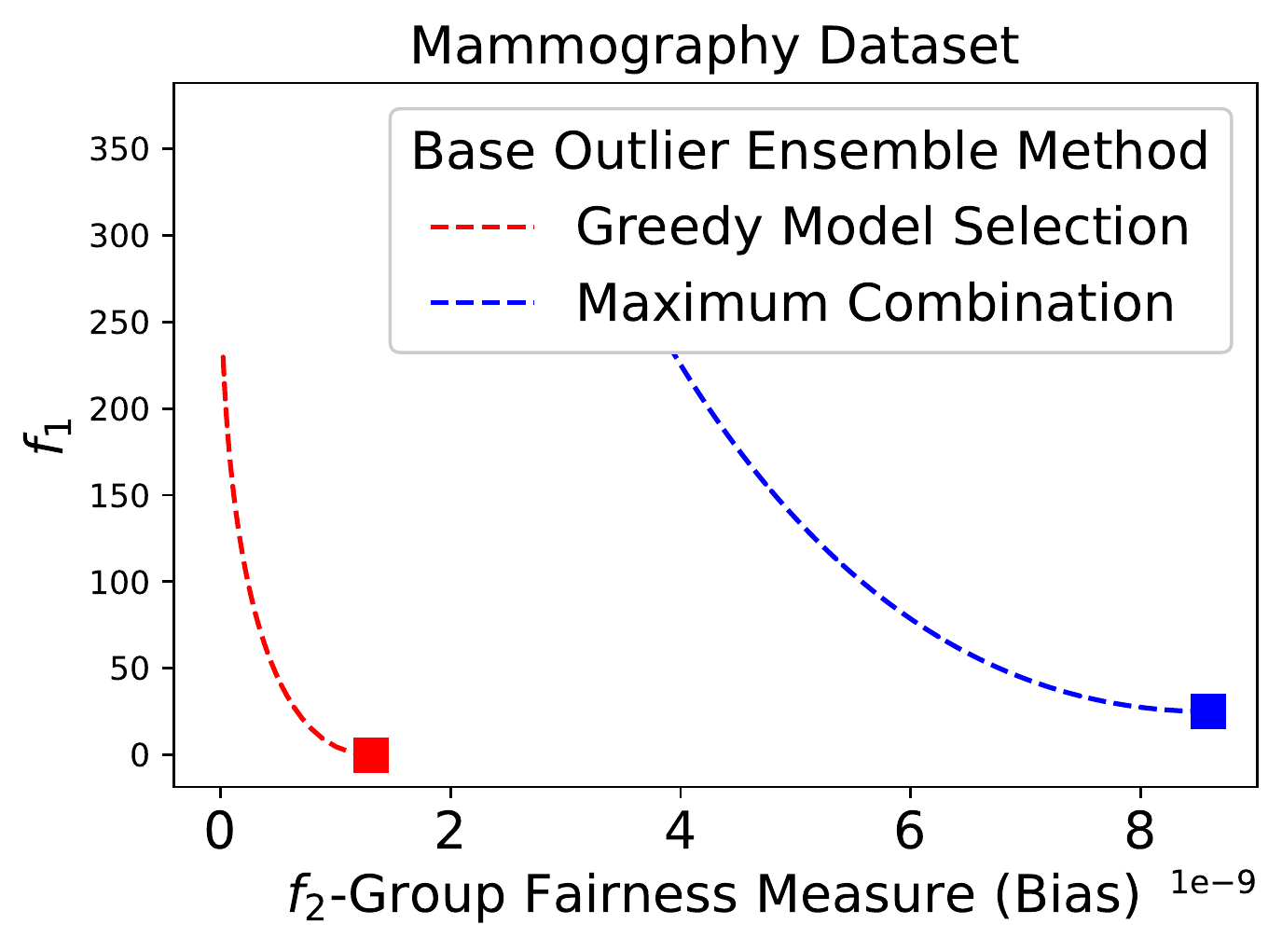}
    }
    \subfigure[Pima Dataset]{
        \includegraphics[height=0.15\textwidth,width=0.20\textwidth]{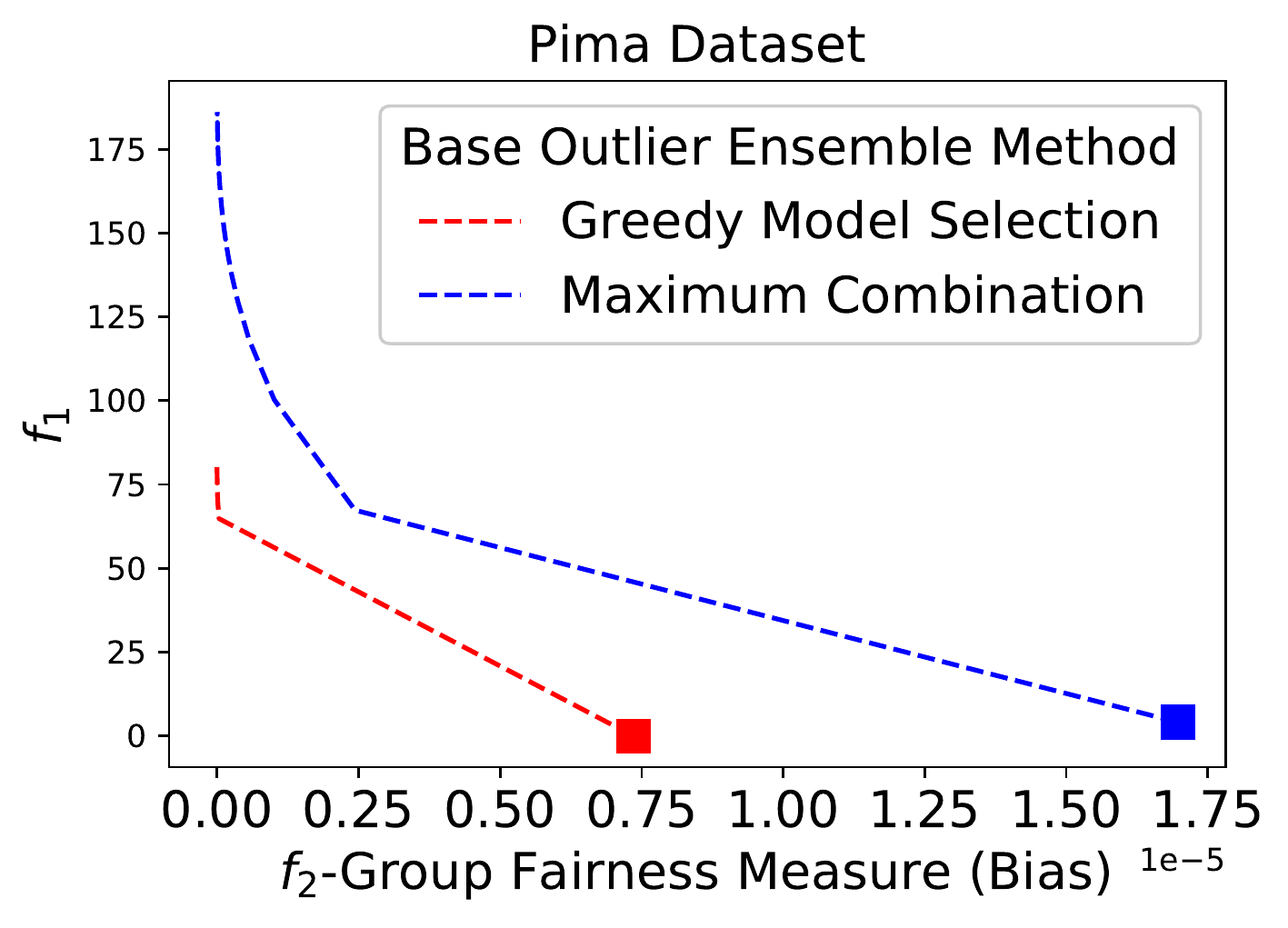}
    }    
    \caption{$f_2$-$f_1$ curves for group fairness.}
    \label{fig:f1f2groupfairness}

\end{figure}
\begin{figure}[!htb]
    \centering
    \subfigure[Japanese Vowels Dataset.]{
        \includegraphics[height=0.15\textwidth,width=0.20\textwidth]{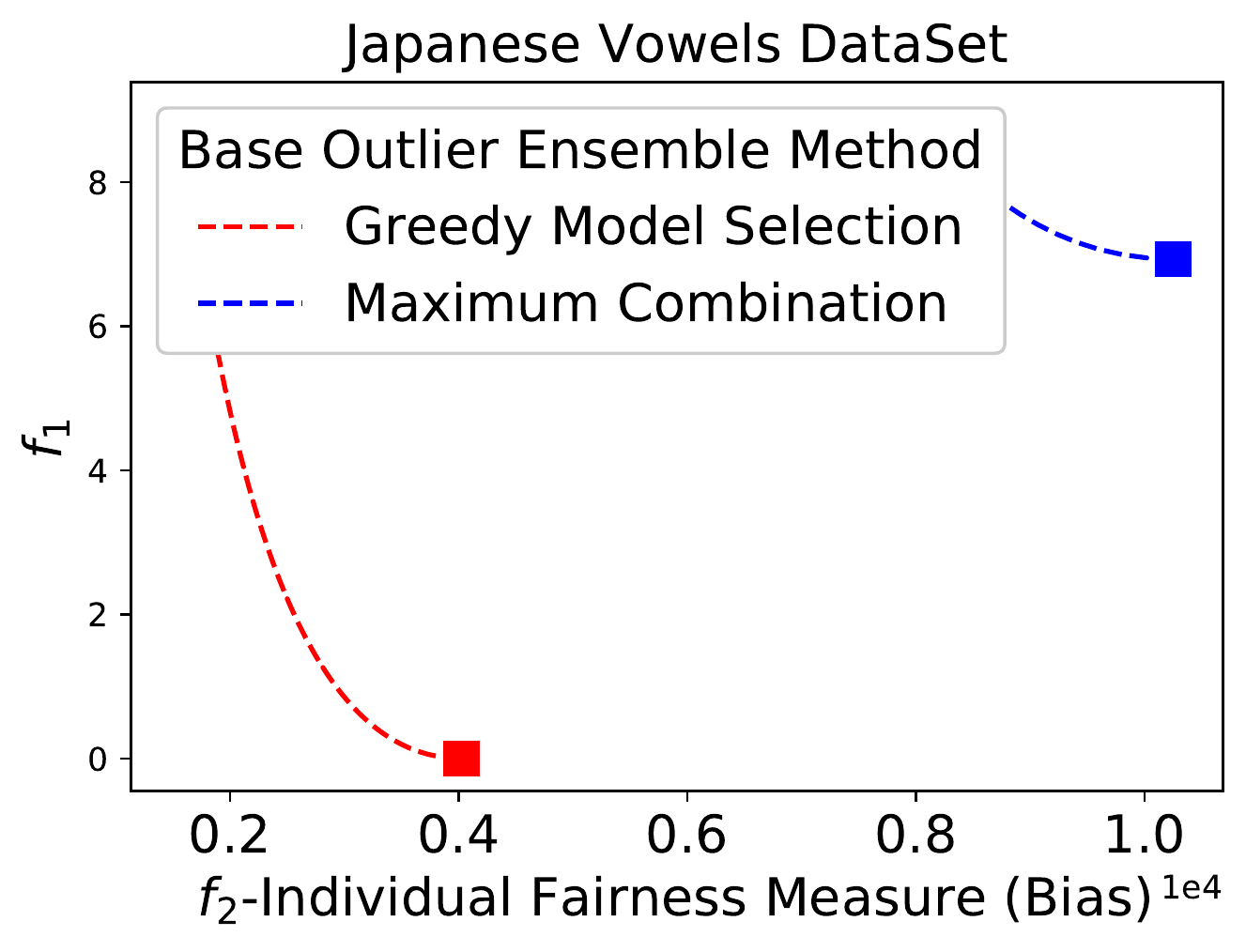}
    }
    \subfigure[Breast Cancer Dataset]{
        \includegraphics[height=0.15\textwidth,width=0.20\textwidth]{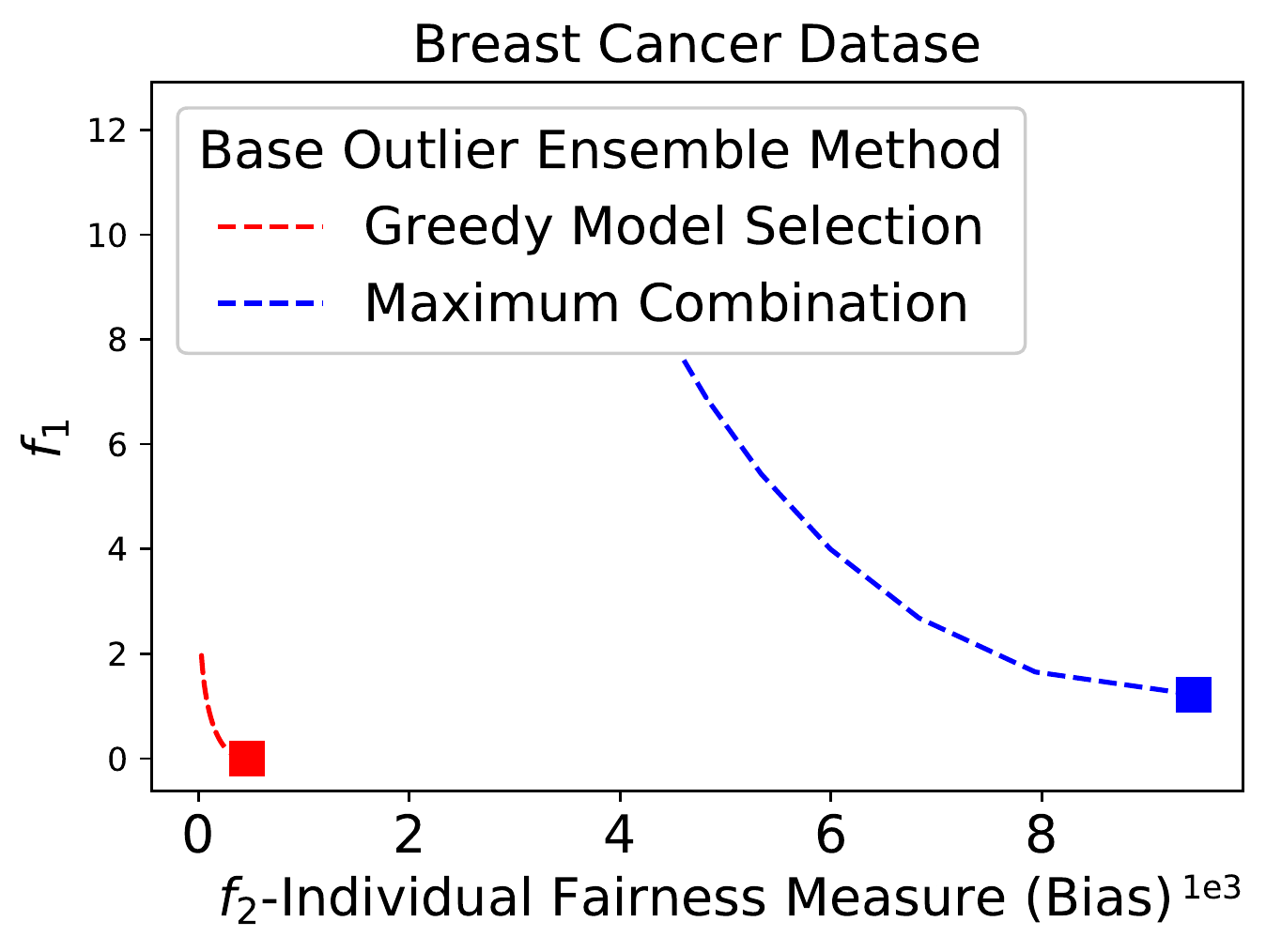}
    }
    
    \subfigure[Mammography Dataset]{
        \includegraphics[height=0.15\textwidth,width=0.20\textwidth]{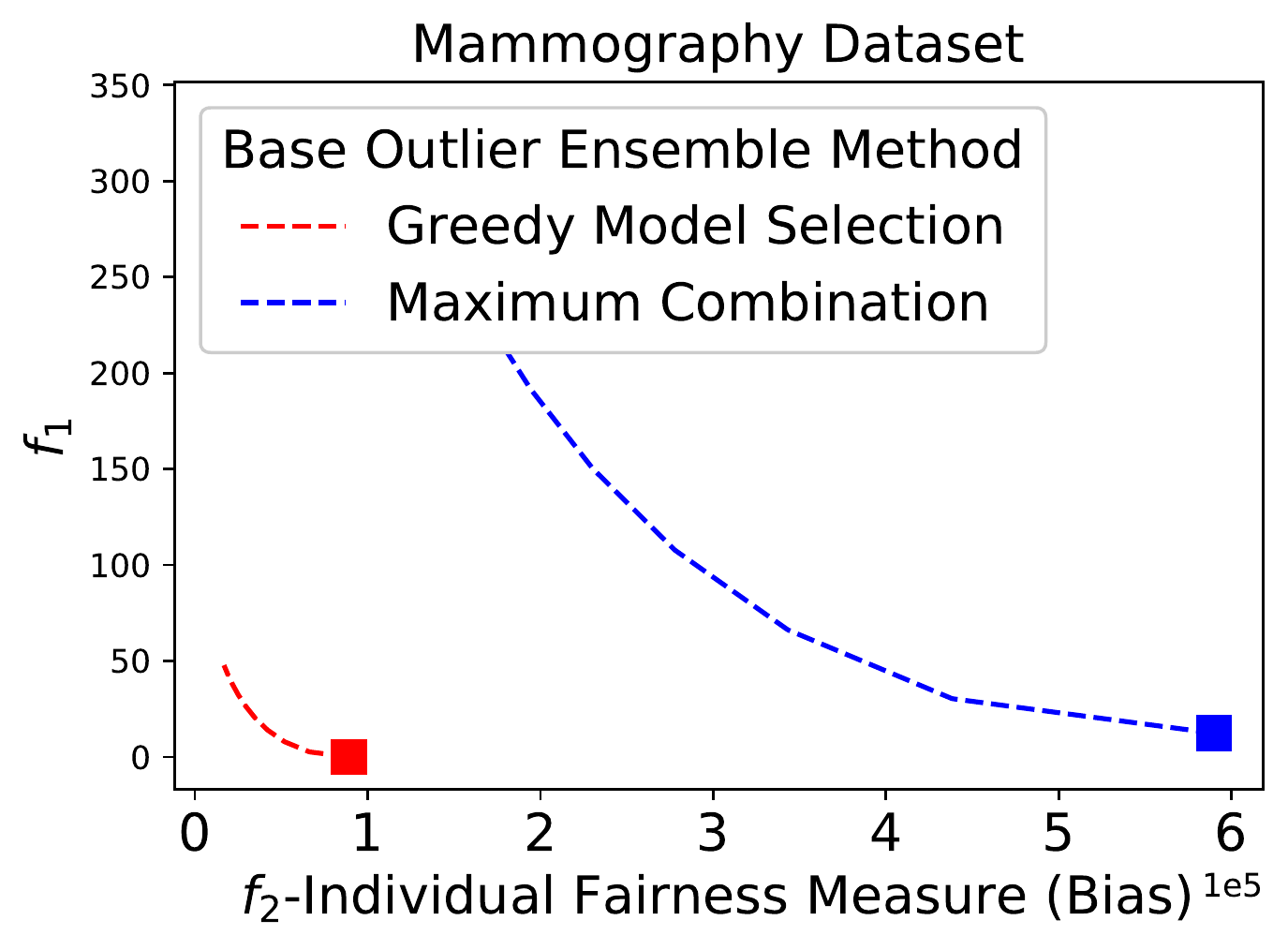}
    }
    \subfigure[Pima Dataset]{
        \includegraphics[height=0.15\textwidth,width=0.20\textwidth]{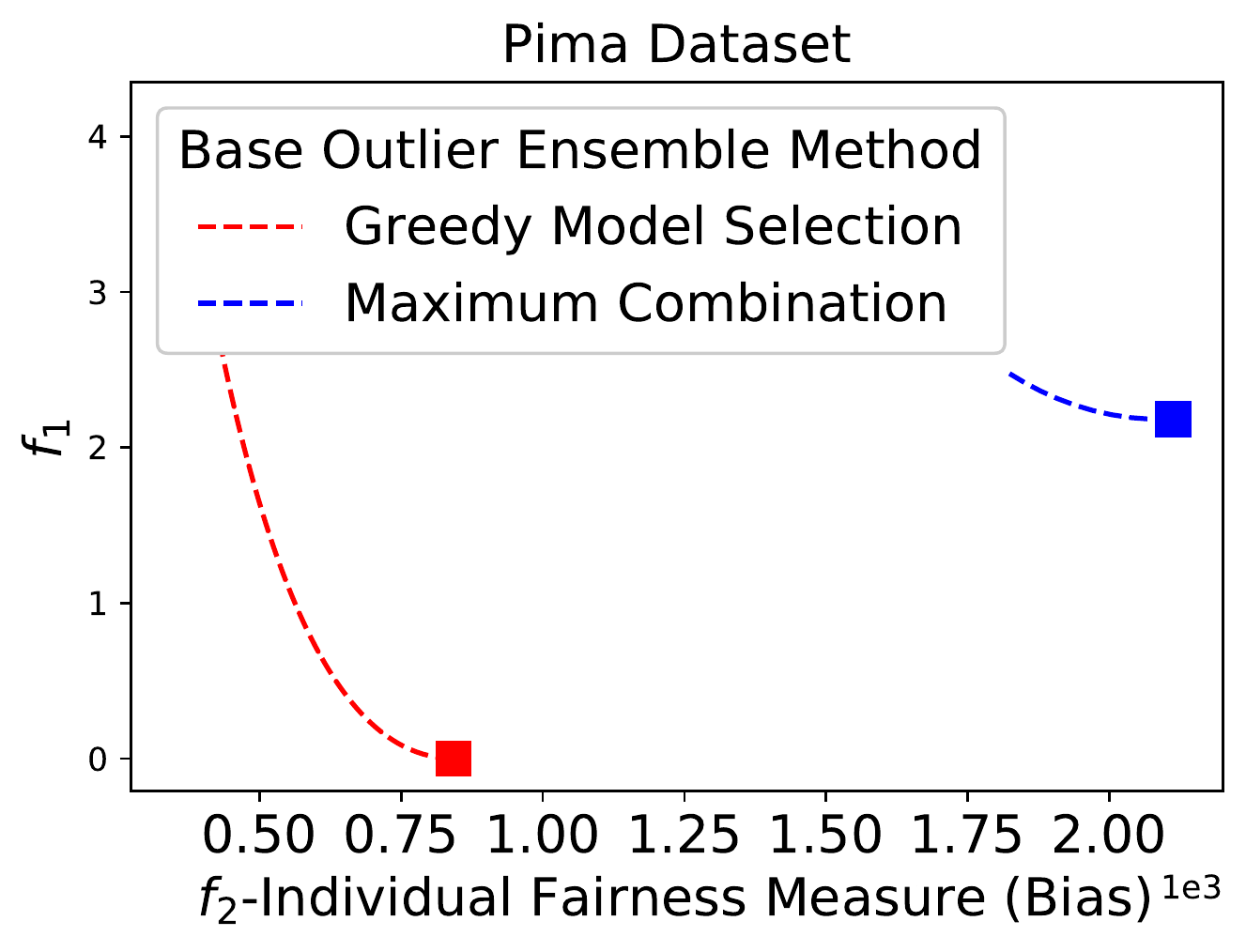}
    }    
    \caption{$f_2$-$f_1$ curves for individual fairness.}
    \label{fig:f1f2individualfairness}
\end{figure}

\subsubsection{Experiment Results}
For the relations between $f_1$ and $f_2$ functions, results for group fairness and individual fairness are presented in Fig.~\ref{fig:f1f2groupfairness} and Fig.~\ref{fig:f1f2individualfairness} respectively.

\begin{figure}[!htb]
    \centering
    \subfigure[Japanese Vowels Dataset.]{
        \includegraphics[height=0.15\textwidth,width=0.20\textwidth]{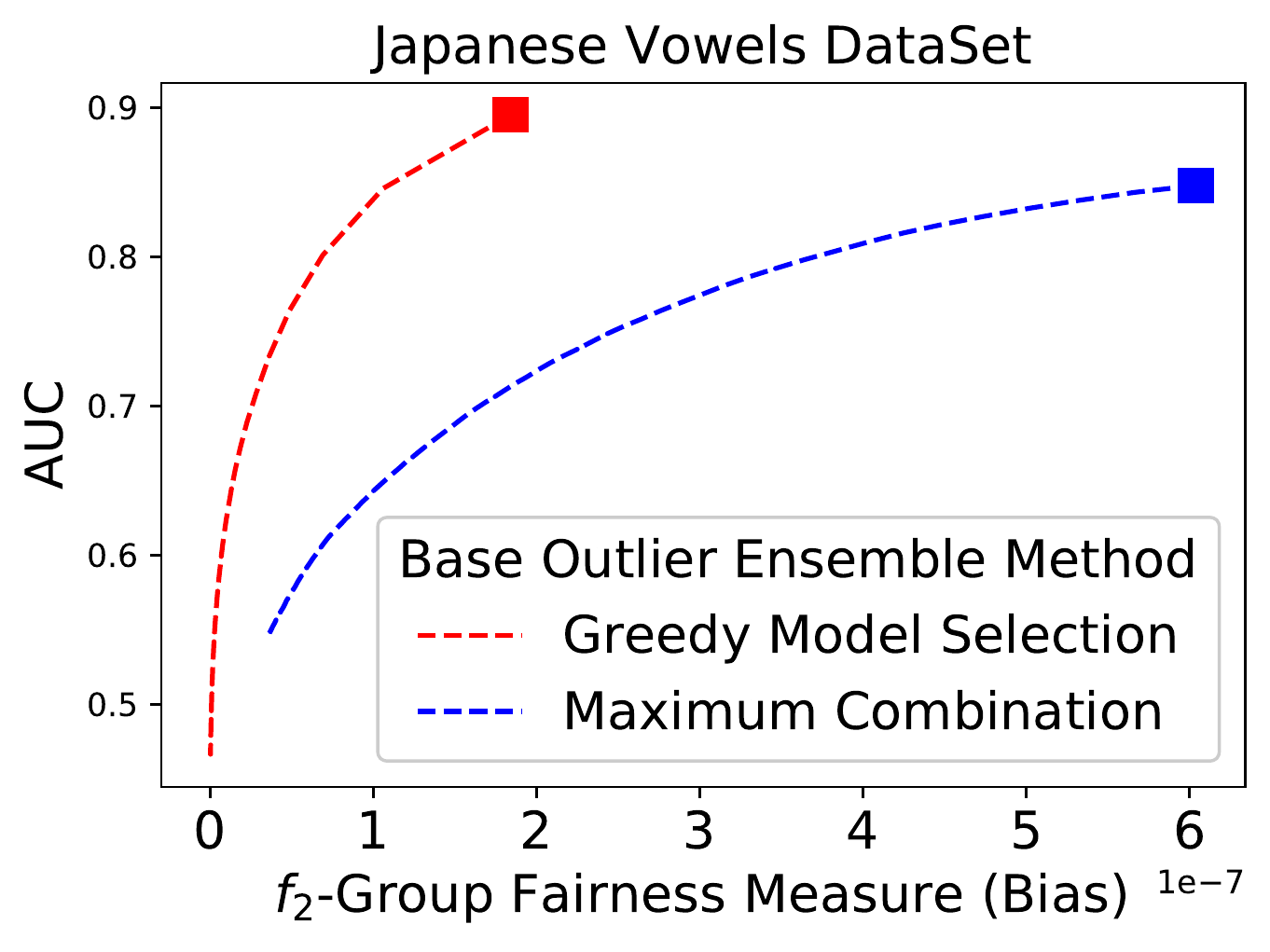}
    }
    \subfigure[Breast Cancer Dataset]{
    \label{fig:breastw_bias_AUC_group_fairness}
        \includegraphics[height=0.15\textwidth,width=0.20\textwidth]{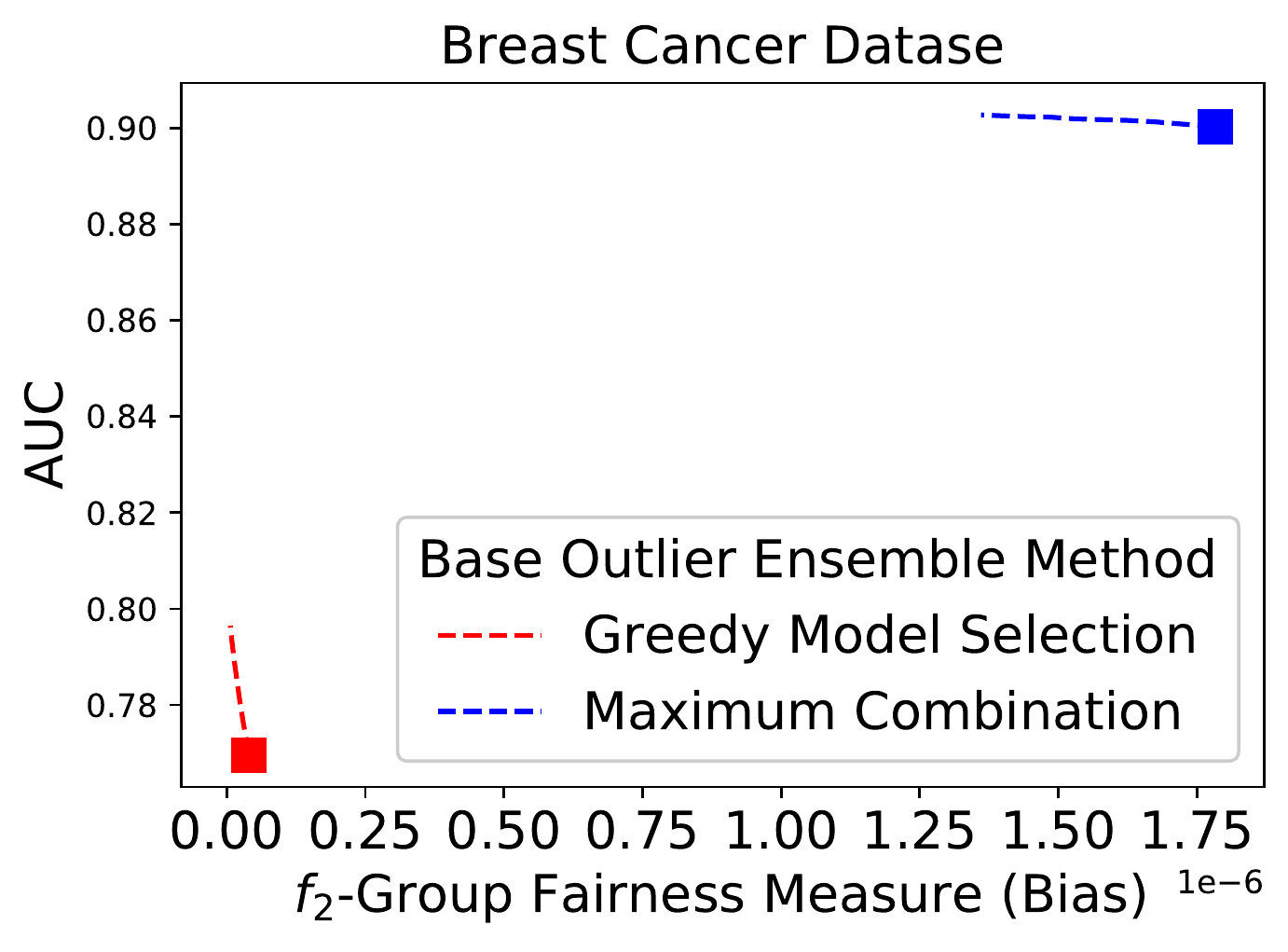}
    }
    
    \subfigure[Mammography Dataset]{
        \includegraphics[height=0.15\textwidth,width=0.20\textwidth]{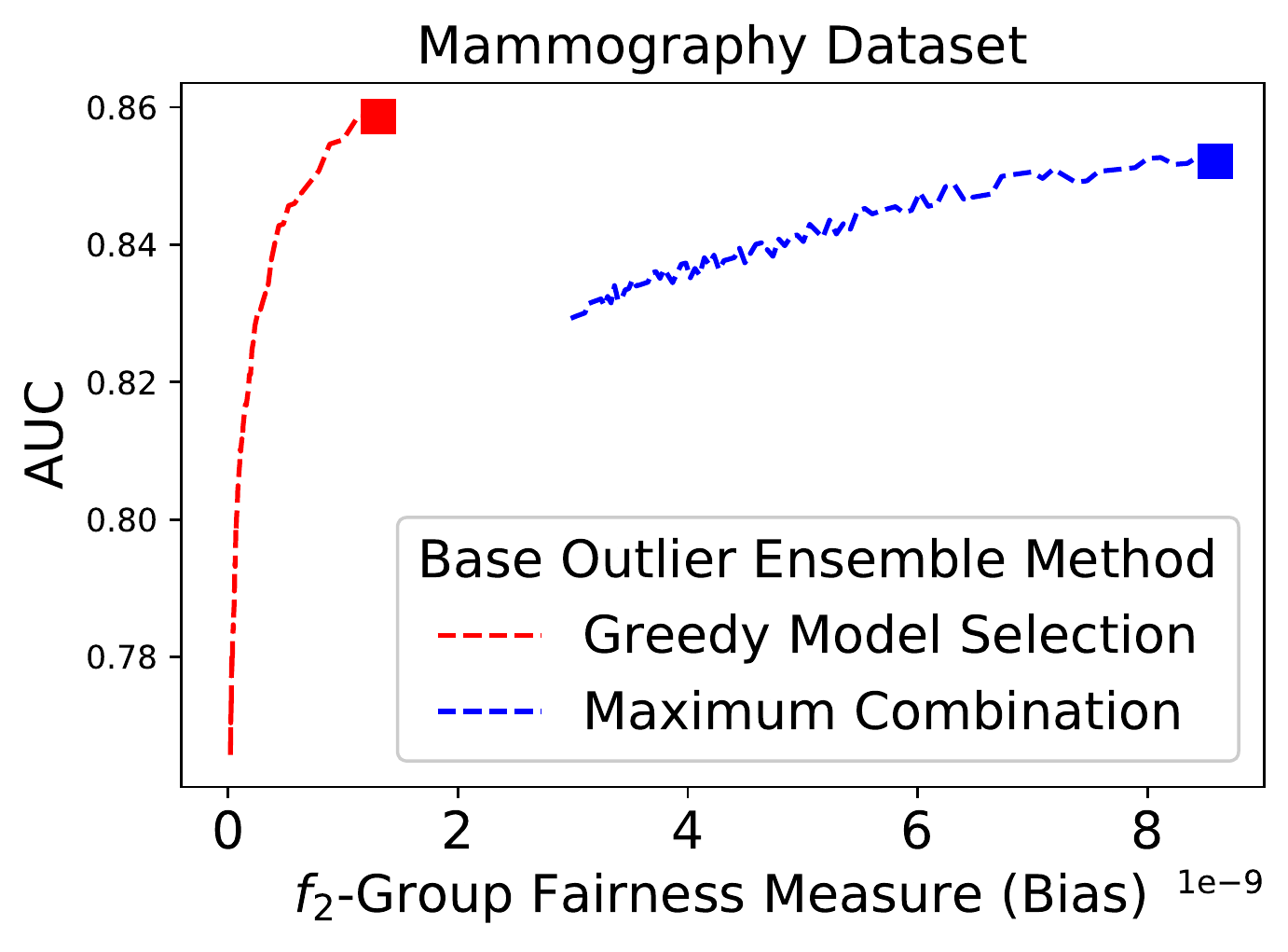}
    }
    \subfigure[Pima Dataset]{
        \includegraphics[height=0.15\textwidth,width=0.20\textwidth]{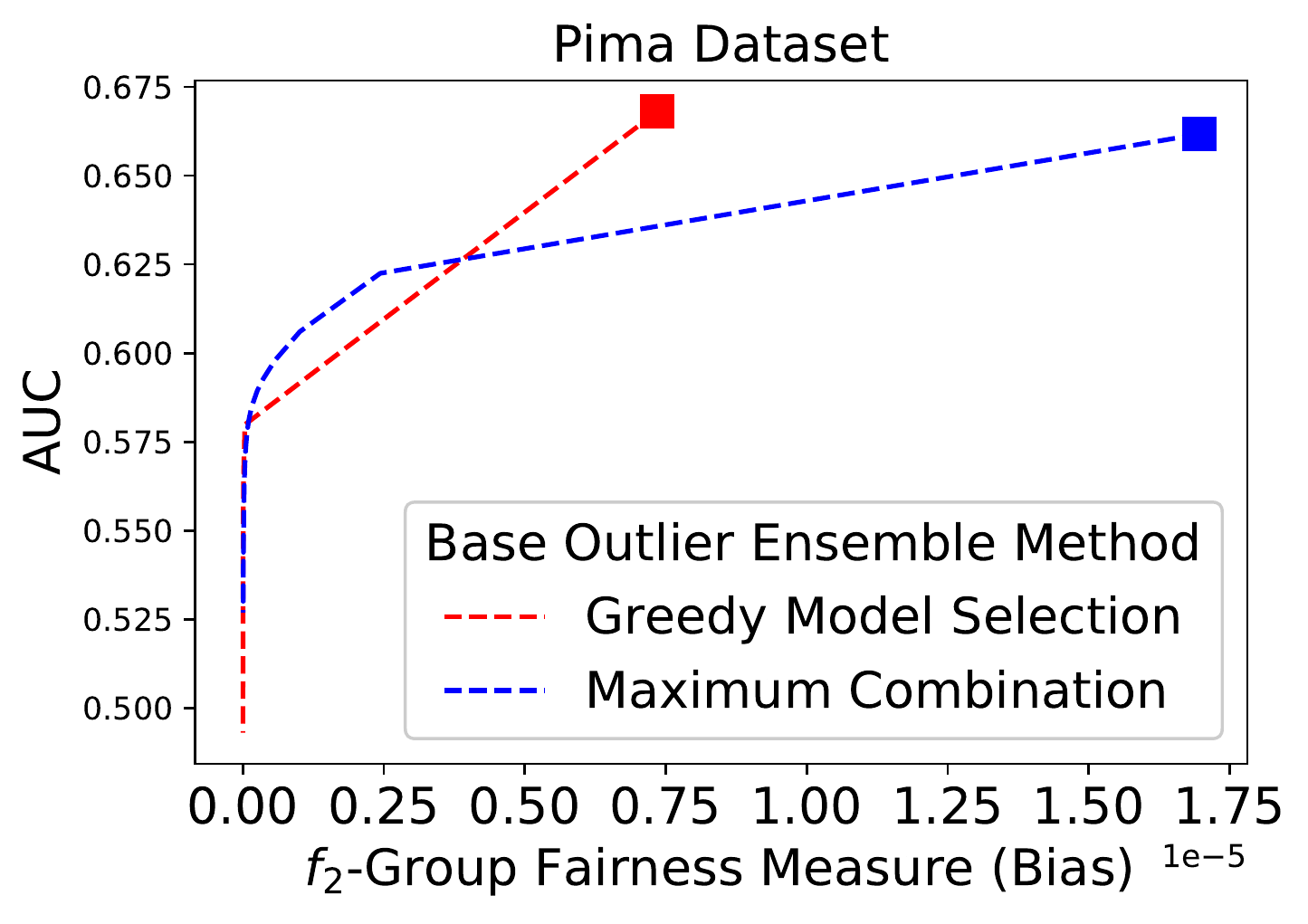}
    }    
    \caption{Bias-AUC curves for group fairness.}
    \label{fig:barelation1}
\end{figure}

\begin{figure}[!htb]
    \centering
    \subfigure[Japanese Vowels Dataset.]{
        \includegraphics[height=0.15\textwidth,width=0.20\textwidth]{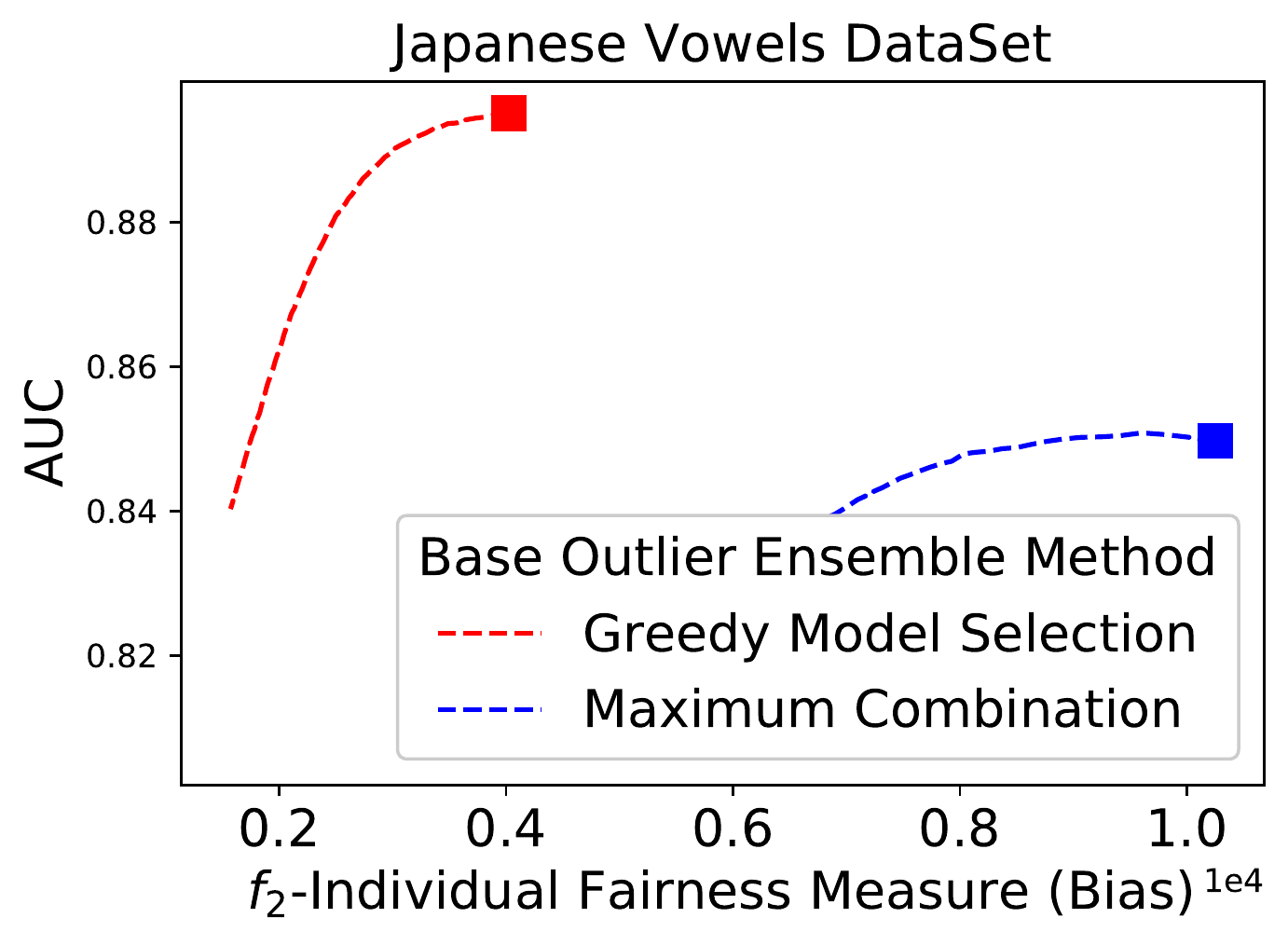}
    }
    \subfigure[Breast Cancer Dataset]{
        
        \includegraphics[height=0.15\textwidth,width=0.20\textwidth]{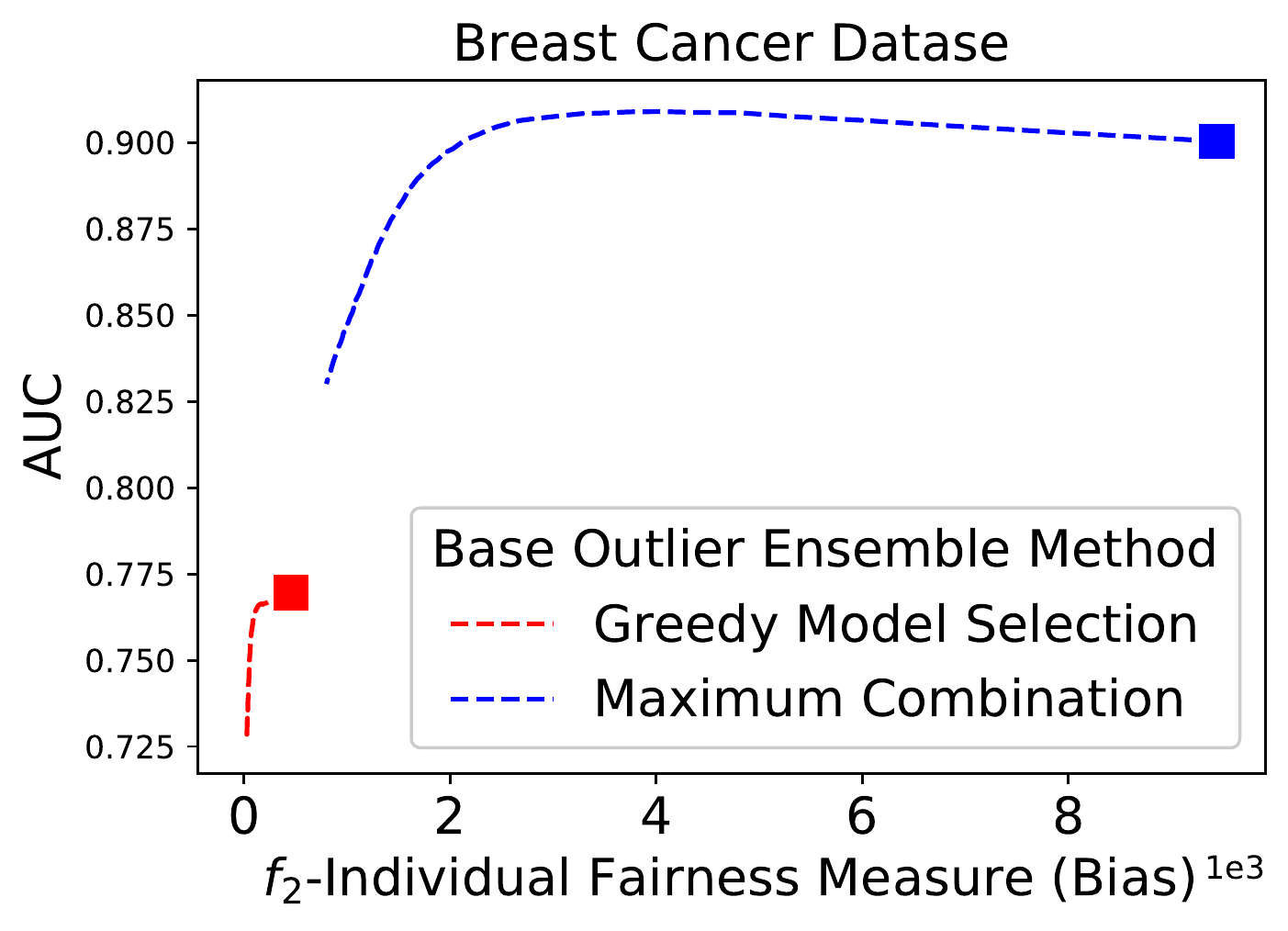}
    }
    
    \subfigure[Mammography Dataset]{
        \includegraphics[height=0.15\textwidth,width=0.20\textwidth]{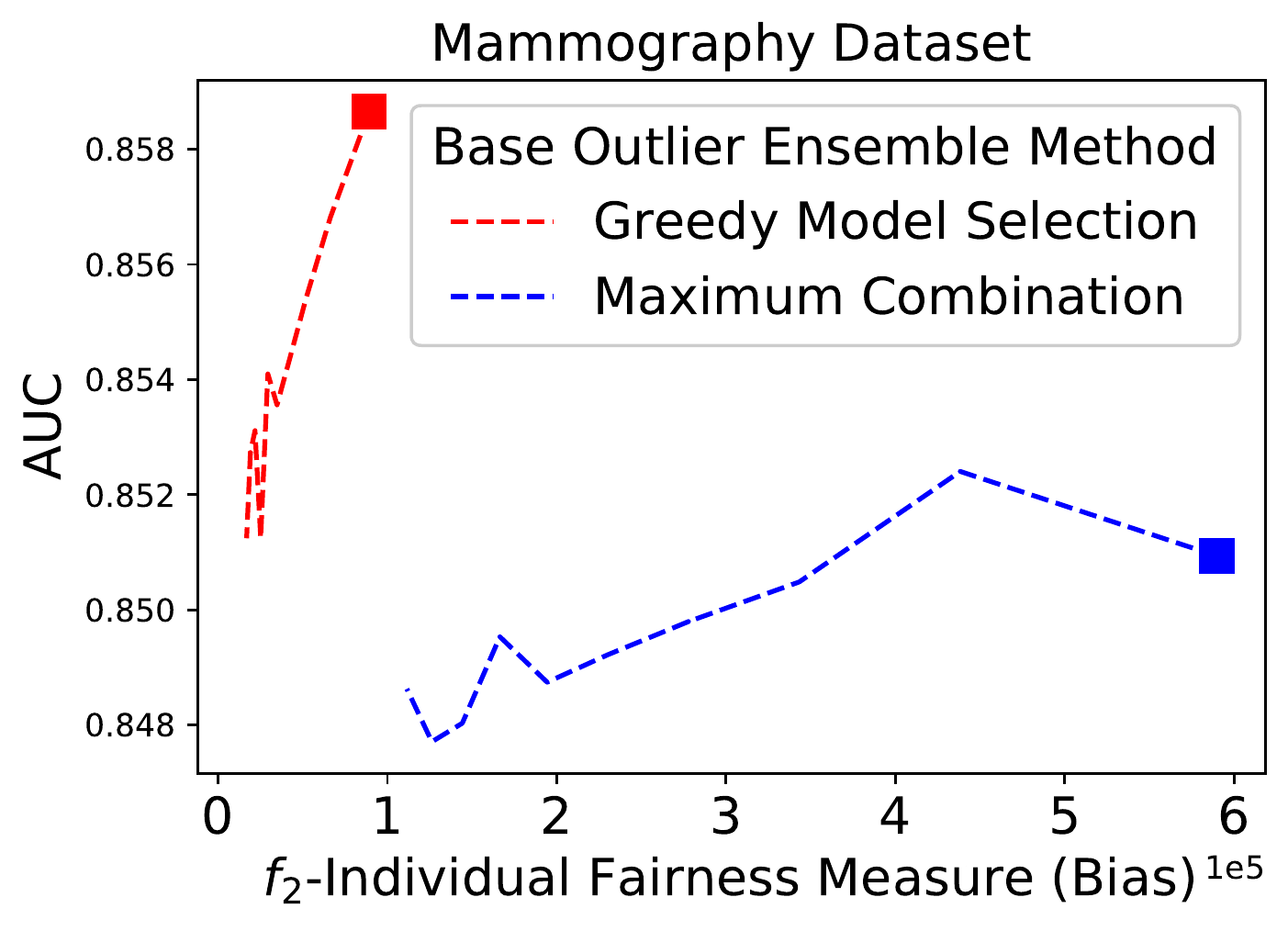}
    }
    \subfigure[Pima Dataset]{
        \includegraphics[height=0.15\textwidth,width=0.20\textwidth]{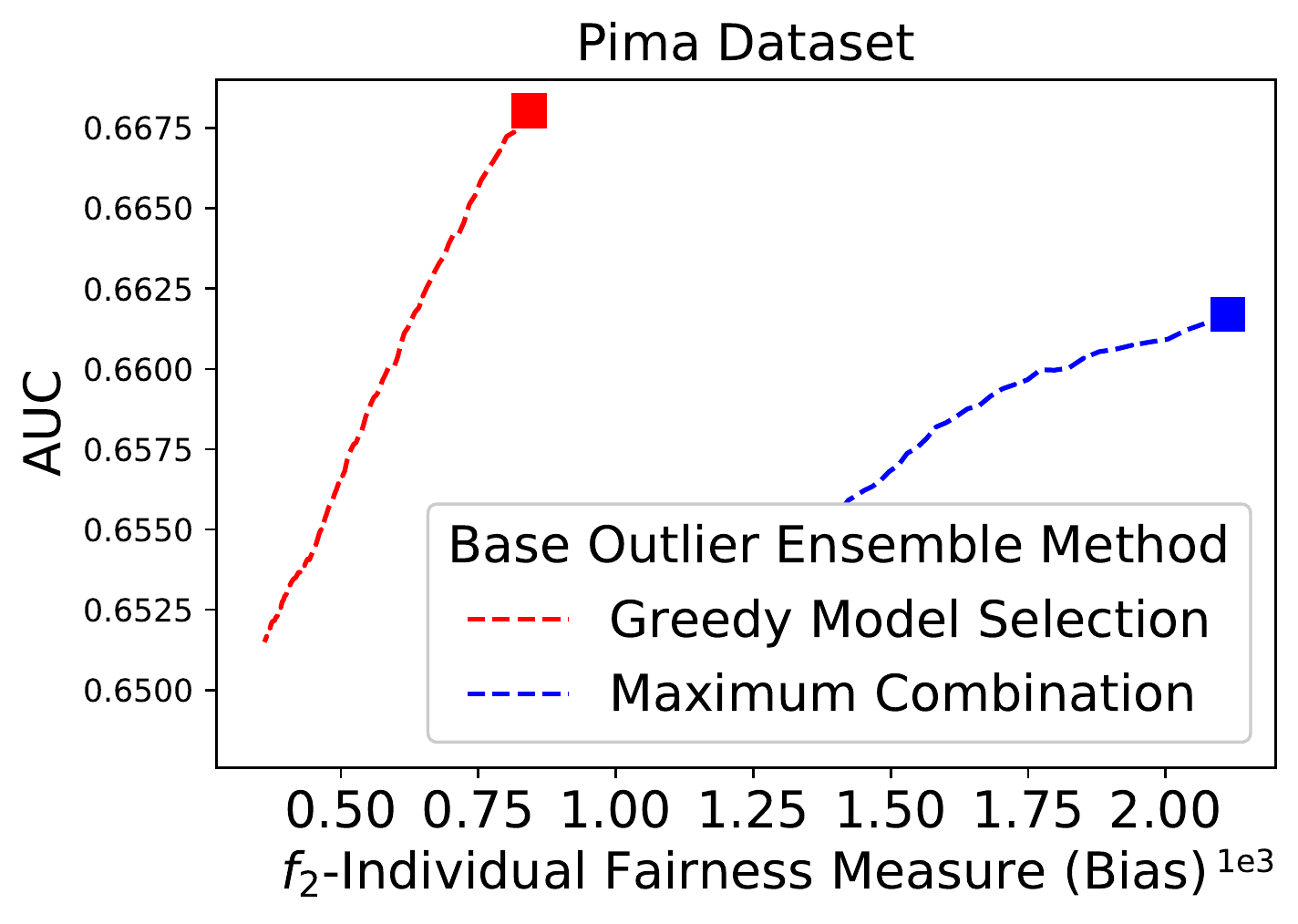}
    }    
    \caption{Bias-AUC curves for individual fairness.}
    \label{fig:barelation2}
\end{figure}

For the relations between AUC and bias, results are presented in Fig.~\ref{fig:barelation1} and Fig.~\ref{fig:barelation2} respectively.
From Fig.~\ref{fig:barelation1} where group fairness is utilized, the experiments show that most AUC values drop during the decrease of the bias values.
However, in the experiment results on the Breast Cancer Dataset, when fairness is decreasing, the AUC is improved.
Since the $f_1$ function minimizes the distance between newly generated results and original ensemble results, if the original ensemble output is not optimal, the slight change may also lead to the improvement of AUC.
For Breast Cancer Dataset in Fig.~\ref{fig:breastw_bias_AUC_group_fairness}, the square marker denotes the experimental results when $\alpha = 0$. We find that the AUC of Greedy Model Selection and Maximum Combination are 0.7696 and 0.8954, respectively, which are far from the optimal (a simple Average Combination method
can obtain 0.9784 of AUC on this dataset).
Therefore, during the improvement of the fairness, the subtle difference with $\mathbf{t}$ results in an improvement of AUC.
Then, for the experiments in Fig.~\ref{fig:barelation2} where individual fairness is utilized, a fluctuation can be found in the results of Mammography Dataset, which also shows that the AUC may increase if the target outlier score is not the optimal.
Regardless of these special situations, most of experimental results indicate when AUC decreases, there exists a \emph{cost} on AUC for improving the fairness of the outlier ensemble result.

\begin{figure}[!h]
    \centering
    \subfigure[Japanese Vowels Dataset.]{
        \includegraphics[height=0.15\textwidth,width=0.20\textwidth]{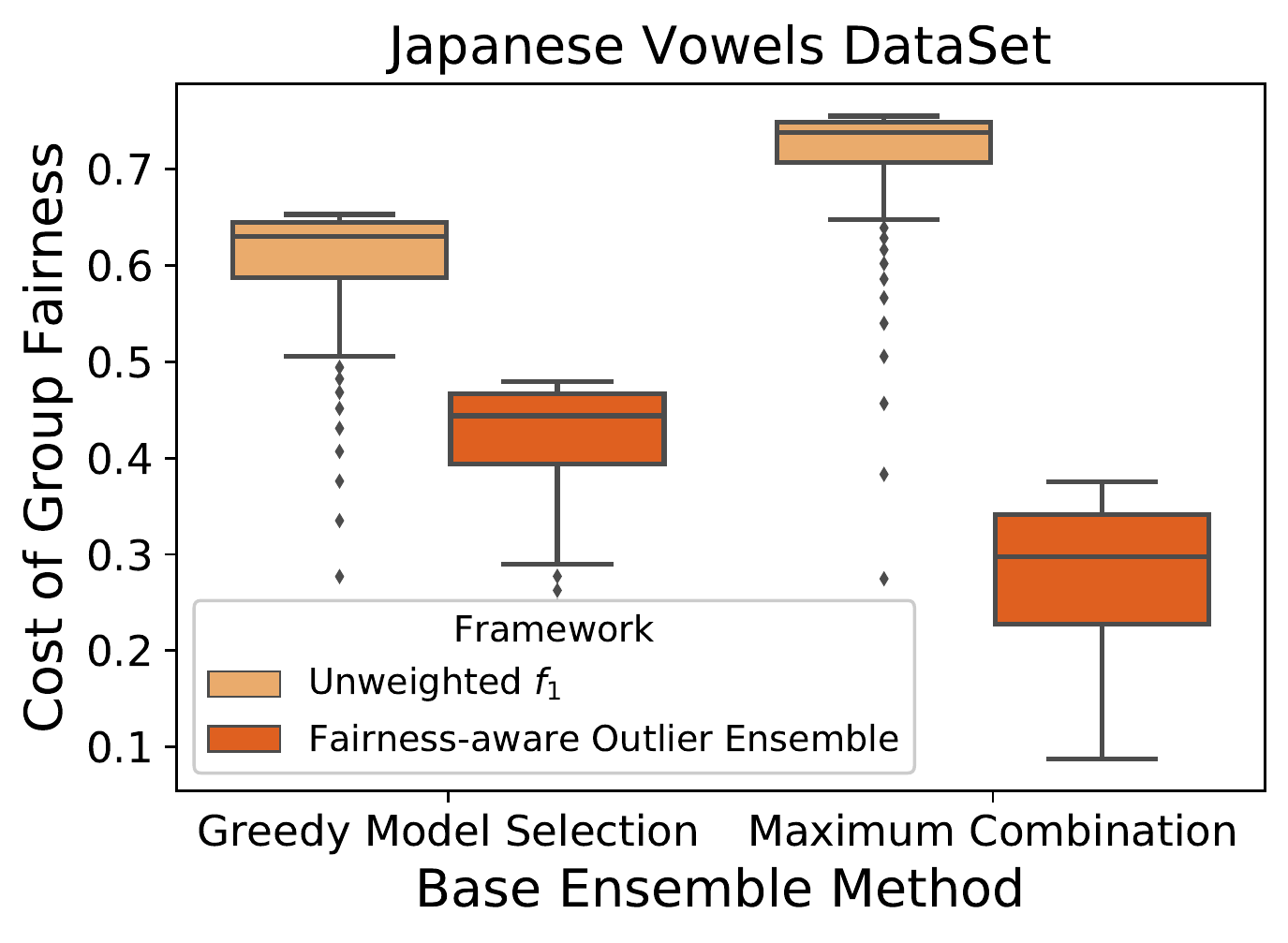}
    }
    \subfigure[Breast Cancer Dataset]{
        \includegraphics[height=0.15\textwidth,width=0.20\textwidth]{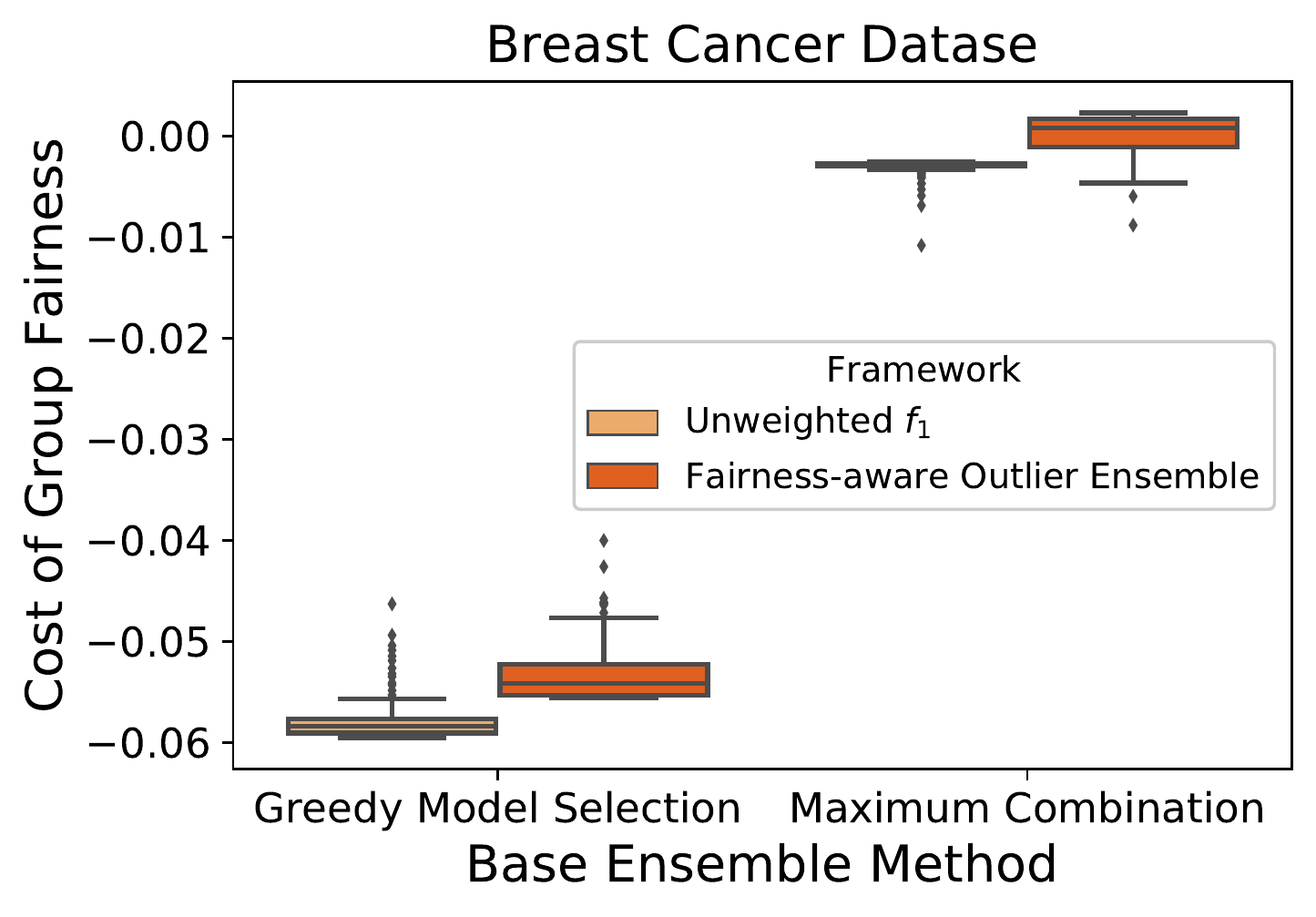}
    }
    \subfigure[Mammography Dataset]{
        \includegraphics[height=0.15\textwidth,width=0.20\textwidth]{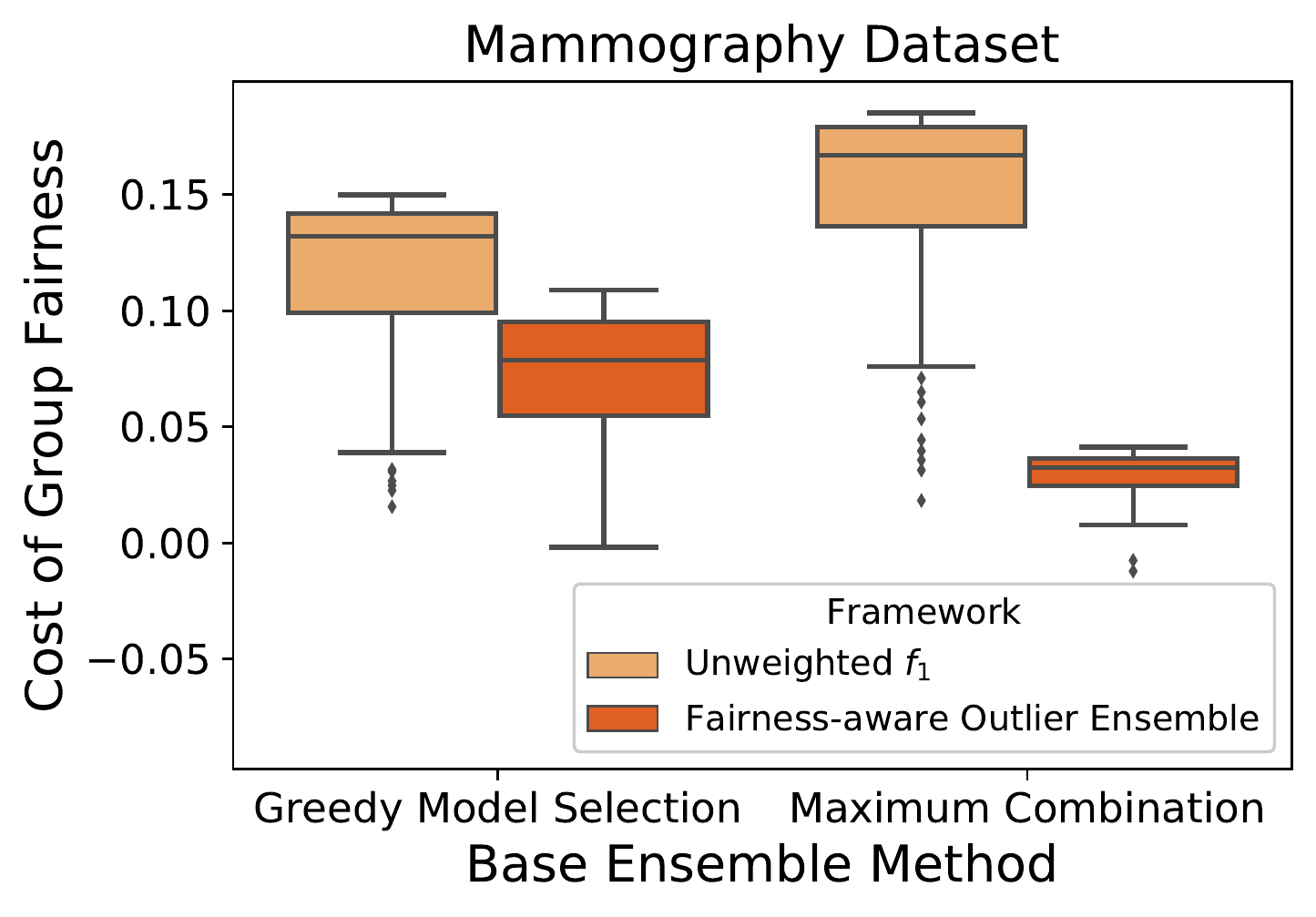}
    }
    \subfigure[Pima Dataset]{
        \includegraphics[height=0.15\textwidth,width=0.20\textwidth]{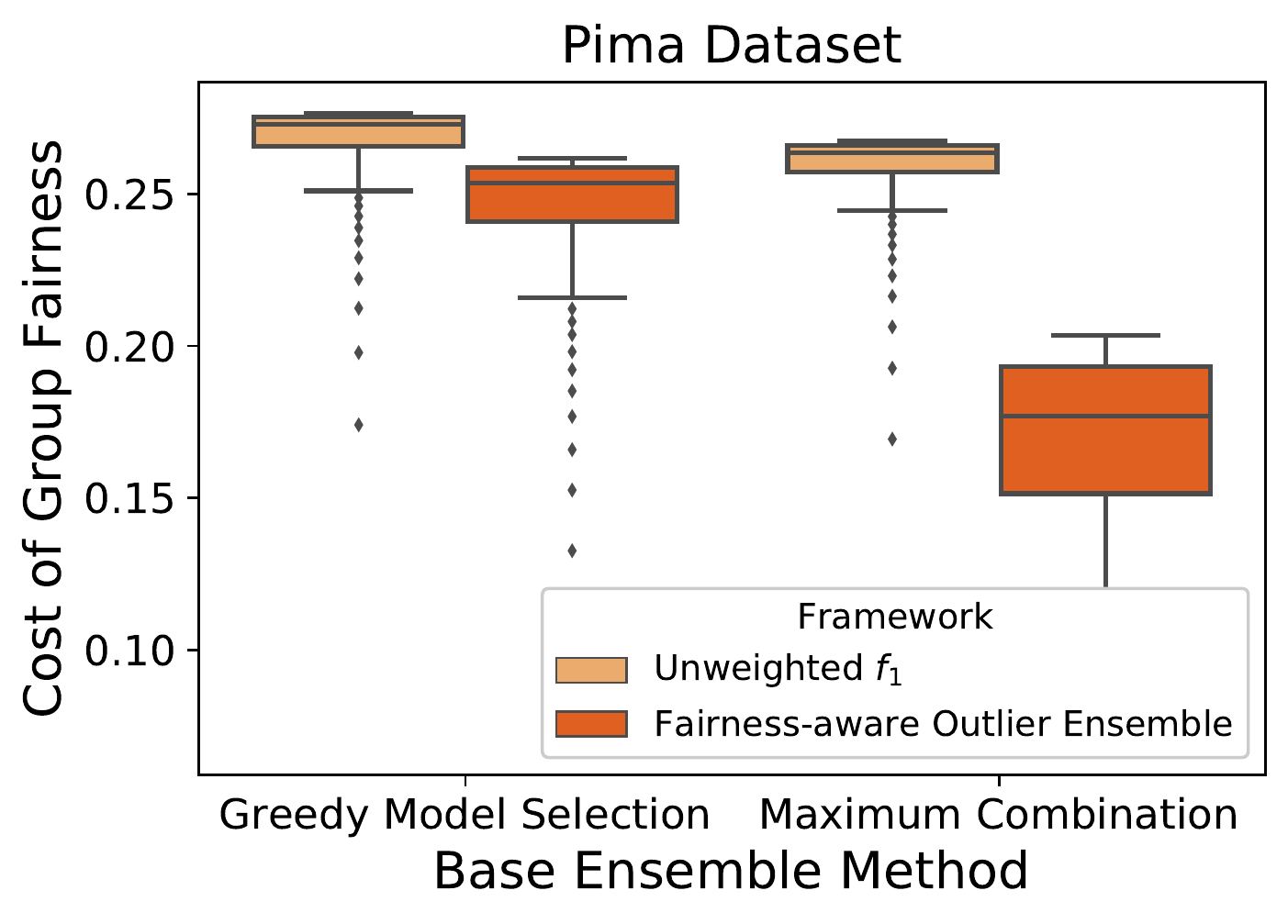}
    }    
    \caption{Cost of group fairness.}
    \label{fig:cogf}
\end{figure}

\begin{figure}[!h]
    \centering
    \subfigure[Japanese Vowels Dataset.]{
        \includegraphics[height=0.15\textwidth,width=0.20\textwidth]{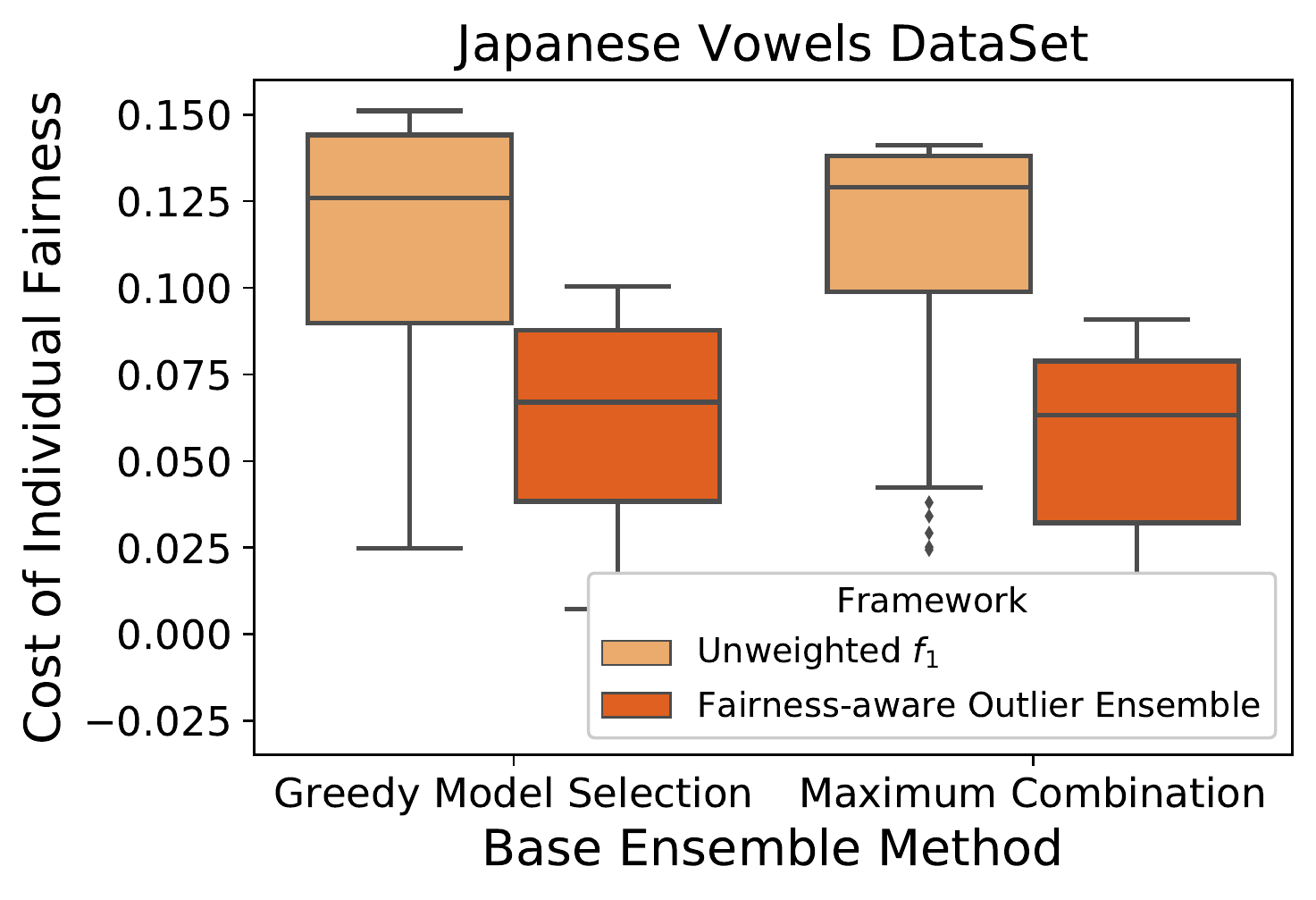}
    }
    \subfigure[Breast Cancer Dataset]{
    \label{fig:breastw_cof_individual_fairness}
        \includegraphics[height=0.15\textwidth,width=0.20\textwidth]{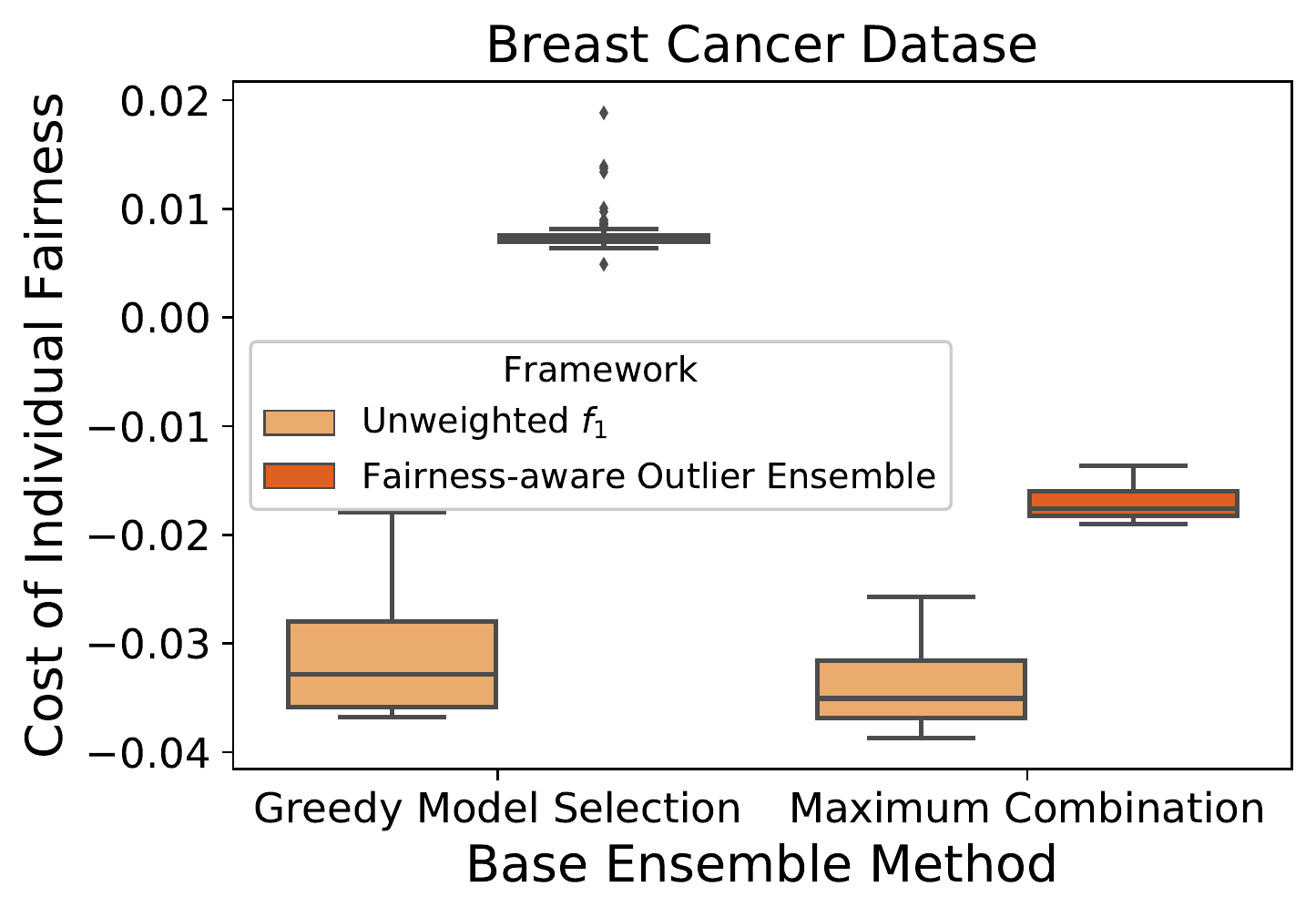}
    }
    \subfigure[Mammography Dataset]{
        \includegraphics[height=0.15\textwidth,width=0.20\textwidth]{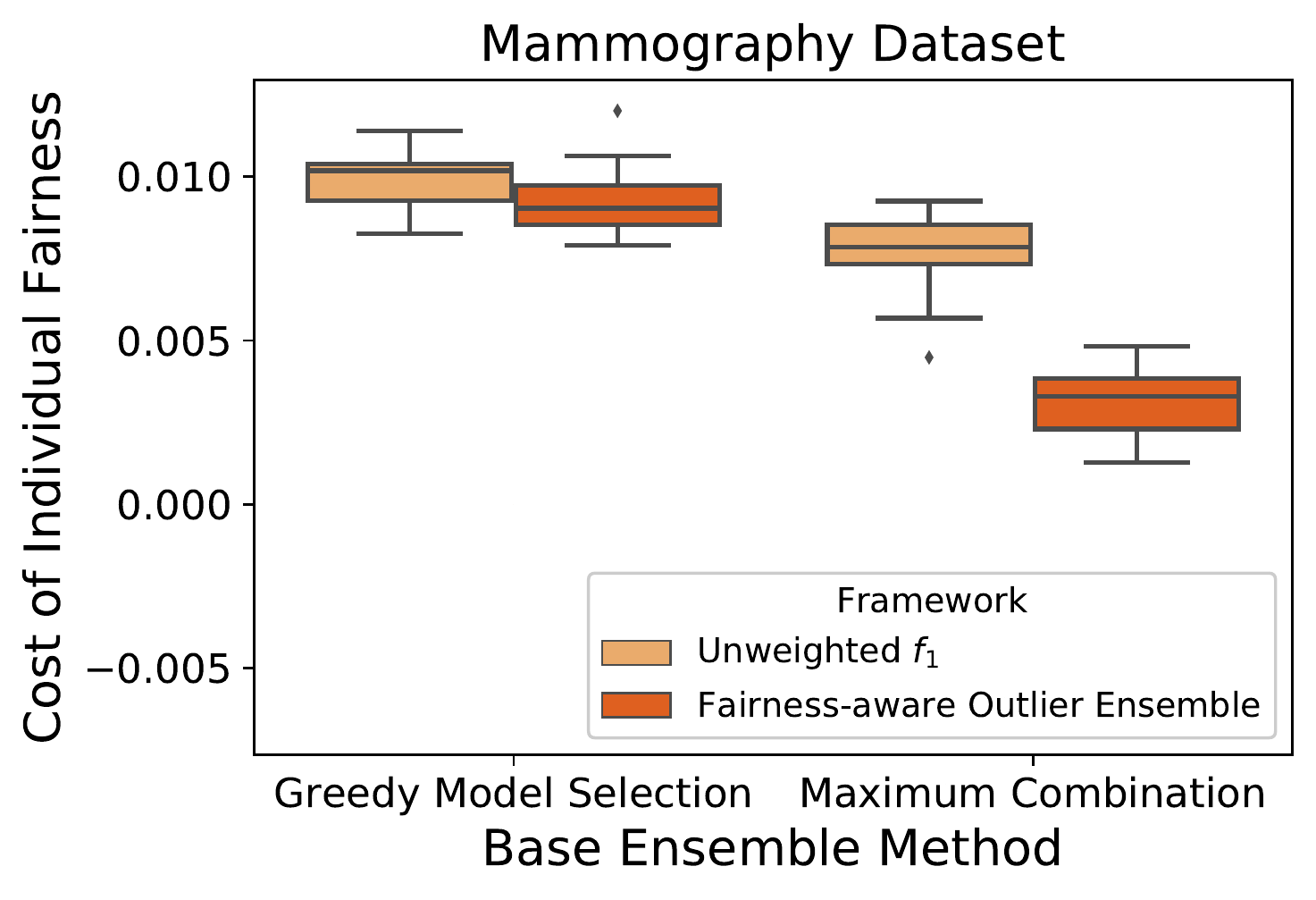}
    }
    \subfigure[Pima Dataset]{
        \includegraphics[height=0.15\textwidth,width=0.20\textwidth]{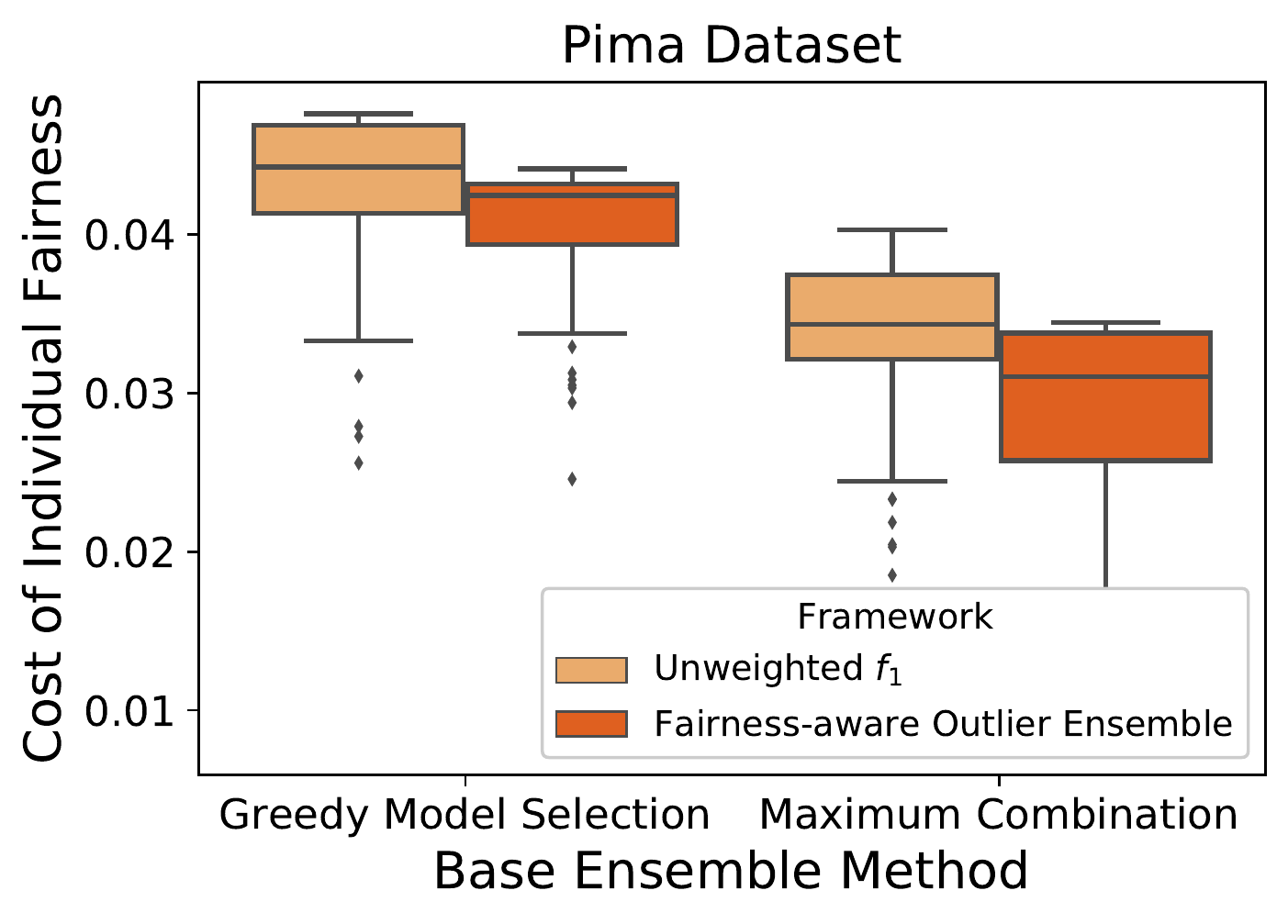}
    }    
    \caption{Cost of individual fairness.}
    \label{fig:coif}
\end{figure}
For the cost of fairness, we present the results in Fig.~\ref{fig:cogf} and Fig.~\ref{fig:coif}.
In most experiments, the proposed fairness-aware outlier ensemble framework has a lower cof than unweighted $f_1$ framework.
For the Breast Cancer Dataset where the opposite results obtained, cof values are negative. The absolute cof value of the proposed framework is lower than that of unweighted $f_1$ framework, which also verifies the effectiveness of the $\beta$ for maintaining the similar detection performance with $\mathbf{t}$.
Similar experiment results are also observed in Fig~\ref{fig:breastw_cof_individual_fairness}, which shows the cof for individual fairness.
The absolute value of cof calculated by the proposed framework is smaller.
Thus, the effectiveness of $\beta$ in $f_1$ can still be experimentally verified.

\end{document}